\documentclass[lettersize,journal]{IEEEtran}
\usepackage{amsmath,amsfonts}
\usepackage{algorithmic}
\usepackage{algorithm}
\usepackage{array}
\usepackage{textcomp}
\usepackage{stfloats}
\usepackage{url}
\usepackage{verbatim}
\usepackage{graphicx}
\usepackage{cite}

\usepackage{tikz}
\usepackage{booktabs}
\usepackage{multirow}
\usepackage{makecell,rotating}
\usepackage{fontawesome} 
\usepackage{pifont}
\usepackage{subfigure}
\usepackage{amsmath}
\usepackage{grffile}
\usepackage{color}
\usepackage{enumitem}
\usepackage{tcolorbox}
\usepackage{amssymb}
\usepackage{orcidlink}
\hypersetup{hidelinks}
\definecolor{redfont}{RGB}{219, 101, 57}
\definecolor{redbackground}{RGB}{255, 239, 234}
\definecolor{greenfont}{RGB}{5, 120, 85}
\definecolor{greenbackground}{RGB}{230, 250, 245}

\usepackage{hyperref}[colorlinks,linkcolor=blue]
\begin{document}

\title{\textsf{TrafficLLM}: Enhancing Large Language Models for Network Traffic Analysis with Generic Traffic Representation}

\author{
Tianyu Cui$^{\orcidlink{0000-0002-4467-2760}}$, Xinjie Lin$^{\orcidlink{0000-0003-0789-7570}}$, Sijia Li, Miao Chen, Qilei Yin, Qi Li,~\IEEEmembership{Senior Member,~IEEE}, \\and Ke Xu,~\IEEEmembership{Fellow,~IEEE} 
\thanks{Tianyu Cui, Xinjie Lin, Sijia Li, Miao Chen, and Qilei Yin are with the Zhongguancun Laboratory, Beijing, 100093, China (e-mail: cuity@zgclab.edu.cn; linxj@zgclab.edu.cn; lisj@zgclab.edu.cn; chenm@zgclab.edu.cn; yinql@zgclab.edu.cn).} 
\thanks{Qi Li and Ke Xu are with the Zhongguancun Laboratory, Beijing 100093, China, and also with the Tsinghua University, Beijing, 100094, China (e-mail: qli01@tsinghua.edu.cn; xuke@tsinghua.edu.cn).}
}




\maketitle

\begin{abstract}
Machine learning (ML) powered network traffic analysis has been widely used for the purpose of threat detection. Unfortunately, their generalization across different tasks and unseen data is very limited. Large language models (LLMs), known for their strong generalization capabilities, have shown promising performance in various domains. However, their application to the traffic analysis domain is limited due to significantly different characteristics of network traffic. To address the issue, in this paper, we propose \textsf{TrafficLLM}, which introduces a dual-stage fine-tuning framework to learn generic traffic representation from heterogeneous raw traffic data. The framework uses traffic-domain tokenization, dual-stage tuning pipeline, and extensible adaptation to help LLM release generalization ability on dynamic traffic analysis tasks, such that it enables traffic detection and traffic generation across a wide range of downstream tasks. We evaluate \textsf{TrafficLLM} across 10 distinct scenarios and 229 types of traffic. \textsf{TrafficLLM} achieves F1-scores of 0.9875 and 0.9483, with up to 80.12\% and 33.92\% better performance than existing detection and generation methods. It also shows strong generalization on unseen traffic with an 18.6\% performance improvement. We further evaluate \textsf{TrafficLLM} in real-world scenarios. The results confirm that \textsf{TrafficLLM} is easy to scale and achieves accurate detection performance on enterprise traffic.
\end{abstract}


\section{Introduction}\label{sec:introduction}


\IEEEPARstart{N}{etwork} traffic is the cornerstone of the Internet, carrying all interactions and transfers within the network. However, as networking techniques evolve, attackers can leverage network traffic to conduct various malicious activities, such as fishing~\cite{hu2024zipzap}, malware campaigns~\cite{fu2022encrypted}, web attacks~\cite{li2023learning}, and exploiting vulnerabilities~\cite{pellegrino2017deemon}. Considerable enterprises have recognized the importance of analyzing traffic data to detect threats, investigate incidents, and monitor environments~\cite{bujlow2015independent,shi2018efficient}. This has facilitated the development of many sophisticated network traffic analyzers (NTA) and security information and event management (SIEM) solutions, e.g., Cisco Secure Network Analytics~\cite{CISCOSecureNetworkAnalytics} and Rapid7 InsightIDR~\cite{Rapid7InsightIDR}. Existing work has achieved great improvement in many tasks, such as encrypted application classification~\cite{van2020flowprint}, website fingerprinting~\cite{wang2020high}, and malicious traffic detection~\cite{fu2023detecting}. 

In recent years, machine learning (ML) based methods~\cite{taylor2017robust,panchenko2016website,al2016adaptive} have been proposed due to their strong representation-learning abilities for diverse traffic patterns. Despite their promising potential, ML-based methods still suffer from the following limitations, leading to low generalization of existing ML-based models: \textit{(i) Generalization across various tasks}. In each sub-field of traffic analysis tasks, existing methods usually learn with handcrafted features and supervised labels to develop complicated ML models for specific tasks~\cite{MengWBXMH20,lin2022bert}. These task-specific models are hardly shared across different tasks due to the specialized handcrafted features or model designs. The development costs will be considerable in covering various tasks. \textit{(ii) Generalization to unseen data}. ML-based methods are widely criticized for their inability to handle unseen data~\cite{fu2023detecting}. These models are usually forced to learn known patterns in the high-quality labeled datasets. When faced with unseen data scenarios like concept drift~\cite{yang2021cade} and zero-day attacks~\cite{tang2020zerowall}, ML models often achieve poor performance due to low generalization.


Thus, it is vital to develop a more generic model to enhance generalization of ML models across different tasks and data distributions~\cite{lin2022bert,netFound}. Recently large language models (LLMs)~\cite{GPT-3,GPT4TechnicalReport,llama,llama2,chatglm} have shown outstanding performance in many complex tasks~\cite{CoT}. Thanks to their pattern mining, generalization to unseen data, and reproducibility across different tasks, LLMs could release remarkable capabilities in various downstream tasks~\cite{zhao2023survey}, which inspired multiple high-level views to develop specialized large-scale models for network traffic analysis. For instance, LLMs' pattern mining and reasoning ability can be utilized to learn generic representations behind IP attributes, flags, timings, and datagram lengths in traffic data. Moreover, the generalization ability allows LLMs to adapt to diverse network environments and attack scenarios. Therefore, LLM could serve as a more powerful ML model to provide traffic representation with strong generalization ability. 



 However, it is non-trivial to utilize LLMs for network traffic analysis. First, the traffic data contains considerable heterogeneous meta-information (i.e., protocol fields in packets and flows) for pattern learning, which is significantly different from the natural language. This input modality gap from the plain text makes it difficult for native LLMs to process the traffic data~\cite{GPT4TechnicalReport,llama2}, which further prevents LLMs from generalizing to traffic data of different network scenarios. Second, different downstream tasks involve diverse domain knowledge and traffic patterns (e.g., botnet traffic and Tor network traffic). Jointly learning the multi-type task-specific instruction semantics and traffic data can confuse LLM, leading to poor generalization across different tasks~\cite{ZhangY0L0C024,netFound}. Third, the traffic domain often faces environment drift like application updates and attack changes~\cite{li2023learning}. Unfortunately, LLM adaptation is extremely time-consuming due to the large model size. The high costs to update models for new environment generalization make LLM impractical in real-world scenarios. 
 
\vspace{-0.02cm}
To overcome these challenges, we propose \textsf{TrafficLLM}, a dual-stage fine-tuning framework for all open-sourced LLMs to learn generic traffic representation from expert instructions and raw traffic data, helping LLMs acquire extensive domain knowledge to enhance the generalization across diverse traffic analysis tasks and unseen scenarios. \textsf{TrafficLLM} implements three core designs: \textit{(i) Traffic-domain tokenization}: To mitigate the input modality gap between traffic and language, \textsf{TrafficLLM} is equipped with a traffic-domain tokenization mechanism to extend LLM's native tokenizers. It helps LLM generalize to different types of traffic data and achieves efficiency improvement by reducing token length. \textit{(ii) Dual-stage tuning pipeline}: \textsf{TrafficLLM} implements a dual-stage tuning pipeline to conduct multimodal learning with text and traffic data. This pipeline helps LLM accurately understand the instruction text of security experts and achieve effective traffic pattern learning in different downstream tasks, forming generic representations across different tasks. \textit{(iii) Extensible adaptation with parameter-effective fine-tuning (EA-PEFT)}: To facilitate LLM's generalization to new environments, \textsf{TrafficLLM} employs extensible adaptation using parameter-effective fine-tuning (PEFT) technique~\cite{liu2021p}. EA-PEFT splits different traffic representation abilities into various PEFT models, which helps \textsf{TrafficLLM} preserve existing capabilities and upgrade the model on new network environments.

We build \textsf{TrafficLLM} prototype to achieve generic traffic representation on ten downstream tasks, which supports traffic analysis for different applications (e.g., mobile apps, websites, and malware), protocols (e.g., HTTP, TLS1.3, and DoH), network environments (e.g., VPN, Tor, and botnet), and threats (e.g., web attacks and APT attacks). \textsf{TrafficLLM} builds generic representation for two key abilities, i.e., traffic detection and traffic generation ability, to assist traffic analyst in their daily attack detection and red teaming work, with 5.90\%-80.12\% and 3.07\%-33.92\% of performance improvement over existing ML methods. We further evaluate \textsf{TrafficLLM} on unseen environments and real-world scenarios. \textsf{TrafficLLM} shows stronger generalization compared to existing ML models.  

\vspace{0.1cm}
\noindent \textbf{Contributions.} Our contributions can be shown as follows:
\begin{itemize}[nolistsep,leftmargin=*]
	\item We develop \textsf{TrafficLLM}, a dual-stage fine-tuning framework learning with extensive expert instruction and raw traffic data, which helps LLM obtain generic traffic representation from the domain knowledge to achieve strong generalization across diverse traffic analysis tasks.  
	\item We build \textsf{TrafficLLM} with three core techniques to overcome the challenges of using LLMs in the traffic domain. \textsf{TrafficLLM} employs traffic-domain tokenization to mitigate the modality gap and generalize to heterogeneous data, dual-stage tuning pipeline to conduct multimodal learning across diverse tasks, and EA-PEFT to realize generalization to new environments.
	\item We construct the first large-scale traffic-domain LLM adaptation dataset for future research. To the best of our knowledge, we have collected the largest LLM tuning dataset for the traffic domain to date, which includes $\approx$ 0.4M samples consisting of instruction text and traffic data supervised by experts and AI assistants. 
	\item We conduct extensive experiments on various downstream tasks to demonstrate the superiority of \textsf{TrafficLLM}. \textsf{TrafficLLM} achieves better performance with generic representation compared to 15 state-of-the-art traffic detection or generation methods. Moreover, \textsf{TrafficLLM} obtains strong generalization on unseen data and real-world settings.
\end{itemize}	

\noindent \textbf{Website demo and datasets.} 
We provide \textsf{TrafficLLM}'s demo, source codes and all the tuning datasets at \href{https://github.com/ZGC-LLM-Safety/TrafficLLM}{https://github.com/ZGC-LLM-Safety/TrafficLLM}.

\section{Problem Statement \& Threat Model}\label{sec:preliminaries}

\begin{table}[t!]
\caption{The basic information and model capabilities of current mainstream LLMs and traffic-domain PLMs. ("\faCheckCircle" = has the ability. "\faTimesCircleO" = doesn't have the ability. "\faCircleO" = has the basic ability but still has shortcomings)}
\vspace{-0.2cm}
\begin{center}
\resizebox{0.48\textwidth}{!}{
\begin{tabular}{l|cccc|cc|c}
\toprule
\multirow{3}{*}{\textbf{Model}} && \multicolumn{2}{c}{\textbf{Basic Information}} && \multicolumn{3}{c}{\textbf{Model Capability}}\\
\cmidrule{2-8}
&& Model & Open && Traffic & Traffic & Language\\
&& Size & Source && Detection & Generation & Processing\\
\midrule
Llama3~\cite{llama3} && 8B/70B & Yes && \faTimesCircleO & \faTimesCircleO & \faCheckCircle\\
Gemini1.5~\cite{Gemini} && Unk. & No && \faTimesCircleO & \faTimesCircleO & \faCheckCircle\\
Claude3~\cite{Claude} && Unk. & No && \faTimesCircleO & \faTimesCircleO & \faCheckCircle\\
Mistral Large 2~\cite{mistral} && Unk. & No && \faTimesCircleO & \faTimesCircleO & \faCheckCircle\\
GPT-4~\cite{GPT4TechnicalReport} && Unk. & No && \faTimesCircleO & \faTimesCircleO & \faCheckCircle\\
GLM-4~\cite{chatglm} && 9B/130B & Yes &&  \faTimesCircleO & \faTimesCircleO & \faCheckCircle\\
Baichuan4~\cite{baichuan} && 53B & Yes && \faTimesCircleO & \faTimesCircleO & \faCheckCircle\\
\midrule
ET-BERT~\cite{lin2022bert} && 0.1B & Yes &&  \faCheckCircle & \faTimesCircleO & \faTimesCircleO\\
PERT~\cite{he2020pert} && 0.04B & Yes && \faCheckCircle & \faTimesCircleO & \faTimesCircleO\\
netFound~\cite{netFound} && 0.2B & No && \faCheckCircle & \faTimesCircleO & \faTimesCircleO\\
NetGPT~\cite{NetGPT} && 0.1B & No && \faCheckCircle & \faCircleO & \faTimesCircleO\\
Lens~\cite{Lens} && 0.25B & No && \faCheckCircle & \faCircleO & \faTimesCircleO\\
\midrule
\textsf{TrafficLLM} && 6B/12B & Yes && \faCheckCircle & \faCheckCircle & \faCheckCircle\\
\bottomrule
\end{tabular}
}
\label{tab1}
\end{center}
\vspace{-0.2cm}
\end{table}

\subsection{Problem Statement}\label{section2.1}

\noindent \textbf{Investigation on Existing LLMs.}
We survey existing LLM's capabilities for network traffic analysis. Due to the difficulty of processing traffic data, we have not witnessed the deployment of the foundation model for network traffic analysis in the industry. As of December 2024, we have compiled the model capabilities of the current mainstream LLMs in Table~\ref{tab1}. Although the existing LLMs have initially possessed certain knowledge in the network security field~\cite{zhao2023survey}, the majority of them can not accomplish traffic analysis tasks due to the lack of capabilities for traffic processing. Most LLMs lack insights from traffic data learning. They can only respond to basic instructions with inaccurate conclusions. To overcome the problem, a series of works like ET-BERT~\cite{lin2022bert} and PERT~\cite{he2020pert} have been developed in recent years to process traffic data with pre-trained language models (PLMs)~\cite{zhao2023survey}, aiming to build effective traffic detection and generation capabilities. However, they have several shortcomings: 
\begin{itemize}[nolistsep,leftmargin=*]
    \item \textbf{High development cost.} These approaches mainly utilize pre-training techniques~\cite{DevlinCLT19}, which has a high training time and resource cost. Compared with fine-tuning, they can not inherit existing LLMs' abilities, which is less practical.
	\item \textbf{Limited model size.} These models are usually smaller than 1B. Strictly speaking, they are not part of LLMs~\cite{zhao2023survey}. These models may lose LLM's surprising emergent abilities~\cite{WeiTBRZBYBZMCHVLDF22} with strong generalization that can not be present in small models.
	\item \textbf{Narrow scenarios.} These efforts only train on traffic datasets. They fall short in handling natural language, rendering them incapable of following instructions and conducting complex traffic analysis tasks~\cite{liu2023pre}, requiring a high user threshold.
	\item \textbf{Defective abilities.} These methods have shortcomings in traffic detection and generation. They do not have the generalization abilities to detect traffic on unseen data~\cite{fu2023detecting}. Moreover, they can only generate 5-tuples of packets and flows, whose practical uses are very limited~\cite{NetGPT,Lens}.
\end{itemize}

In this paper, we aim to exploit LLM to facilitate the work of network traffic analysis with strong generalization across different tasks. Given the language instructions that involve diverse domain knowledge and the traffic data that contains multi-type benign or malicious traffic, an adapted LLM in the traffic domain is expected to learn generic traffic representation with its accuracy and generalization in pattern learning. Note that \textit{all work can be completed through dialogue with the generic model}, which reduces the operational threshold and development costs for security practitioners. However, the characteristics of the traffic domain leave three challenges to realizing LLM's generalization on traffic analysis tasks.


\begin{figure}[t]
\centering
\subfigure{    
\includegraphics[width=2.6cm]{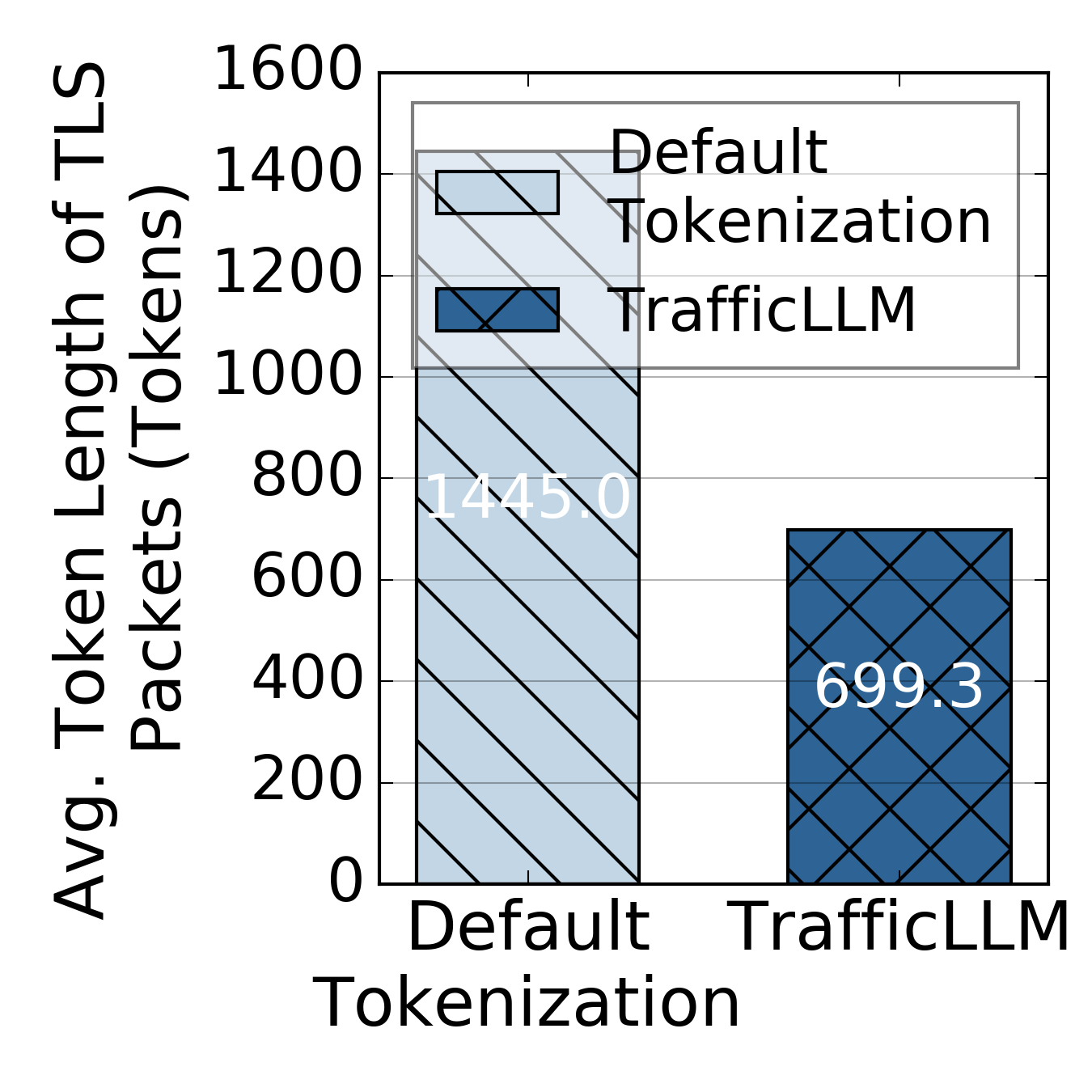}
}\label{fig1a}
\subfigure{ 
\includegraphics[width=2.6cm]{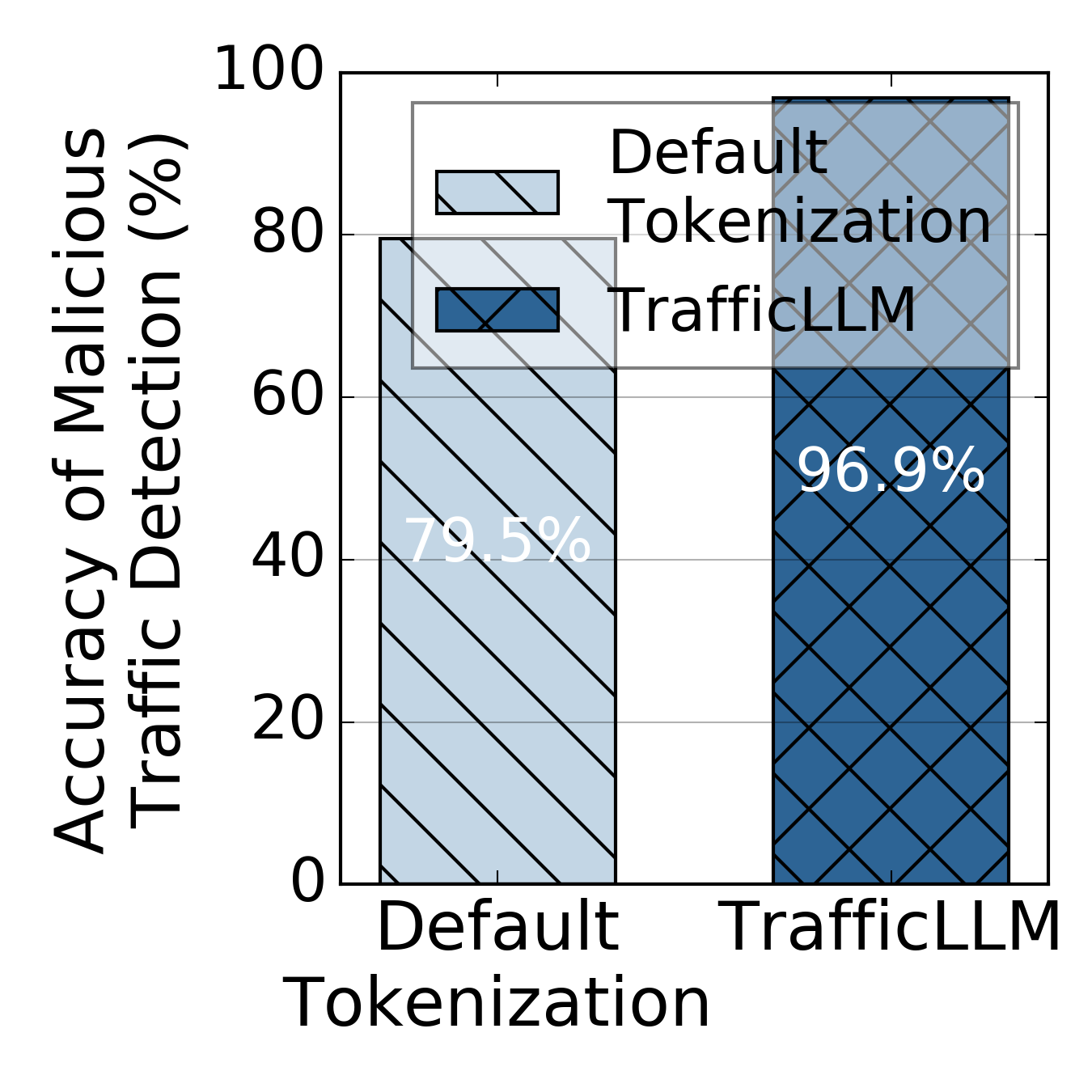}
}\label{fig1b}
\subfigure{ 
\includegraphics[width=2.6cm]{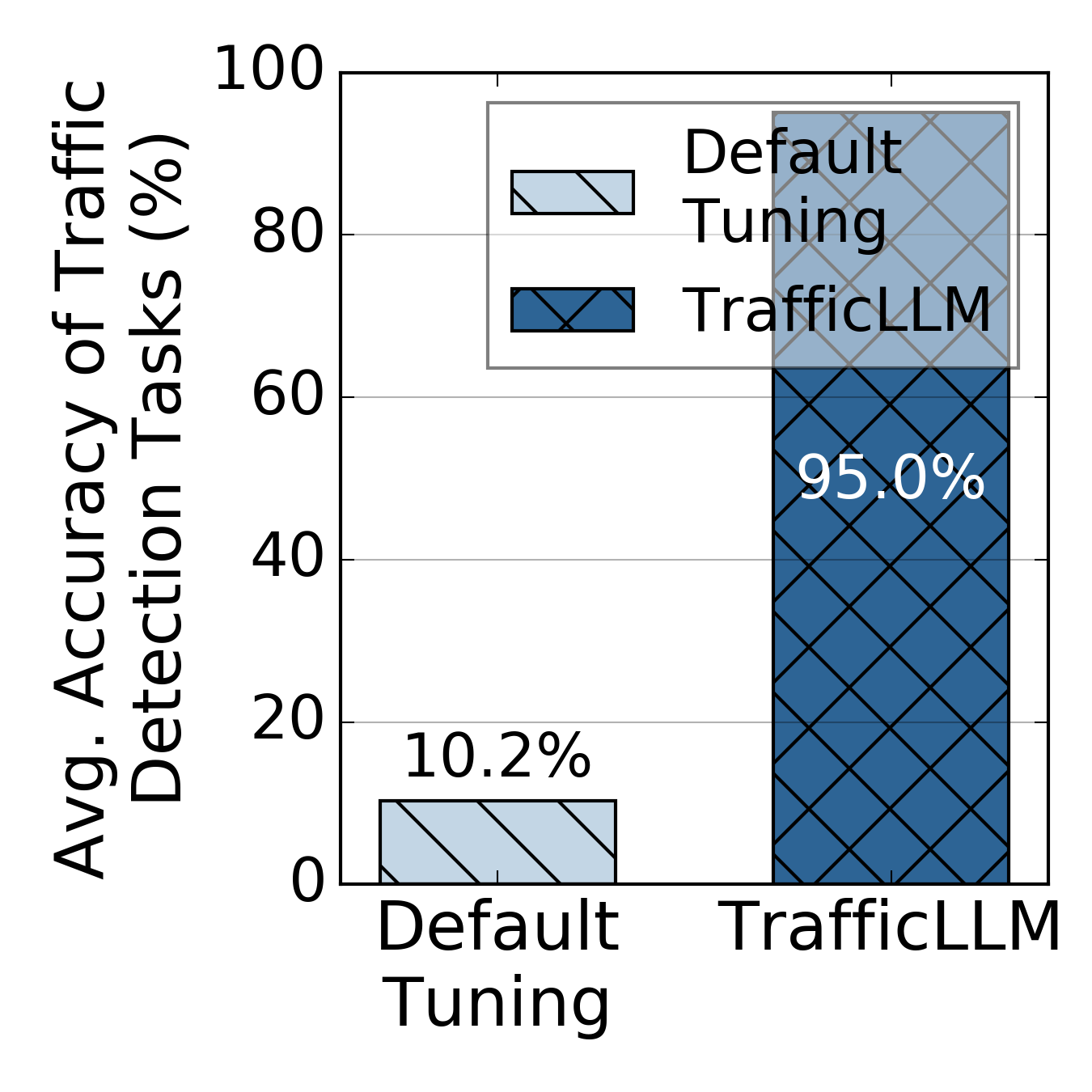}
}\label{fig1c}
\centering
\setlength{\abovecaptionskip}{-0.3cm}
\caption{Native LLM's limitation to handle traffic data with default tokenization and tuning strategies. Left and Middle: LLM is ineffective and inaccurate in loading traffic data with language tokens directly. Right: LLM suffers from learning multi-type semantics and traffic data at the same stage.}
\vspace{-0.4cm}
\label{fig1}
\end{figure}

\begin{figure}[t]
\centering
\subfigure{    
\includegraphics[width=2.6cm]{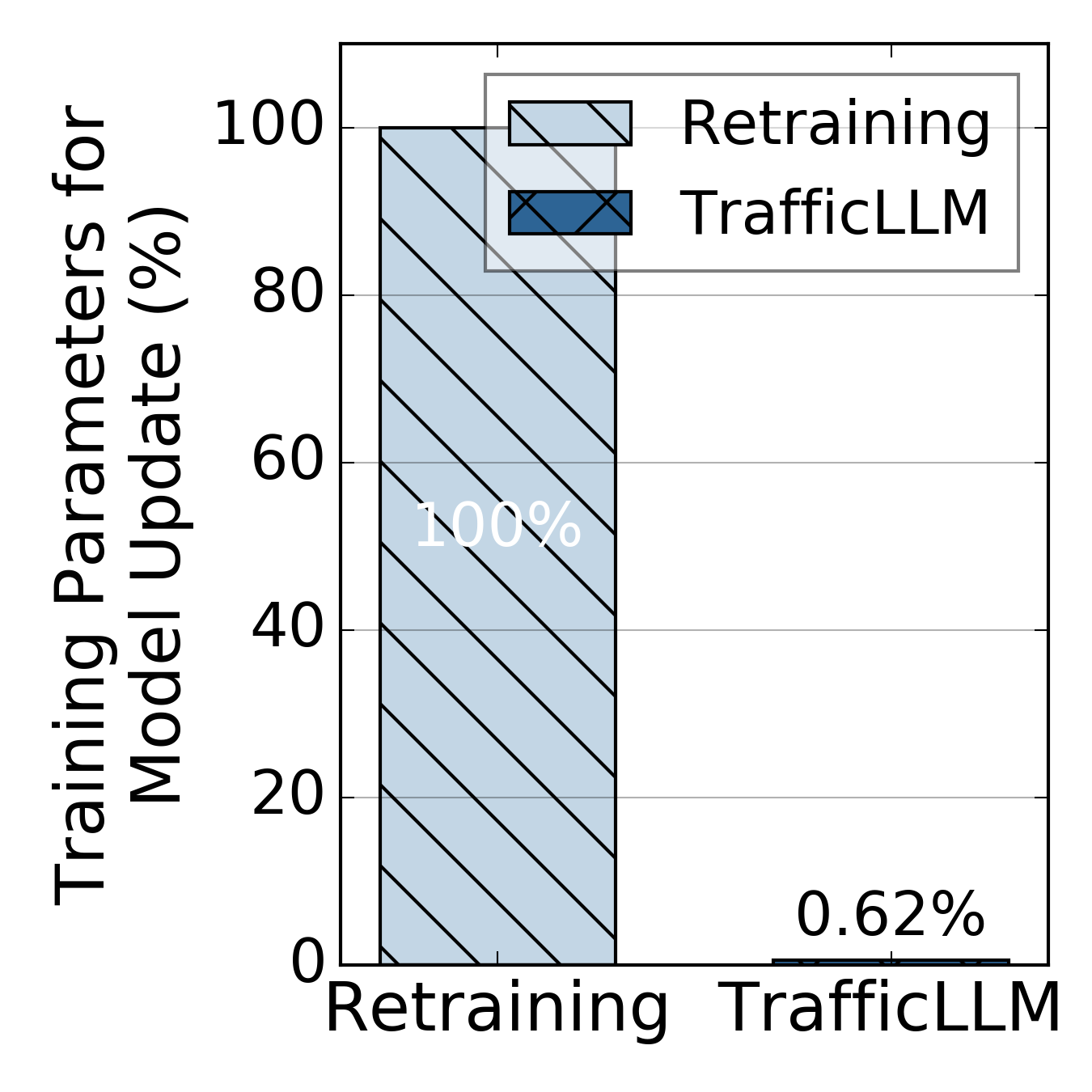}
}\label{fig2a}
\subfigure{ 
\includegraphics[width=2.6cm]{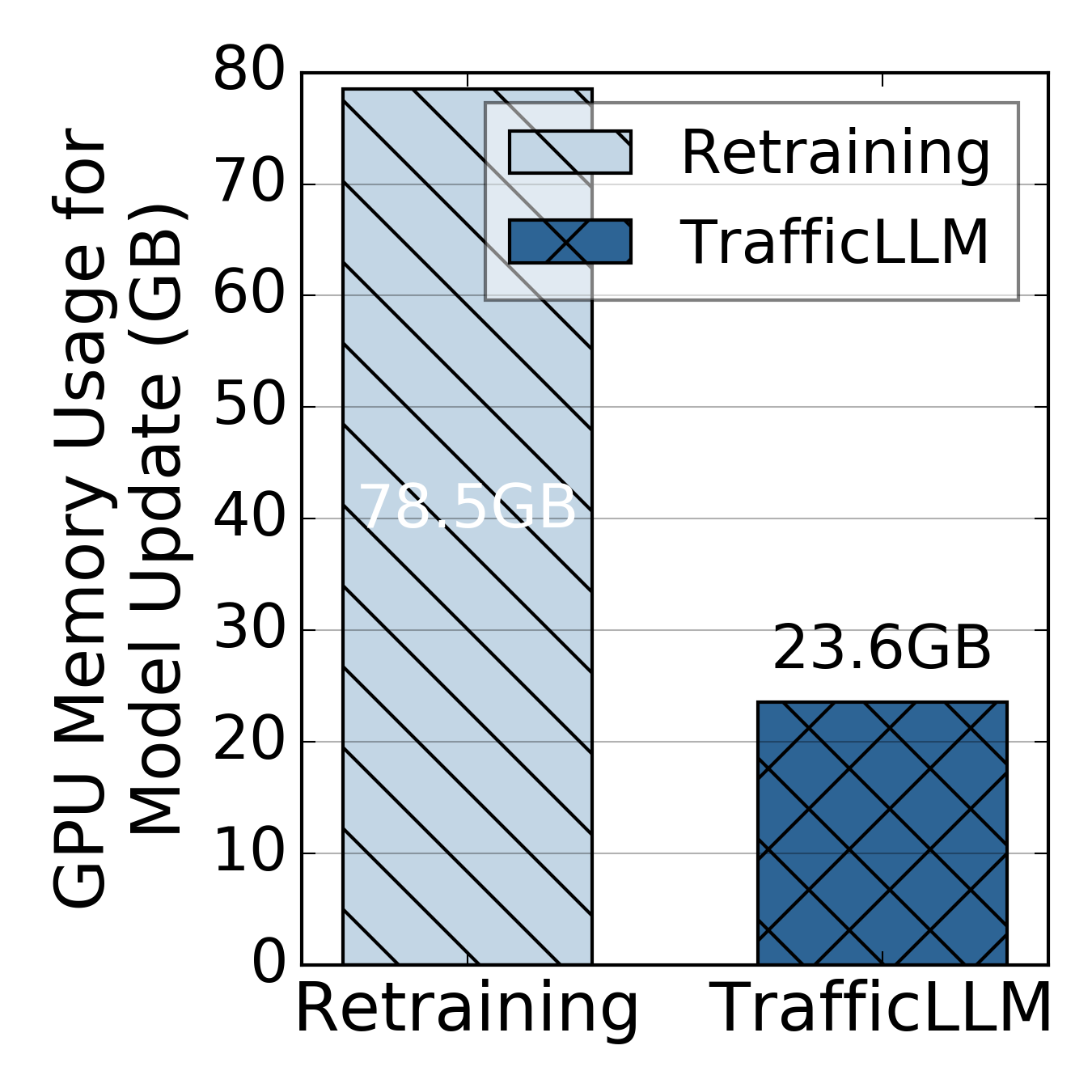}
}\label{fig2b}
\subfigure{ 
\includegraphics[width=2.6cm]{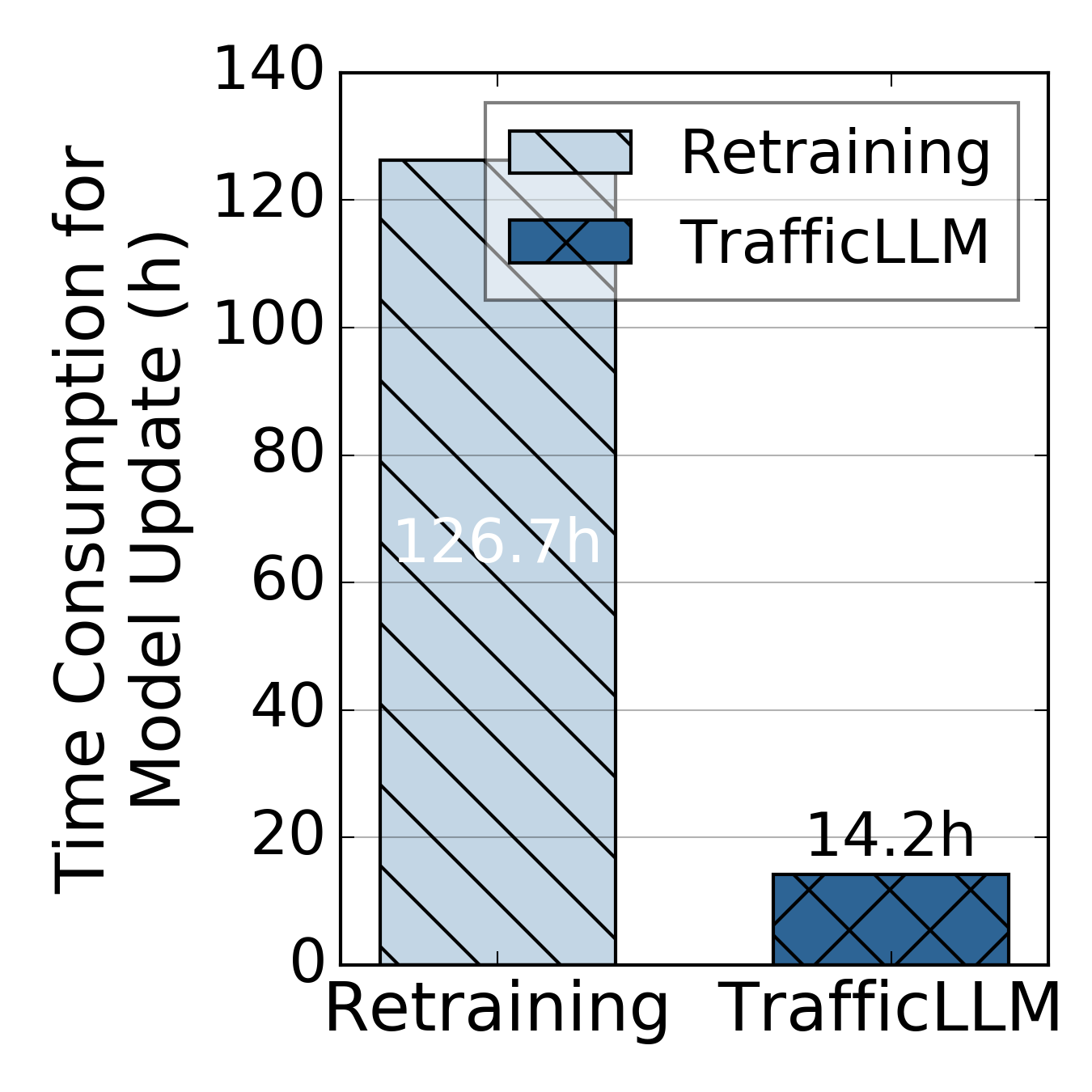}
}\label{fig2c}
\centering
\setlength{\abovecaptionskip}{-0.3cm}
\caption{The adaptation costs of LLM's retraining to update traffic detection capabilities on new scenarios. \textsf{TrafficLLM} employs EA-PEFT to reduce the cost by using multiple external parameters to encapsulate different capabilities.} 
\label{fig2}
\vspace{-0.4cm}
\end{figure}
\vspace{0.1cm}
\noindent \textbf{Challenge 1: Generalization to heterogeneous input of traffic data.} Traffic data consists of structured metadata in packets and flows (e.g., IPs and ports). However, most LLMs are considered as the specialized model for processing plain text, which has a huge gap from the traffic data. Before being fed into LLM, the text is converted into language tokens using a standard tokenizer~\cite{GPT4TechnicalReport,llama2}. These tokenizers are usually trained on large-scale text corpora with tokenization algorithms like WordPiece and Byte-Pair Encoding (BPE), which rarely see heterogeneous traffic data. Consequently, LLM may fail to directly transform traffic data into textual formats and load them with default tokenization. 

To give an instance, we adapt Llama2-7B~\cite{llama2} to conduct the malware traffic detection (MTD)~\cite{wang2017malware} task with its native tokenizer. First, it is not effective to split traffic information as the input. As shown in Figure~\ref{fig1} (left), the default tokenizer will produce many redundancies when processing metadata in TLS packets. It may reduce the efficiency of LLMs in realistic traffic analysis work. Second, the transformed tokens are not performed well to ensure detection accuracy. In Figure~\ref{fig1} (middle), the performance of native LLM is not remarkable on the MTD task (only 79.5\% of accuracy on USTC-TFC 2016 dataset~\cite{wang2017malware}) since the unsuitable tokenization split key features incorrectly, leading to the failure to capture distinct patterns between benign and malicious traffic.

\vspace{0.1cm}
\noindent \textbf{Challenge 2: Generalization across different tasks with multimodal learning.} Network traffic analysis covers a wide range of specific tasks, including detecting and generating attack traffic in different scenarios (e.g., the MTD task). It involves diverse task-specific knowledge in the instruction to prompt LLM to conduct different work. Moreover, these downstream tasks usually point to different network environments, which involve representations learned from multi-type traffic meta-information (e.g., packet lengths in encrypted application classification (EAC)~\cite{liu2019fs} and HTTP request headers in web attack detection (WAD)~\cite{li2023learning}). These complexities of instructions and traffic patterns can easily confuse LLM when facing multimodal learning~\cite{ZhangY0L0C024,netFound}. As depicted in Figure~\ref{fig1} (right), we directly mix the training data of three traffic detection tasks (i.e., MTD, EAC, and WAD tasks) and train Llama2 with the default tuning strategy. Llama2 only reaches 10.2\% of average accuracy, indicating the difficulty for LLM to learn with multimodal across different tasks.

\vspace{0.1cm}
\noindent \textbf{Challenge 3: Generalization to new environments with model update.} The adaptation cost of LLM is quite expensive as it requires training large-scale parameters with extensive datasets~\cite{GPT4TechnicalReport,GPT-3}. However, many traffic analysis tasks often need to update the model's traffic representation to struggle with dynamic scenarios, which are raised by the application version updates (e.g., concept drift~\cite{yang2021cade}) and attack method changes (e.g., APT attacks~\cite{myneni2020dapt}). The high adaptation costs of LLMs prevent the update of traffic representations on new scenarios. As shown in Figure~\ref{fig2}, we measure Llama2-7B's adaptation overhead on 5 NIVIDA A100-80GB GPUs for traffic detection tasks. Traditional retraining methods consume 78.5GB GPU memory and 126.7h to adapt to new environments for one epoch, which is unacceptable in real-world dynamic scenarios.

\begin{figure*}[t]
\centering
\includegraphics[width=1.0\textwidth]{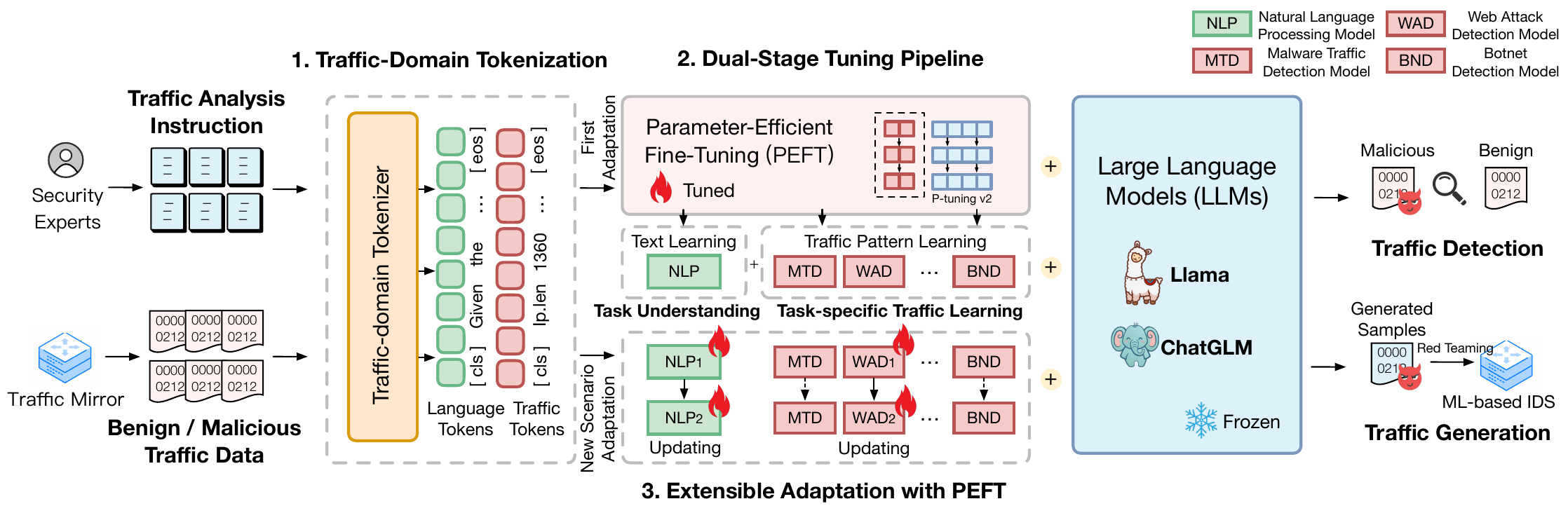}
\setlength{\abovecaptionskip}{-0.3cm} 
\caption{The overall framework of \textsf{TrafficLLM}. \textsf{TrafficLLM} employs three core techniques: \textit{traffic-domain tokenization} to process instructions and traffic data, \textit{dual-stage tuning pipeline} to learn text semantics and traffic patterns across different tasks, \textit{extensible adaptation with parameter-effective fine-tune} to update model parameters for new scenario adaptation.}
\label{fig3:framework}
\vspace{-0.3cm}
\end{figure*}

\subsection{Threat Model} 
Our purpose is to develop an LLM for traffic representation, which can be leveraged to construct traffic detection and generation methods to replace traditional ML-based methods that can be integrated into existing sophisticated network traffic analyzers (NTA)~\cite{CISCOSecureNetworkAnalytics} and security information and event management (SIEM) systems~\cite{SIEM}, which are widely deployed in Security Operation Centers (SOCs) to analyze abnormal events based on traffic mirroring and logs. Different from the threat model of the existing ML-based traffic detection and generation studies~\cite{van2020flowprint,yin2022practical}, \textsf{TrafficLLM} aims to develop LLM-based model to adapt to different tasks, which is more generalized than specific ML models. \textsf{TrafficLLM} can be directly driven by instructions from experts. It achieves generic traffic representation using domain knowledge to extract task-related traffic patterns from raw traffic based on instructions. Leveraging LLM's pattern learning and generalization abilities, we develop \textsf{TrafficLLM} to build the centralized traffic analysis solution, which can achieve the following goals:

\begin{itemize}[nolistsep,leftmargin=*]
	\item \textbf{Attack detection.} \textsf{TrafficLLM} establishes comprehensive traffic detection capabilities to process and analyze diverse benign and malicious traffic. Learning with a variety of heterogeneous traffic data, \textsf{TrafficLLM} can extract generic traffic representation from raw traffic to identify benign and malicious categories~\cite{lotfollahi2020deep,van2020flowprint}, or realize more fine-grained classification (e.g., application types in encrypted application classification (EAC)~\cite{liu2019fs,lin2022bert} and network types in botnet detection (BND)~\cite{beigi2014towards}).
	\item \textbf{Attack synthesis.} \textsf{TrafficLLM} can generate attack samples to facilitate red teaming and enhance the robustness of network-based IDSes (NIDS)~\cite{jan2020throwing} when lacking the high-quality traffic data in practical scenarios. Different from existing ML-based traffic generation studies~\cite{yin2022practical,NetGPT,Lens}, \textsf{TrafficLLM} can generate a wide range of target traffic with \verb|.pcap| format based on LLM's strong memorization. It can help security practitioners simulate traffic attacks to measure the vulnerability of systems and build robust IDSes through data augmentation.
\end{itemize}

\section{Design of \textsf{TrafficLLM}}\label{sec:methodology}

\subsection{Overall Framework}
We develop \textsf{TrafficLLM}, which captures generic traffic representation from diverse traffic-domain instruction text and raw traffic data through a dual-stage fine-tuning framework, aiming to release LLM's strong generalization across diverse traffic analysis tasks. \textsf{TrafficLLM} overcomes the challenges of applying LLM in traffic analysis. It builds domain knowledge by training with extensive expert instructions and traffic data. Driven by expert instructions, \textsf{TrafficLLM} automatically extracts task-related traffic patterns from raw packets and flows, forming the generic representation across different tasks. We present the architecture overview of \textsf{TrafficLLM} in Figure~\ref{fig3:framework}. \textsf{TrafficLLM} designs three modules:

\vspace{0.1cm}
\noindent \textbf{Traffic-Domain Tokenization.} To overcome the modality gap between natural language and heterogeneous traffic data, \textsf{TrafficLLM} employs traffic-domain tokenization to process the diverse input of traffic detection and generation tasks for representation learning. This mechanism effectively extends LLM's native tokenizer by training the specialized tokenization model on large-scale traffic-domain corpora (Section~\ref{sec3:instruction-design}).
 
\vspace{0.1cm}
\noindent \textbf{Dual-Stage Tuning Pipeline.} \textsf{TrafficLLM} designs a dual-stage tuning pipeline to achieve LLM's generic representation learning across different traffic-domain tasks. The pipeline trains LLM to understand instructions and learn task-related traffic patterns at different stages, which builds upon \textsf{TrafficLLM}'s domain knowledge to learn traffic representations for diverse traffic detection and generation tasks (Section~\ref{sec3:instruction-tuning}).

\vspace{0.1cm}
\noindent \textbf{Extensible Adaptation with Parameter-Effective Fine-Tuning (EA-PEFT).} To adapt LLM for generalization to new traffic environments, \textsf{TrafficLLM} proposes an extensible adaptation with parameter-effective fine-tuning (EA-PEFT) to update model parameters with low overhead. The technique splits model capabilities in different PEFT models, which helps minimize the adaptation costs on dynamic scenarios raised by traffic pattern changes (Section~\ref{sec3:applications}).

\subsection{Traffic-Domain Tokenization} \label{sec3:instruction-design}
\textsf{TrafficLLM} utilizes traffic-domain tokenization to encode original inputs of traffic analysis tasks and makes them learnable for LLMs. We extract tuning data from raw traffic to train the specialized tokenizer, which effectively extends LLM's native tokenizer. The principle of traffic-domain tokenization is to map the natural language and traffic data into the same feature space, supporting LLM in accepting heterogeneous traffic data to build the representation.

\vspace{0.1cm}
\noindent \textbf{Tuning Data Extraction.} To help LLMs reduce the modality gap and make them feasible to handle traffic data across different tasks, \textsf{TrafficLLM} directly extracts training data from raw traffic for generic traffic representation. The purpose of using raw traffic is to release the generalization of \textsf{TrafficLLM} in different scenarios, different from the traditional ML-based methods that heavily rely on predefined features. Rather than select for certain features, \textsf{TrafficLLM} leverages the whole meta-information in the packets to learn the important features without human guidance. It helps \textsf{TrafficLLM} obtain strong generalization across different scenarios.

To facilitate LLMs acquire domain knowledge to learn traffic representations across different tasks, \textsf{TrafficLLM} employs instruction-learning~\cite{liu2023pre} to build tuning data templates and adapt LLMs to the traffic-domain semantic space. These instructions can guide LLM to automatically extract task-related patterns in the traffic data to build representations. Then, we utilize Tshark to extract protocol fields in different packet layers. These meta-information are organized by pairs including the field name and the corresponding value (e.g., \verb|tcp.srcport: 443|). To indicate the beginning of traffic data, we define an indicator token \verb|<packet>| in the context. Each packet data is started with this special indicator to form the flow data. This instruction-learning design helps LLM obtain domain knowledge to capture valuable semantics for pattern learning across different tasks. Finally, \textsf{TrafficLLM} combines traffic analysis instructions and the extracted traffic data to build the tuning data. The examples of \textsf{TrafficLLM}'s tuning data are shown as follows.

\begin{tcolorbox}[colback=gray!10, title=Traffic Detection Tuning Data Example]
\textbf{Instruction:} Given the following traffic data that contains protocol fields, traffic features, and payloads. Please conduct the \colorbox{redbackground}{\textcolor{redfont}{Malicious Traffic Detection task}} to determine which application category the encrypted benign or malicious traffic belongs to. \\
\textbf{<packet>:} ip.len: 1360, ip.proto: 6, tcp.srcport: 443, tcp.dstport: 56603, tcp.len: 1308, tcp.window\_size: ...\\
\textbf{Output:} \colorbox{greenbackground}{\textcolor{greenfont}{Zeus}}.
\end{tcolorbox}

\begin{tcolorbox}[colback=gray!10, title=Traffic Generation Tuning Data Example]
\textbf{Instruction:} Based on the protocol fields, traffic features, and payloads of traffic in your knowledge. Please generate \colorbox{redbackground}{\textcolor{redfont}{a packet of Skype traffic}}.\\
\textbf{Output:} \colorbox{greenbackground}{\textcolor{greenfont}{Skype.pcap}}(ip.src: 1.2.102.211, ip.dst: 1.1.210.113, tcp.srcport:443, tcp.dstport: 27567, raw: 1021ac5010000021ac50200000800450002aeea10400\\
0200632f2010266d30101...).
\end{tcolorbox}

\vspace{0.1cm}
\noindent \textbf{Tokenizer Training.} After extracting the tuning data, we build a specialized traffic-domain tokenizer to form the input traffic tokens. We use the BPE method to train the specialized tokenizer on the large-scale tuning data. Since native LLM has hardly ever seen traffic data, it can be considered as an extension to the existing tokenizer. Table~\ref{tab:tokenizer} shows an example of tokenization of \textsf{TrafficLLM} and an open-sourced LLM ChatGLM2~\cite{chatglm}. Tokenizers of existing LLMs tend to split the field names (e.g., \verb|checksum| and \verb|tcp|) since they have learned less of the traffic-domain languages. In contrast, \textsf{TrafficLLM}'s tokenizer can retain these field name indicators based on their appearance frequency in training data. It can also store the common field values (e.g., window sizes and flags), helping LLMs learn the numerical meta-information correctly. Furthermore, due to the accurate tokenization on traffic data, \textsf{TrafficLLM} can produce shorter packet tokens with a 699.36 average token length, compared to ChatGLM2's 1445.04 token length. It helps \textsf{TrafficLLM} obtain faster packet processing efficiency compared to the native LLMs.

As shown in Figure~\ref{fig1} (left) and (middle), \textsf{TrafficLLM}'s tokenization is more effective and accurate in loading traffic data compared to the default tokenization. This mechanism helps \textsf{TrafficLLM} reach 106\% efficiency improvement to process traffic data. \textsf{TrafficLLM} also achieves 17.4\% better performance on MTD tasks by using the traffic-domain tokenization.

\begin{table}[t]
\caption{The tokenization of \textsf{TrafficLLM}'s tokenizer on the traffic data compared to ChatGLM2's tokenizer.}
\vspace{-0.4cm}
\begin{center}
\resizebox{0.48\textwidth}{!}{
\begin{tabular}{p{10.5cm}}
\toprule
\textbf{Default tokenization:}  \colorbox{redbackground}{\textcolor{redfont}{\_ip}} \colorbox{redbackground}{\textcolor{redfont}{.}} \colorbox{redbackground}{\textcolor{redfont}{proto}} \colorbox{redbackground}{\textcolor{redfont}{:}} \colorbox{redbackground}{\textcolor{redfont}{\_}} \colorbox{redbackground}{\textcolor{redfont}{6}} \colorbox{redbackground}{\textcolor{redfont}{,}} \colorbox{redbackground}{\textcolor{redfont}{\_ip}} \colorbox{redbackground}{\textcolor{redfont}{.}} \colorbox{redbackground}{\textcolor{redfont}{che}} \colorbox{redbackground}{\textcolor{redfont}{cks}} \colorbox{redbackground}{\textcolor{redfont}{um}} \colorbox{redbackground}{\textcolor{redfont}{:}} \colorbox{redbackground}{\textcolor{redfont}{\_}} \colorbox{redbackground}{\textcolor{redfont}{0}} \colorbox{redbackground}{\textcolor{redfont}{x}} \colorbox{redbackground}{\textcolor{redfont}{0}} \colorbox{redbackground}{\textcolor{redfont}{0}} \colorbox{redbackground}{\textcolor{redfont}{0}} \colorbox{redbackground}{\textcolor{redfont}{0}} \colorbox{redbackground}{\textcolor{redfont}{1}} \colorbox{redbackground}{\textcolor{redfont}{7}} \colorbox{redbackground}{\textcolor{redfont}{a}} \colorbox{redbackground}{\textcolor{redfont}{7}} \colorbox{redbackground}{\textcolor{redfont}{,}} \colorbox{redbackground}{\textcolor{redfont}{\_ip}} \colorbox{redbackground}{\textcolor{redfont}{.}} \colorbox{redbackground}{\textcolor{redfont}{che}} \colorbox{redbackground}{\textcolor{redfont}{cks}} \colorbox{redbackground}{\textcolor{redfont}{um}} \colorbox{redbackground}{\textcolor{redfont}{.}} \colorbox{redbackground}{\textcolor{redfont}{status}} \colorbox{redbackground}{\textcolor{redfont}{:}} \colorbox{redbackground}{\textcolor{redfont}{\_}} \colorbox{redbackground}{\textcolor{redfont}{2}} \colorbox{redbackground}{\textcolor{redfont}{,}} \colorbox{redbackground}{\textcolor{redfont}{\_ip}} \colorbox{redbackground}{\textcolor{redfont}{.}} \colorbox{redbackground}{\textcolor{redfont}{src}} \colorbox{redbackground}{\textcolor{redfont}{:}} \colorbox{redbackground}{\textcolor{redfont}{\_}} \colorbox{redbackground}{\textcolor{redfont}{1}} \colorbox{redbackground}{\textcolor{redfont}{0}} \colorbox{redbackground}{\textcolor{redfont}{.}} \colorbox{redbackground}{\textcolor{redfont}{0}} \colorbox{redbackground}{\textcolor{redfont}{.}} \colorbox{redbackground}{\textcolor{redfont}{2}} \colorbox{redbackground}{\textcolor{redfont}{.}} \colorbox{redbackground}{\textcolor{redfont}{1}} \colorbox{redbackground}{\textcolor{redfont}{5}} \colorbox{redbackground}{\textcolor{redfont}{,}} \colorbox{redbackground}{\textcolor{redfont}{\_ip}} \colorbox{redbackground}{\textcolor{redfont}{.}} \colorbox{redbackground}{\textcolor{redfont}{d}} \colorbox{redbackground}{\textcolor{redfont}{st}} \colorbox{redbackground}{\textcolor{redfont}{:}} \colorbox{redbackground}{\textcolor{redfont}{\_}} \colorbox{redbackground}{\textcolor{redfont}{1}} \colorbox{redbackground}{\textcolor{redfont}{9}} \colorbox{redbackground}{\textcolor{redfont}{8}} \colorbox{redbackground}{\textcolor{redfont}{.}} \colorbox{redbackground}{\textcolor{redfont}{5}} \colorbox{redbackground}{\textcolor{redfont}{2}} \colorbox{redbackground}{\textcolor{redfont}{.}} \colorbox{redbackground}{\textcolor{redfont}{2}} \colorbox{redbackground}{\textcolor{redfont}{0}} \colorbox{redbackground}{\textcolor{redfont}{0}} \colorbox{redbackground}{\textcolor{redfont}{.}} \colorbox{redbackground}{\textcolor{redfont}{3}} \colorbox{redbackground}{\textcolor{redfont}{9}} \colorbox{redbackground}{\textcolor{redfont}{,}} \colorbox{redbackground}{\textcolor{redfont}{\_t}} \colorbox{redbackground}{\textcolor{redfont}{cp}} \colorbox{redbackground}{\textcolor{redfont}{.}} \colorbox{redbackground}{\textcolor{redfont}{src}} \colorbox{redbackground}{\textcolor{redfont}{port}} \colorbox{redbackground}{\textcolor{redfont}{:}} \colorbox{redbackground}{\textcolor{redfont}{\_}} \colorbox{redbackground}{\textcolor{redfont}{4}} \colorbox{redbackground}{\textcolor{redfont}{3}} \colorbox{redbackground}{\textcolor{redfont}{7}} \colorbox{redbackground}{\textcolor{redfont}{3}} \colorbox{redbackground}{\textcolor{redfont}{1}} \colorbox{redbackground}{\textcolor{redfont}{,}} \colorbox{redbackground}{\textcolor{redfont}{\_t}} \colorbox{redbackground}{\textcolor{redfont}{cp}} \colorbox{redbackground}{\textcolor{redfont}{.}} \colorbox{redbackground}{\textcolor{redfont}{d}} \colorbox{redbackground}{\textcolor{redfont}{st}} \colorbox{redbackground}{\textcolor{redfont}{port}} \colorbox{redbackground}{\textcolor{redfont}{:}} \colorbox{redbackground}{\textcolor{redfont}{\_}} \colorbox{redbackground}{\textcolor{redfont}{4}} \colorbox{redbackground}{\textcolor{redfont}{4}} \colorbox{redbackground}{\textcolor{redfont}{3}} \colorbox{redbackground}{\textcolor{redfont}{,}} \colorbox{redbackground}{\textcolor{redfont}{\_t}} ... \textsf{[Token length : 1405]}\\
\midrule
\textbf{\textsf{TrafficLLM} tokenization:} \colorbox{greenbackground}{\textcolor{greenfont}{\_ip}} \colorbox{greenbackground}{\textcolor{greenfont}{.}} \colorbox{greenbackground}{\textcolor{greenfont}{proto}} \colorbox{greenbackground}{\textcolor{greenfont}{:}} \colorbox{greenbackground}{\textcolor{greenfont}{\_6,}} \colorbox{greenbackground}{\textcolor{greenfont}{\_ip}} \colorbox{greenbackground}{\textcolor{greenfont}{.}} \colorbox{greenbackground}{\textcolor{greenfont}{checksum}} \colorbox{greenbackground}{\textcolor{greenfont}{:}} \colorbox{greenbackground}{\textcolor{greenfont}{\_0}} \colorbox{greenbackground}{\textcolor{greenfont}{x}} \colorbox{greenbackground}{\textcolor{greenfont}{000017}} \colorbox{greenbackground}{\textcolor{greenfont}{a}} \colorbox{greenbackground}{\textcolor{greenfont}{7,}} \colorbox{greenbackground}{\textcolor{greenfont}{\_ip}} \colorbox{greenbackground}{\textcolor{greenfont}{.}} \colorbox{greenbackground}{\textcolor{greenfont}{checksum}} \colorbox{greenbackground}{\textcolor{greenfont}{.}} \colorbox{greenbackground}{\textcolor{greenfont}{status}} \colorbox{greenbackground}{\textcolor{greenfont}{:}} \colorbox{greenbackground}{\textcolor{greenfont}{\_2,}} \colorbox{greenbackground}{\textcolor{greenfont}{\_ip}} \colorbox{greenbackground}{\textcolor{greenfont}{.}} \colorbox{greenbackground}{\textcolor{greenfont}{src}} \colorbox{greenbackground}{\textcolor{greenfont}{:}} \colorbox{greenbackground}{\textcolor{greenfont}{\_10.0.}} \colorbox{greenbackground}{\textcolor{greenfont}{2.15,}} \colorbox{greenbackground}{\textcolor{greenfont}{\_ip}} \colorbox{greenbackground}{\textcolor{greenfont}{.}} \colorbox{greenbackground}{\textcolor{greenfont}{dst}} \colorbox{greenbackground}{\textcolor{greenfont}{:}} \colorbox{greenbackground}{\textcolor{greenfont}{\_198.52.}} \colorbox{greenbackground}{\textcolor{greenfont}{200.39,}} \colorbox{greenbackground}{\textcolor{greenfont}{\_tcp}} \colorbox{greenbackground}{\textcolor{greenfont}{.}} \colorbox{greenbackground}{\textcolor{greenfont}{srcport}} \colorbox{greenbackground}{\textcolor{greenfont}{:}} \colorbox{greenbackground}{\textcolor{greenfont}{\_43731,}} \colorbox{greenbackground}{\textcolor{greenfont}{dstport}} \colorbox{greenbackground}{\textcolor{greenfont}{:}} \colorbox{greenbackground}{\textcolor{greenfont}{\_443,}} ... \textsf{[Token length : 690]} \\
\bottomrule
\end{tabular}
}
\vspace{-0.3cm}
\label{tab:tokenizer}
\end{center}
\end{table}

\subsection{Dual-Stage Tuning Pipeline} \label{sec3:instruction-tuning} 
\textsf{TrafficLLM} proposes a dual-stage tuning pipeline to help LLMs acquire domain knowledge to achieve generic representation learning on diverse traffic analysis tasks. The pipeline can help LLMs obtain two abilities in different stages: (i) understanding the task-related natural language to determine which task should be conducted and (ii) learning the task-specific traffic pattern across different tasks. Guided by expert instructions, \textsf{TrafficLLM} autonomously extracts task-specific traffic patterns from encoded inputs, building generic representations across different tasks.

\vspace{0.1cm}
\noindent \textbf{Tuning Objectives.} \textsf{TrafficLLM} aims to leverage LLMs' pattern mining and generalization abilities to learn generic traffic representations. Based on LLM's strong memorization coming from the deep Transformer architecture, these representations can obtain distinct traffic patterns learned from the meta-information (e.g., lengths, directions, and flags). \textsf{TrafficLLM} automatically integrates these heterogeneous data and finds their importance for different tasks (e.g., lengths for encrypted traffic classification (EAC)~\cite{liu2019fs} and directions for website fingerprinting (WF)~\cite{sirinam2018deep}. \textsf{TrafficLLM} uses the generic representations to realize two mainstream tasks, i.e., \textit{traffic detection} and \textit{traffic generation}.


\begin{itemize}[nolistsep,leftmargin=*]
	\item \textbf{Traffic detection.} Given the security expert's instruction $S = \{s_1, s_2, ..., s_m\}$ that contains $m$ language tokens, the traffic data $X = \{x_0, x_1, ...,  x_n\}$ that contains $n$ traffic tokens to describe the traffic meta-information, \textit{traffic detection} requires the task-related instruction $S_i$ and traffic data $X_i$ (flows or packets) as \textsf{TrafficLLM}'s input $(S_i, X_i)$. Then, \textsf{TrafficLLM} can identify the ground truth label $y_i \in Y = \{y_0, y_1, ..., y_c\}$ of the traffic across different traffic detection tasks (e.g., MTD, WAD, and BND tasks) with its parameter $\theta$:
\begin{equation}
y_i = \textsf{TrafficLLM}((S_i, X_i)|\theta)
\end{equation}
	\item \textbf{Traffic generation.} \textit{Traffic generation} can be regarded as the reversed process of traffic detection tasks. It aims to input the generation instruction $S_i$ to describe the specific scenario and the traffic category $y_i \in Y = \{y_0, y_1, ..., y_c\}$ of the traffic to be generated. \textsf{TrafficLLM} can generate a synthetic packet $\hat{X}_i$ that satisfies the instruction:
\begin{equation}
\hat{X}_i = \textsf{TrafficLLM}((S_i, y_i)|\theta)
\end{equation}
\end{itemize}

\begin{figure}[t]
\centering
\includegraphics[width=0.48\textwidth]{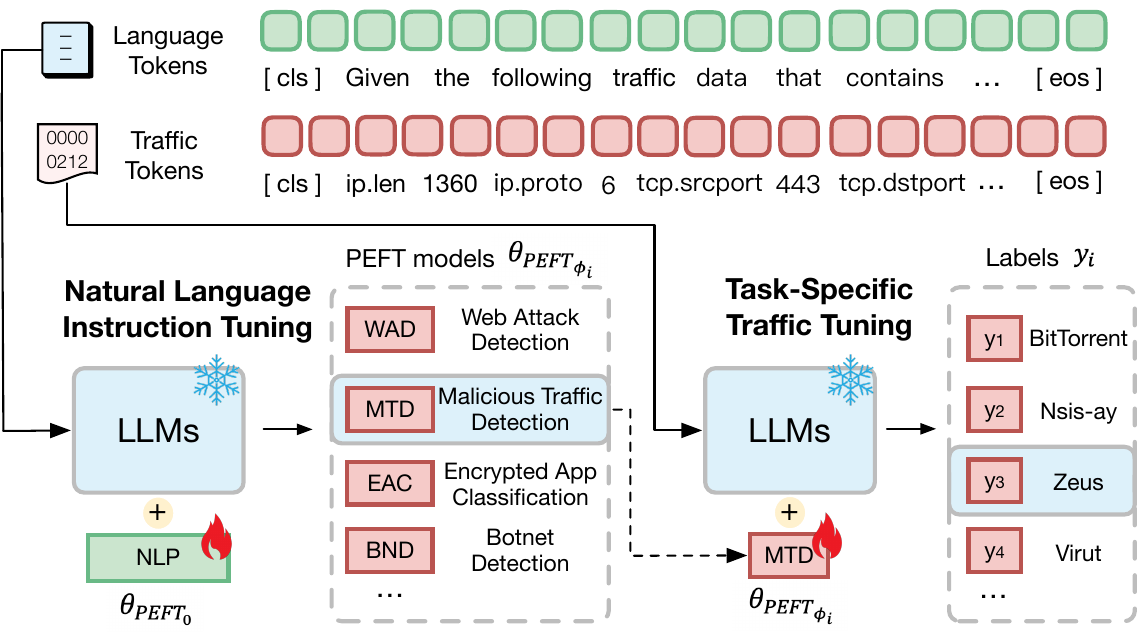}
\setlength{\abovecaptionskip}{-0.1cm}
\caption{Illustration of the dual-stage tuning pipeline to learn natural language and traffic patterns respectively.}
\label{fig4}
\vspace{-0.4cm}
\end{figure}

\vspace{0.1cm}
\noindent \textbf{Natural Language Instruction Tuning.} In the first stage of the dual-stage tuning in \textsf{TrafficLLM}, we introduce natural language instruction tuning to inject the professional task description text from the field of cybersecurity into LLMs. As shown in Figure~\ref{fig4}, the pipeline forces LLM to understand the instructions from security experts and predict the task name $\phi_i$ that needs to be performed.

\begin{equation}
\phi_i = \textsf{TrafficLLM}_{stage_1}(S_i|\theta_1)
\end{equation}
where $\theta_1$ is the trainable parameters in the first stage. $\phi_i$ is the downstream task name to be performed. To learn the context of the security task description, we follow LLM's autoregressive objective function to converge the model. Given the human instruction $S_i = \{s_1, s_2, ..., s_{m}\}$, \textsf{TrafficLLM} calculates the probability $P_m$ of token $s_i$ to model the loss $\mathcal{J}(\theta_1)$:
\begin{equation}
\begin{gathered}
P_m(s_i|s_1, ..., s_{i-1}) = {\rm softmax}(W_sh_{i-1})\\
\mathcal{J}(\theta_1) = \sum \sum_i {\rm log} P_m(s_i|s_1, ..., s_{i-1})
\end{gathered}
\end{equation}
where $W_s$ is the learning parameter matrix for task understanding. $h_{i-1}$ denotes the representation encoded in \textsf{TrafficLLM} with the input of the preceding $i-1$ tokens. The natural language instruction tuning technique plays a crucial role in accurately matching instruction text with the corresponding downstream task, thereby enabling LLMs' domain knowledge to understand different tasks.

\begin{figure}[t]
\centering
\includegraphics[width=0.46\textwidth]{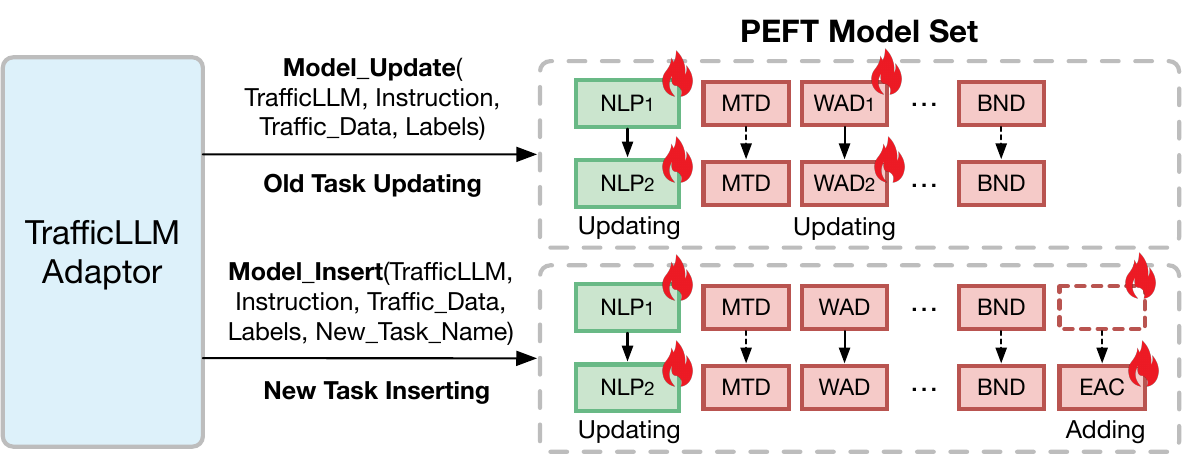}
\caption{The workflow of the extensible adaptation with parameter-effective fine-tuning (EA-PEFT) in \textsf{TrafficLLM}. }
\label{fig5}
\vspace{-0.4cm}
\end{figure}

\vspace{0.1cm}
\noindent \textbf{Task-Specific Traffic Tuning.} The second stage we propose is task-specific traffic tuning. After understanding the task, we force \textsf{TrafficLLM} to learn the traffic pattern under the downstream tasks. In this stage, we fine-tune LLM using the training pairs $(X_i, y_i)$ to model the traffic representation under traffic detection tasks $\phi_{TD}$ and traffic generation tasks $\phi_{TG}$. For the specific downstream task $\phi_i$, \textsf{TrafficLLM} trains the second stage parameters $\theta_2$ to predict traffic labels $y_i$ or generate synthetic traffic $\hat{X}_i$:

\begin{equation}
\begin{gathered}
y_i = \textsf{TrafficLLM}_{stage2}(X_i|(\theta_2;\phi_i \in \phi_{TD}))\\
\hat{X}_i = \textsf{TrafficLLM}_{stage2}(y_i|(\theta_2;\phi_i \in \phi_{TG}))
\end{gathered}
\end{equation}
To inject the knowledge of heterogeneous traffic data into LLMs, \textsf{TrafficLLM} builds and updates the representation of traffic data $X_i = \{x_1, x_2, ..., x_n\}$, by learning the context of traffic packets and flows with the loss function $\mathcal{J}(\theta_2)$:
\begin{equation}
\begin{gathered}
P_n(x_i|x_1, ..., x_{i-1}) = {\rm softmax}(W_lh_{i-1})\\
\mathcal{J}(\theta_2) = \sum \sum_i {\rm log} P_n(x_i|x_1, ..., x_{i-1})
\end{gathered}
\end{equation}
where $W_l$ is the trainable parameter matrix for traffic representation learning. The task-specific traffic tuning aims to align the LLMs with various traffic data under different scenarios, such as VPN and Tor networks, which allows LLMs to accomplish diverse downstream tasks with these traffic representations.

As shown in Figure~\ref{fig1} (right), the dual-stage tuning pipeline helps \textsf{TrafficLLM} achieve 95.0\% of average accuracy across MTD, EAC, and WAD tasks. \textsf{TrafficLLM} has an 84.8\% higher accuracy than directly fine-tuning due to learning text semantics and task-specific traffic patterns in different stages.

\subsection{Extensible Adaptation with PEFT} \label{sec3:applications}
To realize LLM's generalization to new traffic environments, \textsf{TrafficLLM} employs extensible adaptation with parameter-effective fine-tuning (EA-PEFT) to efficiently update representations on dynamic scenarios. The principle of EA-PEFT is to split traffic representation abilities of different tasks into various additional parameters, supporting for \textsf{TrafficLLM} to selectively update parts of capabilities to enable rapid updates of representations in new environments. 

\vspace{0.1cm}
\noindent \textbf{Traffic Domain Adaptation with PEFT.} Assume that the pre-trained LLM's parameters are $\theta_{LLM}$, to adapt to new environments, traditional retraining methods require training the full parameters of models, which indicates the parameter update $\Delta\theta$ is equal to $\theta_{LLM}$, i.e., $|\Delta\theta| = |\theta_{LLM}|$. However, the large parameter size of LLM entails huge costs of repeated retraining in new environments. To address this limitation, \textsf{TrafficLLM} freezes the parameters of the LLM and tunes extra parameters to achieve parameter effective fine-tuning (PEFT)\cite{ding2023parameter}. During the dual-stage tuning, \textsf{TrafficLLM} tunes additional parameters to respectively build $\theta_{PEFT_0}$ for task understanding and $\theta_{PEFT_{\phi_i}}$ for task-specific traffic learning:

\begin{equation}
\begin{gathered}
\theta_1 = \theta_{LLM} + \theta_{PEFT_0}\\
\theta_2 = \theta_{LLM} + \theta_{PEFT_{\phi_i}} 
\end{gathered}
\end{equation}
where $\theta_{PEFT_0}$ and $\theta_{PEFT_{\phi_i}}$ are the parameter updates $\Delta\theta$ in the two stages. This strategy helps \textsf{TrafficLLM} encapsulate abilities of natural language processing and traffic pattern learning across different tasks into specialized PEFT models, which are triggered by instructions from different tasks.

\vspace{0.1cm}
\noindent \textbf{Extensible Adaptation with PEFT Models.} Using the PEFT models trained during traffic domain adaptation, \textsf{TrafficLLM} employs EA-PEFT to organize these models with an extensible adaptation, which can help \textsf{TrafficLLM} easily adapt to new environments. Figure~\ref{fig5} shows the overview of EA-PEFT's workflow implemented by Python scripts. In EA-PEFT, \textsf{TrafficLLM} adaptor allows flexible operations to update old models or register new tasks. For instance, When faced with the traffic update in EAC and WAD tasks raised by client version upgrade (e.g., App version drift) or attack method changes (e.g., HTTP request body changes), the adaptor can call \verb|Model_Update| to update the specific PEFT models by providing new EAC or WAD datasets. Moreover, \textsf{TrafficLLM} can easily add new traffic analysis scenarios. The adaptor can schedule \verb|Model_Insert| to train new PEFT models and insert them in the EA-PEFT framework. Based on these, \textsf{TrafficLLM} easily scales to a wide range of traffic domain tasks with the light-wise adaption scheme of EA-PEFT.

As shown in Figure~\ref{fig2}, the EA-PEFT technique helps \textsf{TrafficLLM} only need to train 0.62\% of parameters when adapting to new traffic environments, which significantly reduces the adaptation costs with the reduction of 69.9\% GPU memory and 88.8\% training time. \textsf{TrafficLLM} effectively mitigates the high adaptation cost of retraining methods, facilitating deploy more traffic analysis tasks in real-world settings.

\section{Experiment Setup \& Implementation}\label{sec:implementation}

In this section, we introduce the experimental setup and implementation of \textsf{TrafficLLM}, including the implementation, datasets, baselines, and metrics used in the experiments.

\subsection{Implementation of \textsf{TrafficLLM}}
\noindent \textbf{Testbed.} We conduct our experiments on a Super GPU server (super-SYS-420GP-TNR) with 5 NVIDIA A100 80G GPUs, Ubuntu 18.04.1 (Linux 5.4.0), and 1TB memory. We use PyTorch 2.0.1 to build the prototype \textsf{TrafficLLM} and deploy Python scripts to incorporate \textsf{TrafficLLM} adaptor and PEFT models in the EA-PEFT framework. We employ Llama2-7B~\cite{llama2} and ChatGLM2-6B~\cite{chatglm} as the base LLMs for most experiments. We choose P-Tuning~v2~\cite{liu2021p} as the PEFT method. The storage of each PEFT model is 7.1MB.

\vspace{0.1cm}
\noindent \textbf{Hyper-parameters.} During the data pre-processing, we employ a data sampling process for each class to avoid the data imbalance issue. The maximum number of flows in each class is 5,000. We set the ratio of training sets, validation sets, and test sets to 8:1:1. In traffic detection tasks, we followed \cite{lin2022bert} to mask the Ethernet layer, IP addresses, and ports to avoid the bias derived by the sensitive meta-information. During the training stage, we set the training steps as 20,000; the initial learning rate is $2\times10^{-2}$. The maximum source and target lengths of generation tasks are set as 128 and 3,072, while detection tasks are 3,072 and 32. During the inference stage, we set top-p and temperature to 0.70 and 0.90 in traffic generation, while 0.90 and 0.10 in traffic detection.

\subsection{Datasets \& Tasks}
To comprehensively evaluate the effectiveness of \textsf{TrafficLLM}, we collect a wide set of traffic datasets and natural language instructions for LLM adaptation and experiments.

\vspace{0.1cm}
\noindent \textbf{Traffic Datasets.} The traffic datasets used in our experiments are shown in Table~\ref{tab:implementation-traffic-dataset}. We use 10 traffic datasets with different scenarios to collect $\approx$ 0.4M training data and build generalization abilities:

\begin{itemize}[nolistsep,leftmargin=*]
\item \textbf{Generalization across different tasks.} \textit{(i) Traffic detection:} To evaluate the detection performance of \textsf{TrafficLLM} on various network scenarios, we choose 8 traffic datasets to measure \textsf{TrafficLLM}'s abilities to detect \textit{malicious and benign traffic}. In \textit{malicious traffic detection tasks}, we introduce malware traffic detection (USTC TFC 2016~\cite{wang2017malware}), botnet detection (ISCX Botnet 2014~\cite{beigi2014towards}), malicious DoH detection (CIC DoHBrw 2020~\cite{montazerishatoori2020detection}), and web attack detection (CSIC 2010~\cite{csic2010}) tasks. In \textit{fine-grained benign traffic detection}, we employ encrypted VPN detection (ISCX VPN 2016~\cite{gil2016characterization}), Tor behavior detection (ISCX Tor 2016~\cite{lashkari2017characterization}), encrypted App classification (CSTNET 2023~\cite{lin2022bert}), and website fingerprinting (CW-100 2024~\cite{ZhaoD0LL0024}). We use traffic detection instructions, flows/packets in the traffic datasets, and the corresponding labels to build the human instructions $S_i$,  traffic data $X_i$, and the target label $y_i$. \textit{(ii) Traffic generation:} To implement the traffic generation capability, we reuse the traffic datasets mentioned above. We use the traffic label $y_i$ to produce the generation task instructions $S_i$ and sample the packets to form the synthetic packets $\hat{X_i}$.

\item \textbf{Generalization to unseen data.} To measure \textsf{TrafficLLM}'s generalization ability on unseen traffic data, we set up the concept drift~\cite{jiang2023zero} and APT attack detection~\cite{myneni2020dapt} scenarios, which are of great concern in the community. We selected APP-53 2023~\cite{jiang2023zero} and DAPT 2020~\cite{myneni2020dapt}, two representative open-sourced datasets containing historical and future-stage traffic for evaluation.

\end{itemize}
We choose these datasets to cover various applications (i.e., mobile apps, websites, and malware), protocols (i.e., HTTP, QUIC, TLS1.3, and DoH), network environments (i.e., VPN, Tor, and botnet), and attacks (i.e., web attacks and APT attacks) in the traffic. This ensures the variety to evaluate model robustness across different scenarios. 

\begin{table*}[t!]
\caption{The detail of 10 traffic datasets used to build the traffic analysis downstream tasks in experiments. \textsf{TrafficLLM} uses the traffic data and labels of these datasets to build traffic detection and generation capabilities respectively.}
\vspace{-0.5cm}
\begin{center}
\resizebox{\textwidth}{!}{
\begin{tabular}{c|c|c|c|c|c|c}
\toprule
\textbf{Tasks} & \textbf{Abbrev.} & \textbf{Traffic Datasets} & \textbf{Description} & \textbf{\#Flows} & \textbf{\#Packets} & \textbf{\#Labels} \\
\midrule
Malware Traffic Detection & MTD & USTC TFC 2016~\cite{wang2017malware} & 10-class malware and 10-class benign Apps & 9,853 & 97,115 & 20\\
Botnet Detection & BND & ISCX Botnet 2014~\cite{beigi2014towards} & 4-class botnets and 1-class benign network& 30,511 & 300,000 & 5\\
Malicious DoH Detection & MDD & CIC DoHBrw 2020~\cite{montazerishatoori2020detection} & 4-class benign DoH and 1-class malicious DoH & 545,463 & 28,341,000 & 5\\
Web Attack Detection & WAD & CSIC 2010~\cite{csic2010} & 1-class Web attack requests and 1-class benign requests & 61,000 & 61,000 & 2\\
APT Attack Detection & AAD & DAPT 2020~\cite{myneni2020dapt} & 1-class APT attack traffic and 1-class benign traffic & 3,000 & 10,000 & 2\\
\midrule
Encrypted VPN Detection & EVD & ISCX VPN 2016~\cite{gil2016characterization} & 19-class VPN encrypted App traffic & 3,694 & 60,000 & 14\\
Tor Behavior Detection & TBD & ISCX Tor 2016~\cite{lashkari2017characterization} & 8-class user behaviors under Tor network & 3,021 & 80,000 & 8\\
Encrypted App Classification & EAC & CSTNET 2023~\cite{lin2022bert} & 20-class mobile App traffic using TLS encryption & 65,128 & 602,568 & 20\\
Website Fingerprinting & WF & CW-100 2024~\cite{ZhaoD0LL0024} & 100-class website accessing traffic under Tor& 9,000 & 603,072 & 100\\
Concept Drift & CD & APP-53 2023~\cite{jiang2023zero} & 53-class mobile App traffic with concept drift & 133,000 & 449,000 & 53 \\
\bottomrule
\end{tabular}
}
\label{tab:implementation-traffic-dataset}
\end{center}
\vspace{-0.3cm}
\end{table*}

\begin{table*}[t!]
\caption{The statistic information and the top words of the natural language instructions we collect for task understanding.}
\vspace{-0.5cm}
\begin{center}
\resizebox{\textwidth}{!}{
\begin{tabular}{ll|ll||ll|ll|ll}
\toprule
\textbf{Statistics} & \textbf{Value} & \textbf{Statistics} & \textbf{Value} & \textbf{Word} & \%\textbf{Hits} & \textbf{Word} & \textbf{\%Hits} & \textbf{Word} & \textbf{\%Hits}\\
\midrule
Total words& 128,248 & Average number of words per instruction & 15.26 & traffic& 4.15\%& packet& 1.01\% & software & 0.48\%\\
Total unique words & 1,999 & Average number of unique words per instruction & 13.92& network & 2.58\%& application& 0.79\% & tunnel & 0.44\%\\
Total sentences& 15,238& Average number of sentence per instruction & 1.65 & data& 1.60\%& IP& 0.56\% & behavior & 0.35\%\\
Total instructions& 9,209 & Type Token Ratio (TTR) & 1.56 & field& 1.54\%& botnet & 0.49\% & set & 0.33\%\\
\bottomrule
\end{tabular}
}
\label{tab4}
\end{center}
\vspace{-0.3cm}
\end{table*}

\vspace{0.1cm}
\noindent \textbf{Natural Language Instructions.} We show the details of the natural language dataset in Table~\ref{tab4}. To build the natural language corpus as the human instructions in \textsf{TrafficLLM}, we invite security experts and college students to provide accurate task descriptions for each downstream task. Moreover, to increase the diversity of the context, we use ChatGPT~\cite{ChatGPT} to rewrite these expert instructions through prompt engineering and remove similar instructions based on human annotation. Each instruction is rewritten 20 times at least. Finally, we collect $\approx$ 10K text instructions to build the training data.

\vspace{-0.1cm}
\subsection{Baselines \& Evaluation Metrics}
\noindent \textbf{Baselines.} To compare the performance of \textsf{TrafficLLM}, we mainly use two types of baselines, including ML-based traffic detection and generation methods.
\begin{itemize}[nolistsep,leftmargin=*]
\item \textbf{ML-based detection methods.} We use state-of-the-art traffic detection methods across different tasks as baselines to evaluate the traffic detection abilities of \textsf{TrafficLLM}. The baselines include (i) Statistical Feature Methods: AppScanner~\cite{taylor2017robust}, CUMUL~\cite{panchenko2016website}, BIND~\cite{al2016adaptive}, k-fingerprinting (K-FP)~\cite{hayes2016k}, and FlowPrint~\cite{van2020flowprint}; (ii) Deep Learning Methods: FS-Net~\cite{liu2019fs}, Deep Fingerprinting (DF)~\cite{sirinam2018deep}, GraphDApp~\cite{shen2021accurate}, TSCRNN~\cite{lin2021tscrnn}, and Deeppacket~\cite{lotfollahi2020deep}; (iii) Pre-training Methods: PERT~\cite{he2020pert} and ET-BERT~\cite{lin2022bert}. 

\item \textbf{ML-based generation methods.} To evaluate the performance of traffic generation, we compare \textsf{TrafficLLM} to state-of-the-art ML-based generation methods. The baselines include (i) The GAN-based IP header trace generation algorithm: Netshare~\cite{yin2022practical}; (ii) The conditional GANs-based augmentation method: PacketCGAN~\cite{cheng2019pac}; (iii) The CNN-GAN-based IP packet generator: PAC-GAN~\cite{wang2020packetcgan}. Note that the rule-based methods are not in the range since they can only simulate network characteristics and are difficult to generate fine-grained packet features with manual configuration (e.g., diverse meta-information in App traffic). 
\end{itemize}
\textbf{Evaluation Metrics.} We use the following metrics to evaluate \textsf{TrafficLLM}: (i) Precision (PR), (ii) Recall (RC), (iii) F1-score (F1), (iv) Accuracy (acc), (v) False Positive (FP), (vi) the macro-average area under ROC curve (Macro-AUC), and (vii) Jensen-Shannon Divergence (JSD). Note that lower JSD and FP denote better fidelity.

\begin{table*}[t!]
\caption{Traffic detection results on APP-53 2023, ISCX Tor 2016, ISCX VPN 2016, CSTNET 2023, and CW-100 2024.}
\vspace{-0.3cm}
\begin{center}
\resizebox{\textwidth}{!}{
\begin{tabular}{l|ccc|ccc|ccc|ccc|ccc}
\toprule
\multirow{2}{*}{\textbf{Method}} & \multicolumn{3}{|c|}{\textbf{ISCX Tor 2016}} & \multicolumn{3}{|c}{\textbf{ISCX VPN 2016}} & \multicolumn{3}{|c|}{\textbf{APP-53 2023}} & \multicolumn{3}{|c}{\textbf{CSTNET 2023}} & \multicolumn{3}{|c}{\textbf{CW-100 2024}} \\
\cmidrule{2-16}
 & \textbf{PR} &\textbf{RC} & \textbf{F1} & \textbf{PR} & \textbf{RC} & \textbf{F1} & \textbf{PR} & \textbf{RC} & \textbf{F1} & \textbf{PR} & \textbf{RC} & \textbf{F1} & \textbf{PR} & \textbf{RC} & \textbf{F1}\\
\midrule
AppScanner~\cite{taylor2017robust} & 0.7251 & 0.6512 & 0.6124\textcolor{red}{\hspace{-0.1em}\raisebox{-0.2ex}{\scriptsize$\blacktriangledown$}} & 0.7395 & 0.7125 & 0.7304 & 0.7035 & 0.6957 & 0.6980 & 0.6481 & 0.6420 & 0.6467 & 0.6780 & 0.6825 & 0.6802 \\
CUMUL~\cite{panchenko2016website} & 0.5672 & 0.5731\textcolor{red}{\hspace{-0.1em}\raisebox{-0.2ex}{\scriptsize$\blacktriangledown$}} & 0.5628 & 0.6322 & 0.6824 & 0.6570 & 0.5563 & 0.5467\textcolor{red}{\hspace{-0.1em}\raisebox{-0.2ex}{\scriptsize$\blacktriangledown$}} & 0.5480\textcolor{red}{\hspace{-0.1em}\raisebox{-0.2ex}{\scriptsize$\blacktriangledown$}} & 0.5373 & 0.5217\textcolor{red}{\hspace{-0.1em}\raisebox{-0.2ex}{\scriptsize$\blacktriangledown$}} & 0.5274\textcolor{red}{\hspace{-0.1em}\raisebox{-0.2ex}{\scriptsize$\blacktriangledown$}} & 0.5623 & 0.5715\textcolor{red}{\hspace{-0.1em}\raisebox{-0.2ex}{\scriptsize$\blacktriangledown$}} & 0.5697\textcolor{red}{\hspace{-0.1em}\raisebox{-0.2ex}{\scriptsize$\blacktriangledown$}}  \\
BIND~\cite{al2016adaptive} & 0.4569\textcolor{red}{\hspace{-0.1em}\raisebox{-0.2ex}{\scriptsize$\blacktriangledown$}} & 0.4385\textcolor{red}{\hspace{-0.1em}\raisebox{-0.2ex}{\scriptsize$\blacktriangledown$}} & 0.4469\textcolor{red}{\hspace{-0.1em}\raisebox{-0.2ex}{\scriptsize$\blacktriangledown$}} & 0.5067\textcolor{red}{\hspace{-0.1em}\raisebox{-0.2ex}{\scriptsize$\blacktriangledown$}} & 0.4975\textcolor{red}{\hspace{-0.1em}\raisebox{-0.2ex}{\scriptsize$\blacktriangledown$}} & 0.5008\textcolor{red}{\hspace{-0.1em}\raisebox{-0.2ex}{\scriptsize$\blacktriangledown$}} & 0.6566\textcolor{red}{\hspace{-0.1em}\raisebox{-0.2ex}{\scriptsize$\blacktriangledown$}} & 0.6456\textcolor{red}{\hspace{-0.1em}\raisebox{-0.2ex}{\scriptsize$\blacktriangledown$}} & 0.6502\textcolor{red}{\hspace{-0.1em}\raisebox{-0.2ex}{\scriptsize$\blacktriangledown$}} & 0.7712 & 0.7689 & 0.7691 & 0.7504 & 0.7489 & 0.7501\\
K-FP~\cite{hayes2016k} & 0.7035 & 0.6789 & 0.6951 & 0.6784 & 0.6967 & 0.6891  & 0.5660\textcolor{red}{\hspace{-0.1em}\raisebox{-0.2ex}{\scriptsize$\blacktriangledown$}} & 0.5260\textcolor{red}{\hspace{-0.1em}\raisebox{-0.2ex}{\scriptsize$\blacktriangledown$}} & 0.5295\textcolor{red}{\hspace{-0.1em}\raisebox{-0.2ex}{\scriptsize$\blacktriangledown$}} & 0.4172\textcolor{red}{\hspace{-0.1em}\raisebox{-0.2ex}{\scriptsize$\blacktriangledown$}} & 0.3981\textcolor{red}{\hspace{-0.1em}\raisebox{-0.2ex}{\scriptsize$\blacktriangledown$}} & 0.4012\textcolor{red}{\hspace{-0.1em}\raisebox{-0.2ex}{\scriptsize$\blacktriangledown$}} & 0.5101\textcolor{red}{\hspace{-0.1em}\raisebox{-0.2ex}{\scriptsize$\blacktriangledown$}} & 0.4995\textcolor{red}{\hspace{-0.1em}\raisebox{-0.2ex}{\scriptsize$\blacktriangledown$}} & 0.5010\textcolor{red}{\hspace{-0.1em}\raisebox{-0.2ex}{\scriptsize$\blacktriangledown$}} \\
FlowPrint~\cite{van2020flowprint} & 0.4201\textcolor{red}{\hspace{-0.1em}\raisebox{-0.2ex}{\scriptsize$\blacktriangledown$}} & 0.3789\textcolor{red}{\hspace{-0.1em}\raisebox{-0.2ex}{\scriptsize$\blacktriangledown$}} & 0.3901\textcolor{red}{\hspace{-0.1em}\raisebox{-0.2ex}{\scriptsize$\blacktriangledown$}} & 0.7084 & 0.6608 & 0.6888 & 0.4890\textcolor{red}{\hspace{-0.1em}\raisebox{-0.2ex}{\scriptsize$\blacktriangledown$}} & 0.5023\textcolor{red}{\hspace{-0.1em}\raisebox{-0.2ex}{\scriptsize$\blacktriangledown$}} & 0.4950\textcolor{red}{\hspace{-0.1em}\raisebox{-0.2ex}{\scriptsize$\blacktriangledown$}} & 0.2371\textcolor{red}{\hspace{-0.1em}\raisebox{-0.2ex}{\scriptsize$\blacktriangledown$}} & 0.2270\textcolor{red}{\hspace{-0.1em}\raisebox{-0.2ex}{\scriptsize$\blacktriangledown$}} & 0.2254\textcolor{red}{\hspace{-0.1em}\raisebox{-0.2ex}{\scriptsize$\blacktriangledown$}} & 0.5237\textcolor{red}{\hspace{-0.1em}\raisebox{-0.2ex}{\scriptsize$\blacktriangledown$}} & 0.5227\textcolor{red}{\hspace{-0.1em}\raisebox{-0.2ex}{\scriptsize$\blacktriangledown$}} & 0.5225\textcolor{red}{\hspace{-0.1em}\raisebox{-0.2ex}{\scriptsize$\blacktriangledown$}} \\
\midrule
GraphDApp~\cite{shen2021accurate} &0.4789\textcolor{red}{\hspace{-0.1em}\raisebox{-0.2ex}{\scriptsize$\blacktriangledown$}} & 0.4878\textcolor{red}{\hspace{-0.1em}\raisebox{-0.2ex}{\scriptsize$\blacktriangledown$}} & 0.4781\textcolor{red}{\hspace{-0.1em}\raisebox{-0.2ex}{\scriptsize$\blacktriangledown$}} & 0.6478 & 0.6488 & 0.6476  & 0.6860 & 0.6450 & 0.6550 & 0.6329 & 0.5965 & 0.6078 & 0.6530 & 0.6974 & 0.6870\\
FS-Net~\cite{liu2019fs} & 0.6283\textcolor{red}{\hspace{-0.1em}\raisebox{-0.2ex}{\scriptsize$\blacktriangledown$}} & 0.6274\textcolor{red}{\hspace{-0.1em}\raisebox{-0.2ex}{\scriptsize$\blacktriangledown$}} & 0.5916\textcolor{red}{\hspace{-0.1em}\raisebox{-0.2ex}{\scriptsize$\blacktriangledown$}} & 0.7693 & 0.7488 & 0.7507 & 0.8550 & 0.8349 & 0.8376 & 0.8291 & 0.8061 & 0.8195 & 0.4582\textcolor{red}{\hspace{-0.1em}\raisebox{-0.2ex}{\scriptsize$\blacktriangledown$}} & 0.4781\textcolor{red}{\hspace{-0.1em}\raisebox{-0.2ex}{\scriptsize$\blacktriangledown$}} & 0.4668\textcolor{red}{\hspace{-0.1em}\raisebox{-0.2ex}{\scriptsize$\blacktriangledown$}}\\
DF~\cite{sirinam2018deep}& 0.6072\textcolor{red}{\hspace{-0.1em}\raisebox{-0.2ex}{\scriptsize$\blacktriangledown$}} & 0.6123\textcolor{red}{\hspace{-0.1em}\raisebox{-0.2ex}{\scriptsize$\blacktriangledown$}} & 0.6090\textcolor{red}{\hspace{-0.1em}\raisebox{-0.2ex}{\scriptsize$\blacktriangledown$}} & 0.6296\textcolor{red}{\hspace{-0.1em}\raisebox{-0.2ex}{\scriptsize$\blacktriangledown$}} & 0.6051\textcolor{red}{\hspace{-0.1em}\raisebox{-0.2ex}{\scriptsize$\blacktriangledown$}} & 0.6139\textcolor{red}{\hspace{-0.1em}\raisebox{-0.2ex}{\scriptsize$\blacktriangledown$}} & 0.7689\textcolor{red}{\hspace{-0.1em}\raisebox{-0.2ex}{\scriptsize$\blacktriangledown$}} & 0.7523\textcolor{red}{\hspace{-0.1em}\raisebox{-0.2ex}{\scriptsize$\blacktriangledown$}} & 0.7604\textcolor{red}{\hspace{-0.1em}\raisebox{-0.2ex}{\scriptsize$\blacktriangledown$}} & 0.7729\textcolor{red}{\hspace{-0.1em}\raisebox{-0.2ex}{\scriptsize$\blacktriangledown$}} & 0.7621\textcolor{red}{\hspace{-0.1em}\raisebox{-0.2ex}{\scriptsize$\blacktriangledown$}} & 0.7682\textcolor{red}{\hspace{-0.1em}\raisebox{-0.2ex}{\scriptsize$\blacktriangledown$}} & 0.9120 & 0.9046 & 0.9075\\
TSCRNN~\cite{lin2021tscrnn} & 0.9051 & 0.9178 & 0.9105 & 0.9346 & 0.9367 & 0.9349  & 0.7057\textcolor{red}{\hspace{-0.1em}\raisebox{-0.2ex}{\scriptsize$\blacktriangledown$}} & 0.6890\textcolor{red}{\hspace{-0.1em}\raisebox{-0.2ex}{\scriptsize$\blacktriangledown$}} & 0.6995\textcolor{red}{\hspace{-0.1em}\raisebox{-0.2ex}{\scriptsize$\blacktriangledown$}} & 0.7529\textcolor{red}{\hspace{-0.1em}\raisebox{-0.2ex}{\scriptsize$\blacktriangledown$}} & 0.7566\textcolor{red}{\hspace{-0.1em}\raisebox{-0.2ex}{\scriptsize$\blacktriangledown$}} & 0.7558\textcolor{red}{\hspace{-0.1em}\raisebox{-0.2ex}{\scriptsize$\blacktriangledown$}} & 0.8350 & 0.8210\textcolor{red}{\hspace{-0.1em}\raisebox{-0.2ex}{\scriptsize$\blacktriangledown$}} & 0.8260\\
Deeppacket~\cite{lotfollahi2020deep} & 0.7456\textcolor{red}{\hspace{-0.1em}\raisebox{-0.2ex}{\scriptsize$\blacktriangledown$}} & 0.7469\textcolor{red}{\hspace{-0.1em}\raisebox{-0.2ex}{\scriptsize$\blacktriangledown$}} & 0.7400\textcolor{red}{\hspace{-0.1em}\raisebox{-0.2ex}{\scriptsize$\blacktriangledown$}} & 0.9467 & 0.9508 & 0.9503 & 0.5590\textcolor{red}{\hspace{-0.1em}\raisebox{-0.2ex}{\scriptsize$\blacktriangledown$}} & 0.5489\textcolor{red}{\hspace{-0.1em}\raisebox{-0.2ex}{\scriptsize$\blacktriangledown$}} & 0.5506\textcolor{red}{\hspace{-0.1em}\raisebox{-0.2ex}{\scriptsize$\blacktriangledown$}} & 0.4013\textcolor{red}{\hspace{-0.1em}\raisebox{-0.2ex}{\scriptsize$\blacktriangledown$}} & 0.2965\textcolor{red}{\hspace{-0.1em}\raisebox{-0.2ex}{\scriptsize$\blacktriangledown$}} & 0.3890\textcolor{red}{\hspace{-0.1em}\raisebox{-0.2ex}{\scriptsize$\blacktriangledown$}} & 0.8243\textcolor{red}{\hspace{-0.1em}\raisebox{-0.2ex}{\scriptsize$\blacktriangledown$}} & 0.8246\textcolor{red}{\hspace{-0.1em}\raisebox{-0.2ex}{\scriptsize$\blacktriangledown$}} & 0.8244\textcolor{red}{\hspace{-0.1em}\raisebox{-0.2ex}{\scriptsize$\blacktriangledown$}}\\
\midrule
PERT~\cite{he2020pert}& 0.7480\textcolor{red}{\hspace{-0.1em}\raisebox{-0.2ex}{\scriptsize$\blacktriangledown$}} & 0.4952\textcolor{red}{\hspace{-0.1em}\raisebox{-0.2ex}{\scriptsize$\blacktriangledown$}} & 0.4874\textcolor{red}{\hspace{-0.1em}\raisebox{-0.2ex}{\scriptsize$\blacktriangledown$}} & 0.8573 & 0.7394\textcolor{red}{\hspace{-0.1em}\raisebox{-0.2ex}{\scriptsize$\blacktriangledown$}} & 0.7481\textcolor{red}{\hspace{-0.1em}\raisebox{-0.2ex}{\scriptsize$\blacktriangledown$}} & 0.8458 & 0.8369 & 0.8403 & 0.8896 & 0.8721 & 0.8771 & 0.8247 & 0.8355 & 0.8300 \\
ET-BERT~\cite{lin2022bert} & 0.9809 & 0.9830 & 0.9810 & 0.9890 & 0.9890 & 0.9890 & 0.8540\textcolor{red}{\hspace{-0.1em}\raisebox{-0.2ex}{\scriptsize$\blacktriangledown$}} & 0.8494\textcolor{red}{\hspace{-0.1em}\raisebox{-0.2ex}{\scriptsize$\blacktriangledown$}} & 0.8506\textcolor{red}{\hspace{-0.1em}\raisebox{-0.2ex}{\scriptsize$\blacktriangledown$}} & 0.9581 & 0.9478 & 0.9496 & 0.8670\textcolor{red}{\hspace{-0.1em}\raisebox{-0.2ex}{\scriptsize$\blacktriangledown$}} & 0.8650\textcolor{red}{\hspace{-0.1em}\raisebox{-0.2ex}{\scriptsize$\blacktriangledown$}} & 0.8660\textcolor{red}{\hspace{-0.1em}\raisebox{-0.2ex}{\scriptsize$\blacktriangledown$}}\\
\midrule
\textsf{TrafficLLM} & \textbf{0.9810} & \textbf{0.9871} & \textbf{0.9810} & \textbf{0.9960} & \textbf{0.9970} & \textbf{0.9960} & \textbf{0.9325} & \textbf{0.9315} & \textbf{0.9320} & \textbf{0.9678} & 0.9369 & \textbf{0.9599} & \textbf{0.9370} & \textbf{0.9360} & \textbf{0.9366} \\
\bottomrule
\multicolumn{16}{l}{$^1$ We use \textcolor{red}{\hspace{-0.1em}\raisebox{-0.2ex}{\scriptsize$\blacktriangledown$}} to mark the results of each baseline with a decrease of more than 10\% compared to their best results among 5 fine-grained benign traffic detection tasks.} \\
\end{tabular}
}
\label{tab:evaluation-traffic-classification-1}
\end{center}
\vspace{-0.3cm}
\end{table*}

\begin{table*}[t!]
\caption{Traffic detection results on ISCX Botnet 2014, USTC TFC 2016, DoHBrw 2020, CSIC 2010, and DAPT 2020.}
\vspace{-0.3cm}
\begin{center}
\resizebox{\textwidth}{!}{
\begin{tabular}{l|ccc|ccc|ccc|ccc|ccc}
\toprule
\multirow{2}{*}{\textbf{Method}} & \multicolumn{3}{|c|}{\textbf{ISCX Botnet 2014}} & \multicolumn{3}{|c|}{\textbf{USTC TFC 2016}} & \multicolumn{3}{|c|}{\textbf{CIC DoHBrw 2020}} & \multicolumn{3}{|c}{\textbf{DAPT 2020}} & \multicolumn{3}{|c}{\textbf{CSIC 2010}}\\
\cmidrule{2-16}
 & \textbf{PR} & \textbf{RC} & \textbf{F1} & \textbf{PR} & \textbf{RC} & \textbf{F1} & \textbf{PR} & \textbf{RC} & \textbf{F1} & \textbf{PR} & \textbf{RC} & \textbf{F1} & \textbf{PR} & \textbf{RC} & \textbf{F1}\\
\midrule
AppScanner~\cite{taylor2017robust} & 0.9021 & 0.9004 & 0.9008 & 0.8872 & 0.8910 & 0.8972 & 0.7331\textcolor{red}{\hspace{-0.1em}\raisebox{-0.2ex}{\scriptsize$\blacktriangledown$}} & 0.7063\textcolor{red}{\hspace{-0.1em}\raisebox{-0.2ex}{\scriptsize$\blacktriangledown$}} & 0.7106\textcolor{red}{\hspace{-0.1em}\raisebox{-0.2ex}{\scriptsize$\blacktriangledown$}} & 0.7590\textcolor{red}{\hspace{-0.1em}\raisebox{-0.2ex}{\scriptsize$\blacktriangledown$}} & 0.7226\textcolor{red}{\hspace{-0.1em}\raisebox{-0.2ex}{\scriptsize$\blacktriangledown$}} & 0.7408\textcolor{red}{\hspace{-0.1em}\raisebox{-0.2ex}{\scriptsize$\blacktriangledown$}} & 0.7465\textcolor{red}{\hspace{-0.1em}\raisebox{-0.2ex}{\scriptsize$\blacktriangledown$}} & 0.7112\textcolor{red}{\hspace{-0.1em}\raisebox{-0.2ex}{\scriptsize$\blacktriangledown$}} & 0.7220\textcolor{red}{\hspace{-0.1em}\raisebox{-0.2ex}{\scriptsize$\blacktriangledown$}} \\
CUMUL~\cite{panchenko2016website}& 0.8791 & 0.8320 & 0.8417 & 0.6074\textcolor{red}{\hspace{-0.1em}\raisebox{-0.2ex}{\scriptsize$\blacktriangledown$}} & 0.5239\textcolor{red}{\hspace{-0.1em}\raisebox{-0.2ex}{\scriptsize$\blacktriangledown$}} & 0.5437\textcolor{red}{\hspace{-0.1em}\raisebox{-0.2ex}{\scriptsize$\blacktriangledown$}} & 0.5623\textcolor{red}{\hspace{-0.1em}\raisebox{-0.2ex}{\scriptsize$\blacktriangledown$}} & 0.5281\textcolor{red}{\hspace{-0.1em}\raisebox{-0.2ex}{\scriptsize$\blacktriangledown$}} & 0.5301\textcolor{red}{\hspace{-0.1em}\raisebox{-0.2ex}{\scriptsize$\blacktriangledown$}} & 0.6509\textcolor{red}{\hspace{-0.1em}\raisebox{-0.2ex}{\scriptsize$\blacktriangledown$}} & 0.6486\textcolor{red}{\hspace{-0.1em}\raisebox{-0.2ex}{\scriptsize$\blacktriangledown$}} & 0.6492\textcolor{red}{\hspace{-0.1em}\raisebox{-0.2ex}{\scriptsize$\blacktriangledown$}} & 0.6905\textcolor{red}{\hspace{-0.1em}\raisebox{-0.2ex}{\scriptsize$\blacktriangledown$}} & 0.6870\textcolor{red}{\hspace{-0.1em}\raisebox{-0.2ex}{\scriptsize$\blacktriangledown$}} & 0.6894\textcolor{red}{\hspace{-0.1em}\raisebox{-0.2ex}{\scriptsize$\blacktriangledown$}} \\
BIND~\cite{al2016adaptive} & 0.6798\textcolor{red}{\hspace{-0.1em}\raisebox{-0.2ex}{\scriptsize$\blacktriangledown$}} & 0.6489\textcolor{red}{\hspace{-0.1em}\raisebox{-0.2ex}{\scriptsize$\blacktriangledown$}} & 0.6582\textcolor{red}{\hspace{-0.1em}\raisebox{-0.2ex}{\scriptsize$\blacktriangledown$}} & 0.8268 & 0.8014 &  0.8209 & 0.7137\textcolor{red}{\hspace{-0.1em}\raisebox{-0.2ex}{\scriptsize$\blacktriangledown$}} & 0.7003\textcolor{red}{\hspace{-0.1em}\raisebox{-0.2ex}{\scriptsize$\blacktriangledown$}} & 0.7077\textcolor{red}{\hspace{-0.1em}\raisebox{-0.2ex}{\scriptsize$\blacktriangledown$}} & 0.7224\textcolor{red}{\hspace{-0.1em}\raisebox{-0.2ex}{\scriptsize$\blacktriangledown$}} & 0.7026\textcolor{red}{\hspace{-0.1em}\raisebox{-0.2ex}{\scriptsize$\blacktriangledown$}} & 0.7115\textcolor{red}{\hspace{-0.1em}\raisebox{-0.2ex}{\scriptsize$\blacktriangledown$}} & 0.7022\textcolor{red}{\hspace{-0.1em}\raisebox{-0.2ex}{\scriptsize$\blacktriangledown$}} & 0.7002\textcolor{red}{\hspace{-0.1em}\raisebox{-0.2ex}{\scriptsize$\blacktriangledown$}} & 0.7010\textcolor{red}{\hspace{-0.1em}\raisebox{-0.2ex}{\scriptsize$\blacktriangledown$}} \\
K-FP~\cite{hayes2016k}& 0.8398 & 0.8960 & 0.8591 & 0.6447\textcolor{red}{\hspace{-0.1em}\raisebox{-0.2ex}{\scriptsize$\blacktriangledown$}} & 0.4172\textcolor{red}{\hspace{-0.1em}\raisebox{-0.2ex}{\scriptsize$\blacktriangledown$}} & 0.3981\textcolor{red}{\hspace{-0.1em}\raisebox{-0.2ex}{\scriptsize$\blacktriangledown$}} & 0.7035\textcolor{red}{\hspace{-0.1em}\raisebox{-0.2ex}{\scriptsize$\blacktriangledown$}} & 0.6789\textcolor{red}{\hspace{-0.1em}\raisebox{-0.2ex}{\scriptsize$\blacktriangledown$}} & 0.6951\textcolor{red}{\hspace{-0.1em}\raisebox{-0.2ex}{\scriptsize$\blacktriangledown$}} & 0.6585\textcolor{red}{\hspace{-0.1em}\raisebox{-0.2ex}{\scriptsize$\blacktriangledown$}} & 0.6496\textcolor{red}{\hspace{-0.1em}\raisebox{-0.2ex}{\scriptsize$\blacktriangledown$}} & 0.6542\textcolor{red}{\hspace{-0.1em}\raisebox{-0.2ex}{\scriptsize$\blacktriangledown$}} & 0.6855\textcolor{red}{\hspace{-0.1em}\raisebox{-0.2ex}{\scriptsize$\blacktriangledown$}} & 0.6960\textcolor{red}{\hspace{-0.1em}\raisebox{-0.2ex}{\scriptsize$\blacktriangledown$}} & 0.6920\textcolor{red}{\hspace{-0.1em}\raisebox{-0.2ex}{\scriptsize$\blacktriangledown$}} \\
FlowPrint~\cite{van2020flowprint}& 0.5898\textcolor{red}{\hspace{-0.1em}\raisebox{-0.2ex}{\scriptsize$\blacktriangledown$}} & 0.6309 & 0.5967 & 0.6609\textcolor{red}{\hspace{-0.1em}\raisebox{-0.2ex}{\scriptsize$\blacktriangledown$}} & 0.6596 & 0.6584 & 0.7712 & 0.2371\textcolor{red}{\hspace{-0.1em}\raisebox{-0.2ex}{\scriptsize$\blacktriangledown$}} & 0.2270\textcolor{red}{\hspace{-0.1em}\raisebox{-0.2ex}{\scriptsize$\blacktriangledown$}} & 0.6960 & 0.6894& 0.6935 & 0.7025 & 0.7009 & 0.7015\\
\midrule
GraphDApp~\cite{shen2021accurate}& 0.7578& 0.7598& 0.7538& 0.8027 & 0.8320 & 0.8263 & 0.6478\textcolor{red}{\hspace{-0.1em}\raisebox{-0.2ex}{\scriptsize$\blacktriangledown$}} & 0.6791\textcolor{red}{\hspace{-0.1em}\raisebox{-0.2ex}{\scriptsize$\blacktriangledown$}} & 0.6512\textcolor{red}{\hspace{-0.1em}\raisebox{-0.2ex}{\scriptsize$\blacktriangledown$}} & 0.8350 & 0.8342 & 0.8345& 0.8050 & 0.8240 & 0.8120 \\
FS-Net~\cite{liu2019fs} & 0.7303 & 0.8546 & 0.7876 & 0.5964\textcolor{red}{\hspace{-0.1em}\raisebox{-0.2ex}{\scriptsize$\blacktriangledown$}} & 0.7174\textcolor{red}{\hspace{-0.1em}\raisebox{-0.2ex}{\scriptsize$\blacktriangledown$}} & 0.6371\textcolor{red}{\hspace{-0.1em}\raisebox{-0.2ex}{\scriptsize$\blacktriangledown$}} & 0.7123 & 0.6991\textcolor{red}{\hspace{-0.1em}\raisebox{-0.2ex}{\scriptsize$\blacktriangledown$}} & 0.7053 & 0.8056 & 0.7783 & 0.7946 & 0.6255\textcolor{red}{\hspace{-0.1em}\raisebox{-0.2ex}{\scriptsize$\blacktriangledown$}} & 0.6145\textcolor{red}{\hspace{-0.1em}\raisebox{-0.2ex}{\scriptsize$\blacktriangledown$}} & 0.6190\textcolor{red}{\hspace{-0.1em}\raisebox{-0.2ex}{\scriptsize$\blacktriangledown$}}\\
DF~\cite{sirinam2018deep}& 0.8267 & 0.8509 & 0.7980 & 0.7623& 0.7598 & 0.7604 & 0.7078\textcolor{red}{\hspace{-0.1em}\raisebox{-0.2ex}{\scriptsize$\blacktriangledown$}} & 0.6986\textcolor{red}{\hspace{-0.1em}\raisebox{-0.2ex}{\scriptsize$\blacktriangledown$}} & 0.7022\textcolor{red}{\hspace{-0.1em}\raisebox{-0.2ex}{\scriptsize$\blacktriangledown$}} & 0.7892 & 0.7759 & 0.7805\textcolor{red}{\hspace{-0.1em}\raisebox{-0.2ex}{\scriptsize$\blacktriangledown$}} & 0.6470\textcolor{red}{\hspace{-0.1em}\raisebox{-0.2ex}{\scriptsize$\blacktriangledown$}} & 0.6455\textcolor{red}{\hspace{-0.1em}\raisebox{-0.2ex}{\scriptsize$\blacktriangledown$}} & 0.6464\textcolor{red}{\hspace{-0.1em}\raisebox{-0.2ex}{\scriptsize$\blacktriangledown$}}\\
TSCRNN~\cite{lin2021tscrnn}& 0.9206 & 0.8976& 0.8989& 0.9538 & 0.9428 & 0.9503 & 0.8837 & 0.8672 & 0.8695 & 0.8453\textcolor{red}{\hspace{-0.1em}\raisebox{-0.2ex}{\scriptsize$\blacktriangledown$}} & 0.8550 & 0.8503 & 0.6480\textcolor{red}{\hspace{-0.1em}\raisebox{-0.2ex}{\scriptsize$\blacktriangledown$}} & 0.6326\textcolor{red}{\hspace{-0.1em}\raisebox{-0.2ex}{\scriptsize$\blacktriangledown$}} & 0.6370\textcolor{red}{\hspace{-0.1em}\raisebox{-0.2ex}{\scriptsize$\blacktriangledown$}}\\
Deeppacket~\cite{lotfollahi2020deep}& 0.9408 & 0.9520 & 0.9496& 0.9369 & 0.9292& 0.9338 & 0.8930 & 0.8977 & 0.8965 & 0.8934 & 0.8924& 0.8930 & 0.6469\textcolor{red}{\hspace{-0.1em}\raisebox{-0.2ex}{\scriptsize$\blacktriangledown$}} & 0.6510\textcolor{red}{\hspace{-0.1em}\raisebox{-0.2ex}{\scriptsize$\blacktriangledown$}} & 0.6505\textcolor{red}{\hspace{-0.1em}\raisebox{-0.2ex}{\scriptsize$\blacktriangledown$}}\\
\midrule
PERT~\cite{he2020pert}& 0.9268 & 0.9078 & 0.9096 & 0.9605 & 0.9611 & 0.9574 & 0.9378& 0.9052 & 0.8977 &  0.8990 & 0.8868 & 0.8960 & 0.8274\textcolor{red}{\hspace{-0.1em}\raisebox{-0.2ex}{\scriptsize$\blacktriangledown$}} & 0.7588\textcolor{red}{\hspace{-0.1em}\raisebox{-0.2ex}{\scriptsize$\blacktriangledown$}} & 0.7685\textcolor{red}{\hspace{-0.1em}\raisebox{-0.2ex}{\scriptsize$\blacktriangledown$}}\\
ET-BERT~\cite{lin2022bert}& 0.9503 & 0.9462 & 0.9489 & 0.9930 & 0.9930 & 0.9930 & 0.8927\textcolor{red}{\hspace{-0.1em}\raisebox{-0.2ex}{\scriptsize$\blacktriangledown$}} & 0.8674\textcolor{red}{\hspace{-0.1em}\raisebox{-0.2ex}{\scriptsize$\blacktriangledown$}} & 0.8467\textcolor{red}{\hspace{-0.1em}\raisebox{-0.2ex}{\scriptsize$\blacktriangledown$}} & 0.9450 & 0.9423 & 0.9435 & 0.9021 & 0.8920\textcolor{red}{\hspace{-0.1em}\raisebox{-0.2ex}{\scriptsize$\blacktriangledown$}} & 0.8995 \\
\midrule
\textsf{TrafficLLM} & \textbf{0.9800} & \textbf{0.9861} & \textbf{0.9800} & \textbf{0.9950} & \textbf{0.9957} & \textbf{0.9950} & \textbf{0.9640}& \textbf{0.9640} & \textbf{0.9639} & \textbf{0.9820} & \textbf{0.9806} & \textbf{0.9810} & \textbf{0.9870} & \textbf{0.9823} & \textbf{0.9845}\\
\bottomrule
\multicolumn{16}{l}{$^1$ We use \textcolor{red}{\hspace{-0.1em}\raisebox{-0.2ex}{\scriptsize$\blacktriangledown$}} to mark the results of each baseline with a performance decrease of more than 10\% compared to their best results among 5 malicious traffic detection tasks.}
\end{tabular}
}
\label{tab:evaluation-traffic-classification-2}
\end{center}
\vspace{-0.3cm}
\end{table*}

\section{Evaluation}\label{sec:evaluation}

In this section, we evaluate \textsf{TrafficLLM}'s generalization abilities across different scenarios, including various detection and generation tasks, unseen scenarios, and real-world settings.

\subsection{Generalization across Detection Tasks}\label{sec:evaluation-traffic-classification}
We first evaluate whether \textsf{TrafficLLM} can reach robust detection performance across different scenarios and analyze the reason for its generalization ability on traffic detection.

\vspace{0.1cm}
\noindent \textbf{Generalization on Different Tasks.} Table~\ref{tab:evaluation-traffic-classification-1} and Table~\ref{tab:evaluation-traffic-classification-2} present the performance of \textsf{TrafficLLM} on 10 datasets for various traffic detection tasks. Results indicate that \textsf{TrafficLLM} can classify all 229 types of traffic with F1-score ranging from 0.9320 to 0.9960. \textsf{TrafficLLM} achieves at most 80.12\% better results than all baselines. Pre-training methods like PERT and ET-BERT obtained acceptable results with the average F1-scores of 0.8128 and 0.9324 since they additionally put the traffic bytes of pre-trained datasets into the model compared to prior works. Nevertheless, since \textsf{TrafficLLM} leverages LLM's pattern mining and generalization abilities, the performance outperforms PERT and ET-BERT with an improvement of 9.63\% at most on the F1-score metric. 


Furthermore, due to the difference between various detection scenarios, most works keep a poor generalization ability to share their models across different tasks (e.g., FlowPrint keeps F1-scores ranging from 0.2254 to 0.7015 with a variance of 3.396\%). Results indicate that previous ML-based methods usually show low generalization due to their dependence on handcrafted features and predefined model structures. For instance, Deeppacket gets an F1-score of 0.9503 on the EVD task (ISCX VPN 2016) but only obtains an F1-score of 0.3890 on the EAC task (CSTNET 2023). Conversely, \textsf{TrafficLLM} outperforms existing methods with an average F1-score of 0.9875 and a variance of 0.018\%, compared to the pre-trained model ET-BERT's 0.9324 average F1-score and 0.151\% variance.

\begin{figure}[t]
\vspace{-0.3cm}
\centering
\subfigure[\textsf{TrafficLLM} on Masked Features]{    
\includegraphics[width=4.1cm]{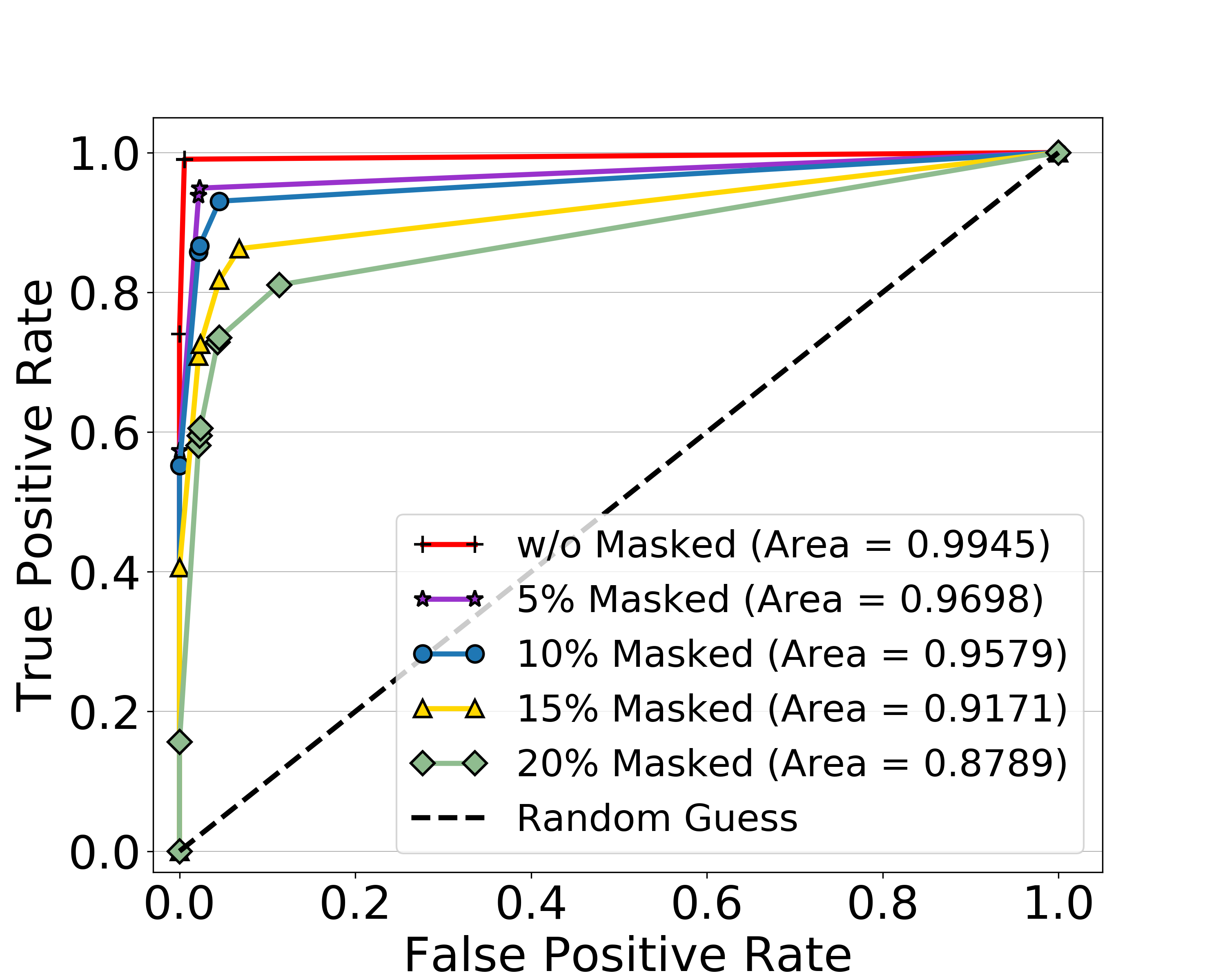}
}\label{fig6a}
\subfigure[Compared to ET-BERT and PERT]{ 
\includegraphics[width=4.1cm]{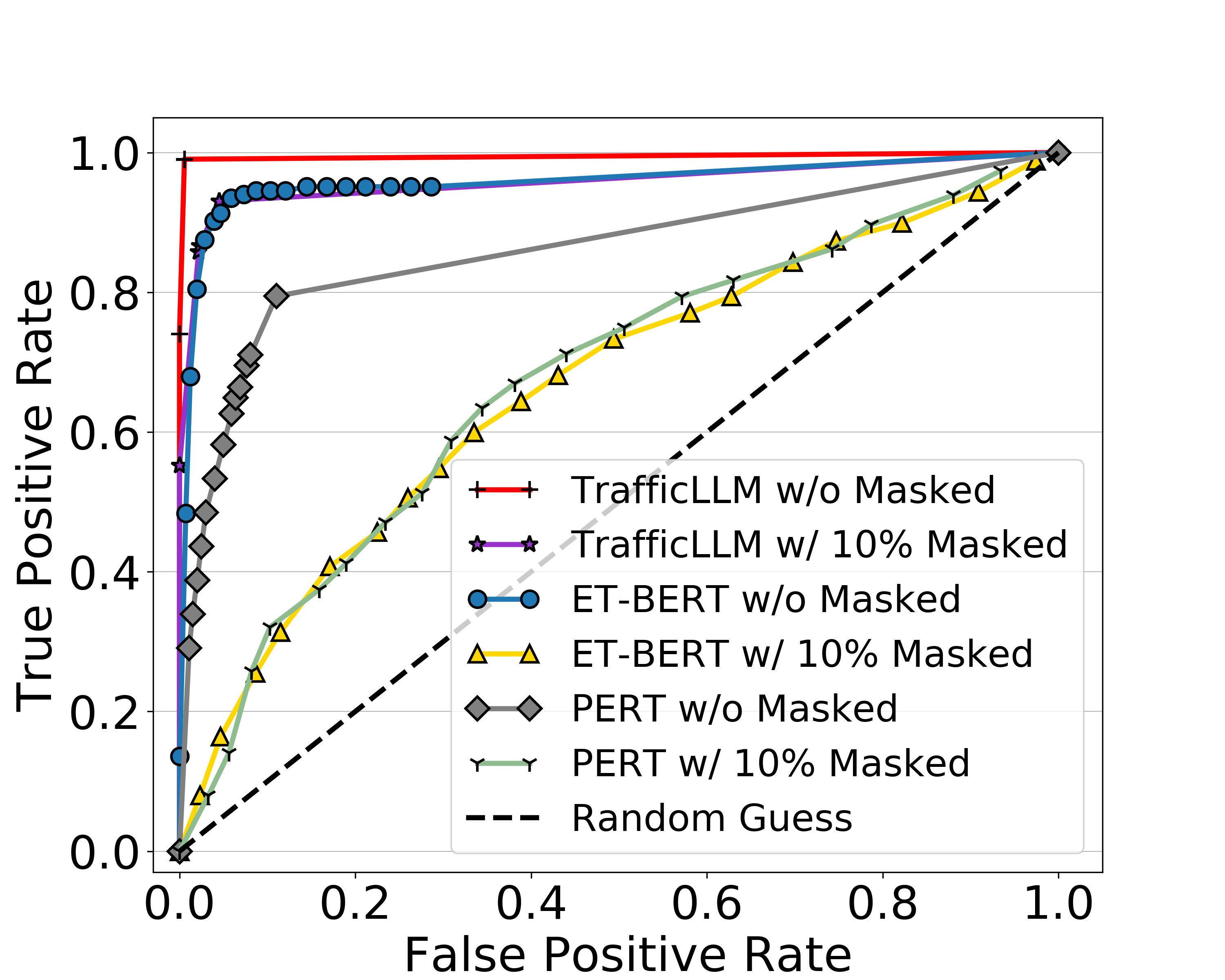}
}\label{fig6b}
\centering
\setlength{\abovecaptionskip}{-0.03cm}
\caption{The Macro-AUC of \textsf{TrafficLLM} and baselines with different ratios of masked features on ISCX Tor 2016.} 
\label{fig6}
\vspace{-0.3cm}
\end{figure}

\vspace{0.1cm}
\noindent \textbf{Robustness against masking features.}
To understand why \textsf{TrafficLLM} keeps such generalization ability on traffic detection tasks, we randomly mask partial packet meta-information in the inference stage to test detection performance. In Figure~\ref{fig6}, results indicate that \textsf{TrafficLLM} can obtain a Macro-AUC of 0.9171 even when 15\% of features missing. However, the performance of pre-trained models ET-BERT and PERT significantly declines. For instance, for a target FPR = $1\times10^{-1}$, while \textsf{TrafficLLM} achieves a TPR of 0.90, both two baselines provide TPRs less than 0.40. The robustness against missing features comes from the pattern reasoning and generalization abilities inherited from LLMs, while previous work does not. Using the raw traffic data as inputs, \textsf{TrafficLLM} does not heavily rely on the partial features. It benefits from the ability to automatically learn the importance and relationships of meta-information to the specific task to build generic traffic representation, helping \textsf{TrafficLLM} release strong generalization across different scenarios.



\begin{figure}[t]
\centering
\subfigure[5-Tuples JSD]{    
\includegraphics[width=4.0cm]{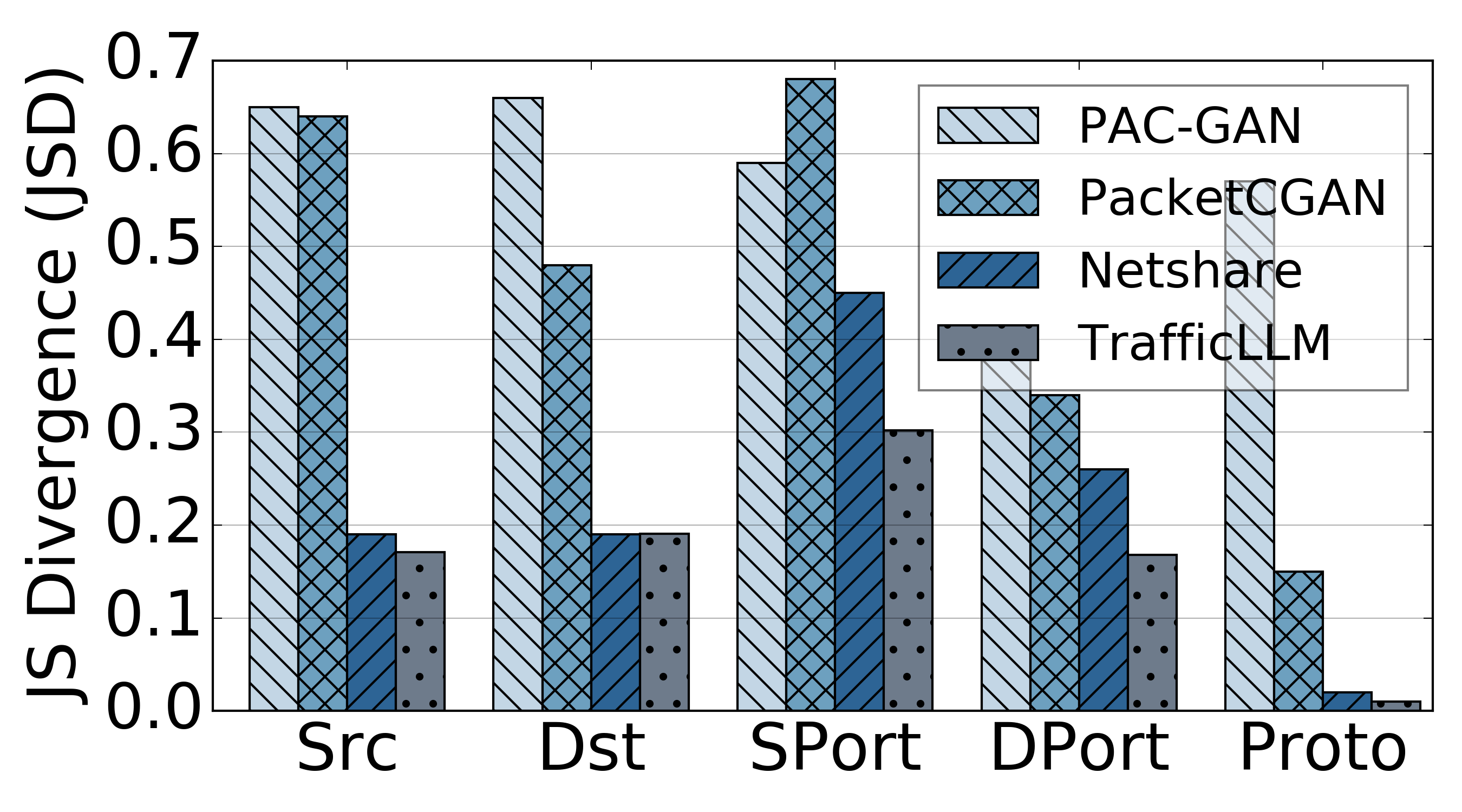}
}\label{fig7a}
\subfigure[Destination IP JSD]{ 
\includegraphics[width=4.0cm]{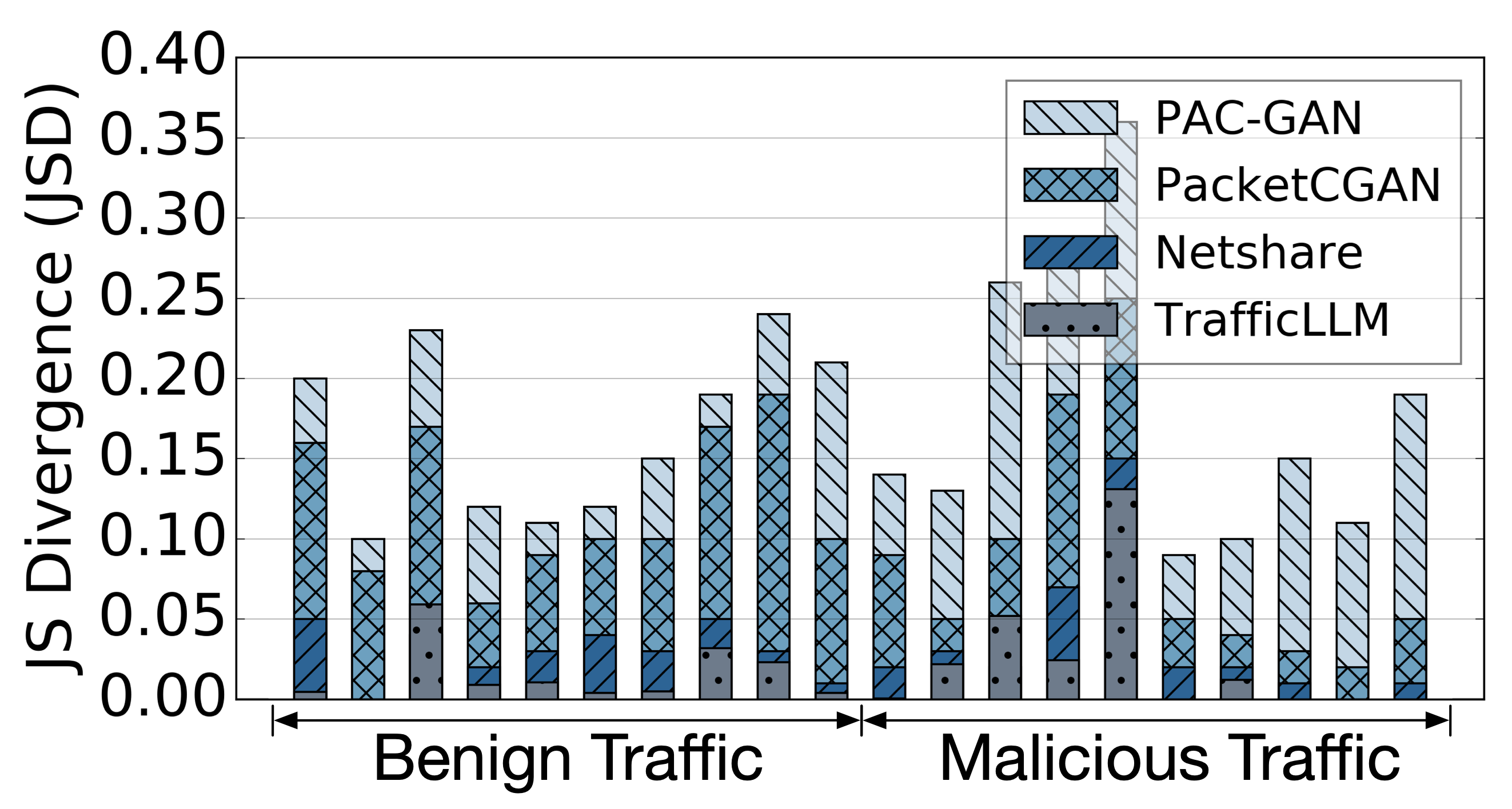}
}\label{fig7b}
\centering
\setlength{\abovecaptionskip}{-0.03cm}
\caption{JSD between real and synthetic distributions on ISCX Botnet 2014 and USTC TFC 2016 ($\Downarrow$ JSD is better).} 
\vspace{-0.3cm}
\label{fig7}
\end{figure}

\begin{table*}[t!]
\caption{The performance of identifying the 2k synthetic and real packets using the classifiers built from the same size of real and synthetic data under USTC TFC 2016, ISCX Botnet 2014, ISCX VPN 2016, and CSTNET 2023 datasets.}
\vspace{-0.4cm}
\begin{center}
\resizebox{\textwidth}{!}{
\begin{tabular}{l|c|ccc|ccc|ccc|ccc}
\toprule
\multirow{2}{*}{\textbf{Method}} & \multirow{2}{*}{\textbf{Setting}$^1$} & \multicolumn{3}{|c|}{\textbf{USTC TFC 2016}} & \multicolumn{3}{|c|}{\textbf{ISCX Botnet 2014}} & \multicolumn{3}{|c|}{\textbf{ISCX VPN 2016}} & \multicolumn{3}{|c}{\textbf{CSTNET 2023}}\\
\cmidrule{3-14}
& & \textbf{PR} & \textbf{RC} & \textbf{F1} & \textbf{PR} & \textbf{RC} & \textbf{F1} & \textbf{PR} & \textbf{RC} & \textbf{F1} & \textbf{PR} & \textbf{RC} & \textbf{F1}\\
\midrule
PAC-GAN~\cite{cheng2019pac} & \multirow{2}{*}{\ding{182}} & 0.7825 & 0.7432 & 0.7453 & 0.7234 & 0.7421 & 0.7256 & 0.8196 & 0.7945 & 0.8204 & 0.8402 & 0.8794 & 0.8571\\
PacketCGAN~\cite{wang2020packetcgan} & \multirow{2}{*}{R-Train} & 0.7673 & 0.7925 & 0.7529 &  0.8057 & 0.8245 & 0.8050 & 0.8342 & 0.8454 & 0.8345 & 0.8590 & 0.8881 & 0.8530\\
NetShare~\cite{yin2022practical} & \multirow{2}{*}{S-Test} & 0.8178 & 0.8157 & 0.8042 & 0.9021 & 0.9105 & 0.9042 & 0.9859 & 0.9675 & 0.9734 & 0.9314 & 0.9146 & 0.9345\\
\textsf{TrafficLLM} & & \textbf{0.9315} & \textbf{0.8604} & \textbf{0.8354} & \textbf{0.9778} & \textbf{0.9730} & \textbf{0.9727} & 0.9802 & \textbf{0.9754} & \textbf{0.9779} &  \textbf{0.9861} & \textbf{0.9852} & \textbf{0.9852}\\
\midrule
PAC-GAN~\cite{cheng2019pac} & \multirow{2}{*}{\ding{183}} & 0.6459 & 0.6375 & 0.6431 & 0.6204 & 0.6079 & 0.6202 & 0.8192 & 0.8023 & 0.8205 & 0.7056 & 0.6859 & 0.6980\\
PacketCGAN~\cite{wang2020packetcgan}& \multirow{2}{*}{S-Train} & 0.6724 & 0.6458 & 0.6532 & 0.7057 & 0.6894 & 0.6995 & 0.8525 & 0.8678 & 0.8548 & 0.6861 & 0.6404 & 0.6489\\
NetShare~\cite{yin2022practical} & \multirow{2}{*}{R-Test} & 0.8521 & 0.8254 & 0.8322 & 0.7974 & 0.7854 & 0.8024 & 0.9045 & 0.8845 & 0.8964 & 0.8420 & 0.8299 & 0.8405\\
\textsf{TrafficLLM} & & \textbf{0.8843} & \textbf{0.8641} & \textbf{0.8589} & \textbf{0.8634} & \textbf{0.8375} & \textbf{0.8334} & \textbf{0.9716} & \textbf{0.9686} & \textbf{0.9688} &  \textbf{0.8630} & 0.8289 &  \textbf{0.8340}\\
\bottomrule
\multicolumn{14}{l}{$^1$ Setting \ding{182} uses the real packets to train classifiers to test synthetic packets. Setting \ding{183} trains with synthetic packets and test on real packets.}\\
\end{tabular}
}
\vspace{-0.5cm}
\label{tab:classifier-real-against-fake}
\end{center}
\end{table*}

\subsection{Generalization across Generation Tasks}\label{sec:evaluation-packet-simulation} 

We evaluate \textsf{TrafficLLM}'s generation ability across different scenarios and analyze the practicality of the generated traffic.

\begin{figure}[t]
\centering
\subfigure[Source port]{    
\includegraphics[width=4.1cm]{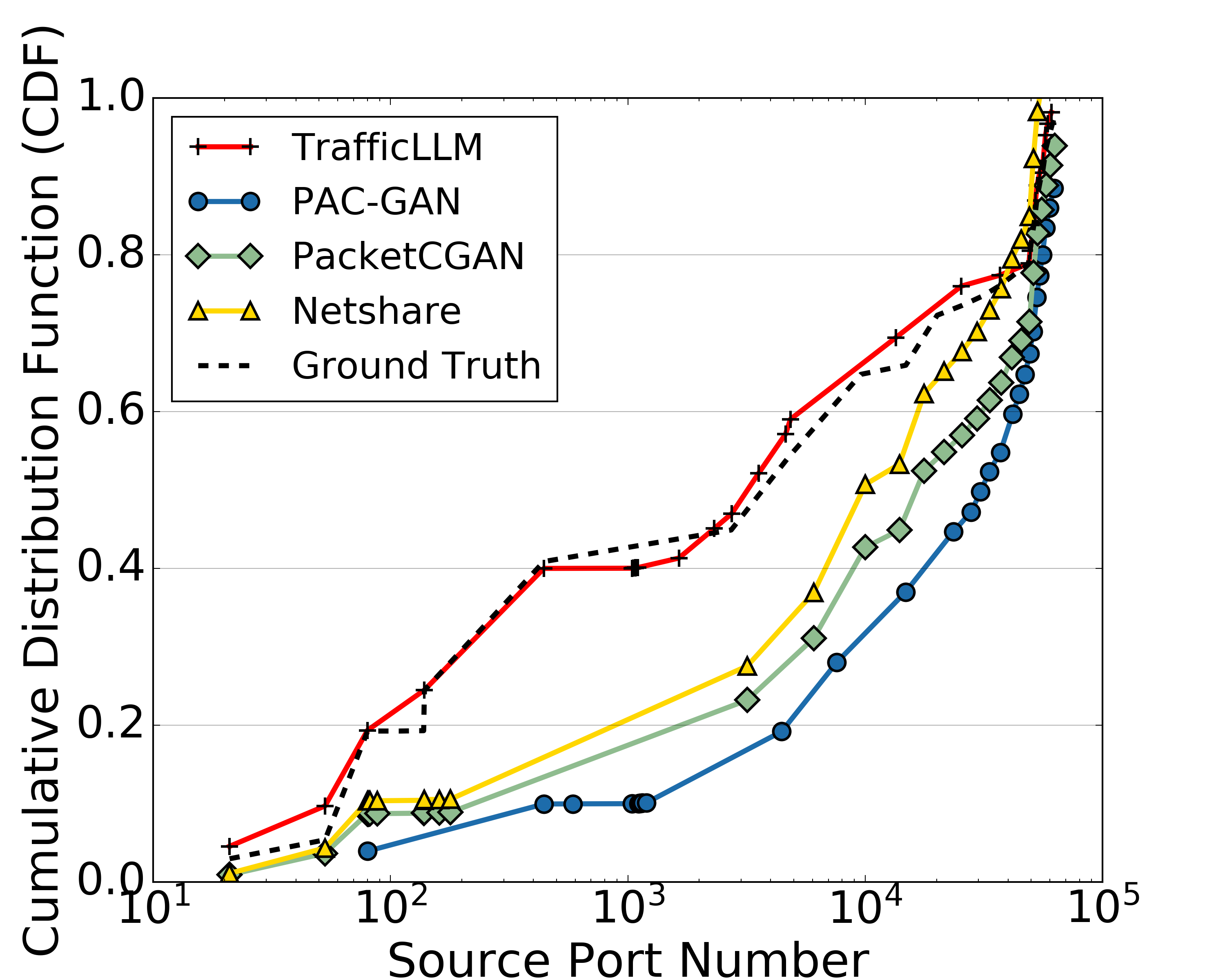}
}\label{fig8a}
\subfigure[Packet length (bytes)]{ 
\includegraphics[width=4.1cm]{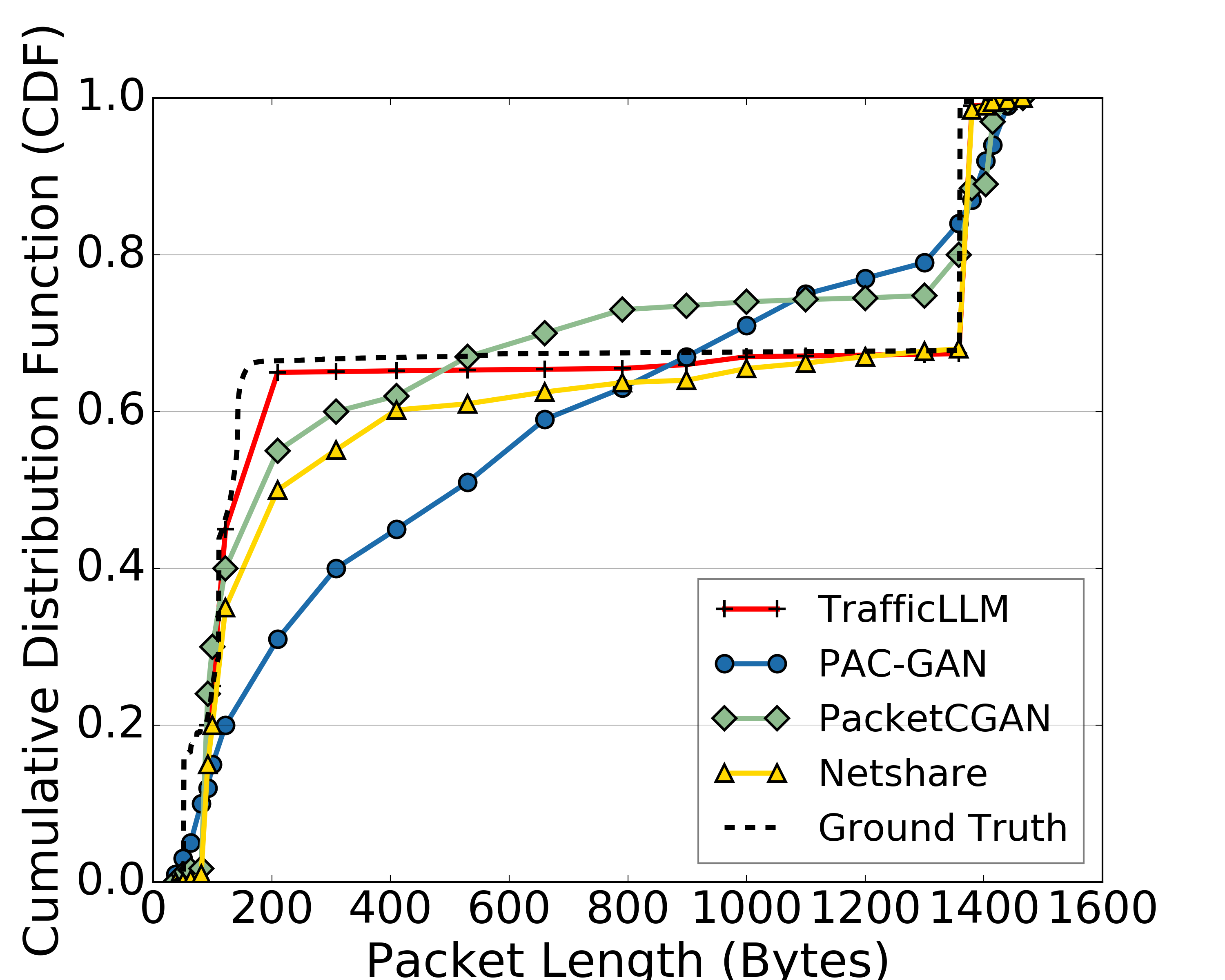}
}\label{fig8b}
\centering
\setlength{\abovecaptionskip}{-0.03cm}
\caption{Source port and packet length CDFs compared with baselines and the ground truth on ISCX VPN 2016.} 
\label{fig8}
\vspace{-0.3cm}
\end{figure}

\vspace{0.1cm}
\noindent \textbf{Distribution Metrics.} To evaluate the performance of the traffic generation capability, we compute the distribution metrics between the real and synthetic distribution of the meta-information in the packets. Figure~\ref{fig7} shows the results on 5-tuples compared to 3 baselines. We find that \textsf{TrafficLLM} is at most 73.76\% better than existing methods to degrade the gap from the real distribution. For different categories of benign and malicious traffic, \textsf{TrafficLLM} achieves an average JSD of 0.0179, which is 39.32\% better than the state-of-the-art traffic generation method NetShare's 0.0295 average JSD. For a more detailed analysis, Figure~\ref{fig8} shows the CDFs of the source port and packet length. Results indicate that existing GAN-based methods can not capture the distribution well. \textsf{TrafficLLM}'s memorization advantage from parameter volume can help restore the field values of the original traffic data. The results also indicate that \textsf{TrafficLLM} effectively learns the fine-grained traffic representations to make the distribution consistent with the ground truth. 

\vspace{0.1cm}
\noindent \textbf{Practicality of Synthetic Samples.} \textsf{TrafficLLM} can rebuild the packets of raw traffic using the generated meta-information. Furthermore, we consider that the generation ability of \textsf{TrafficLLM} has two promising applications: (i) generating packets for security tests and (ii) building classifiers under few-shot scenarios. First, to verify the quality of the generated packets, we use 2k traffic traces of real datasets to build a robust ensemble model based on Multinomial Naive Bayesian and SGD classifiers to construct a prototype of ML-based NIDS. Then we use the synthetic traces to test whether the model can identify these packets. In Table~\ref{tab:classifier-real-against-fake}, we find that the generated packets of \textsf{TrafficLLM} can be identified by the real-world classifier with an average F1-score of 0.9483, which is 4.68\% better than the state-of-the-art method NetShare. It means that \textsf{TrafficLLM} can generate test samples with extremely high confidence. To measure the second application of \textsf{TrafficLLM}, we use 2k generated packets to build the classifiers and treat the real packets as the test sets. Results indicate that \textsf{TrafficLLM} outperforms baselines on most metrics to detect real-world traffic using the classifiers built from synthetic data. \textsf{TrafficLLM} can be applied in data augmentation for benign or malicious traffic with a 0.8739 average F1-score, which is 3.07\%-33.92\% better than baselines.

\begin{figure}[t!]
\centering
\subfigure[No Drift]{    
\includegraphics[width=2.6cm]{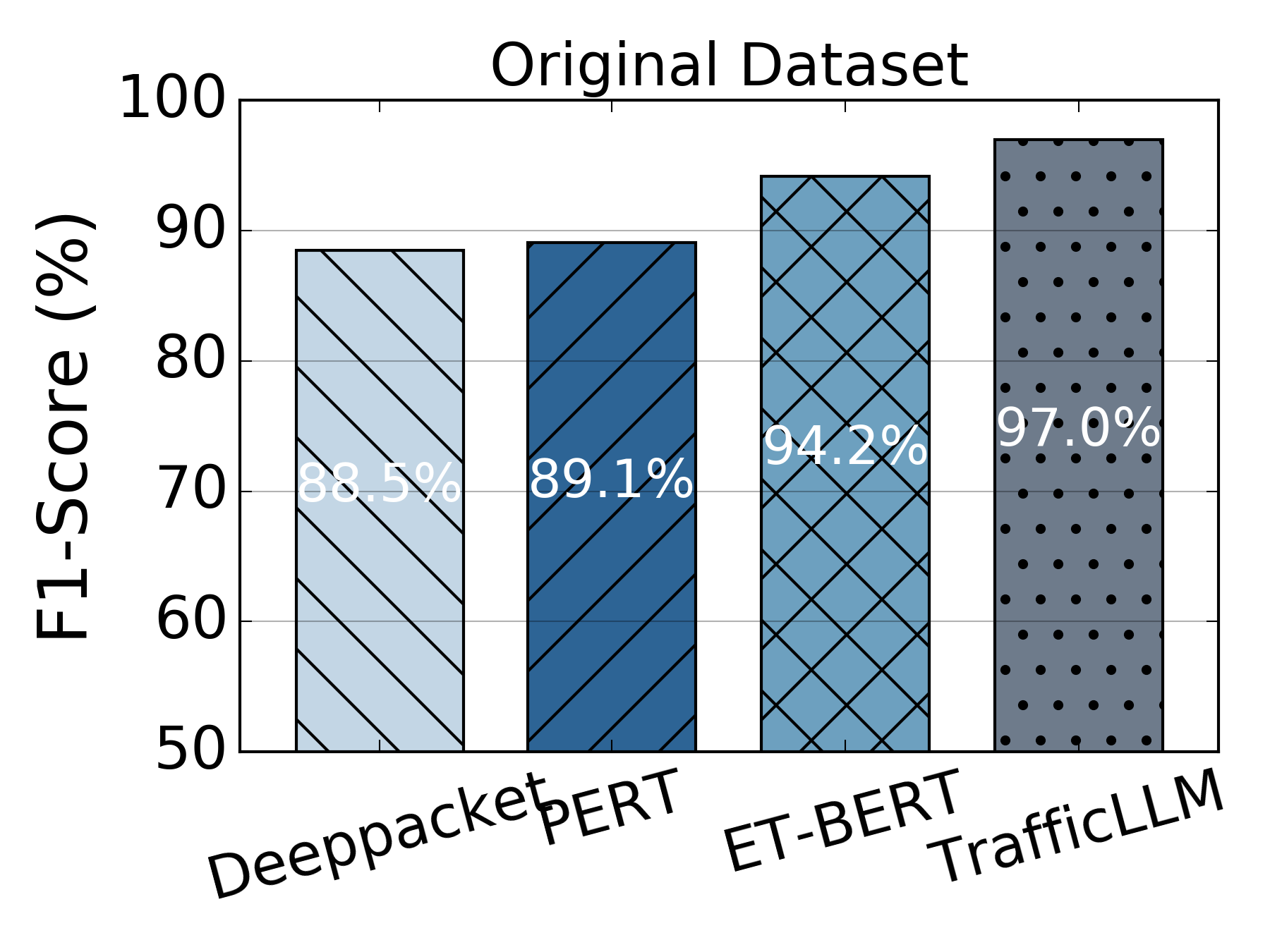}
}\label{fig11a}
\subfigure[Time Drift]{ 
\includegraphics[width=2.6cm]{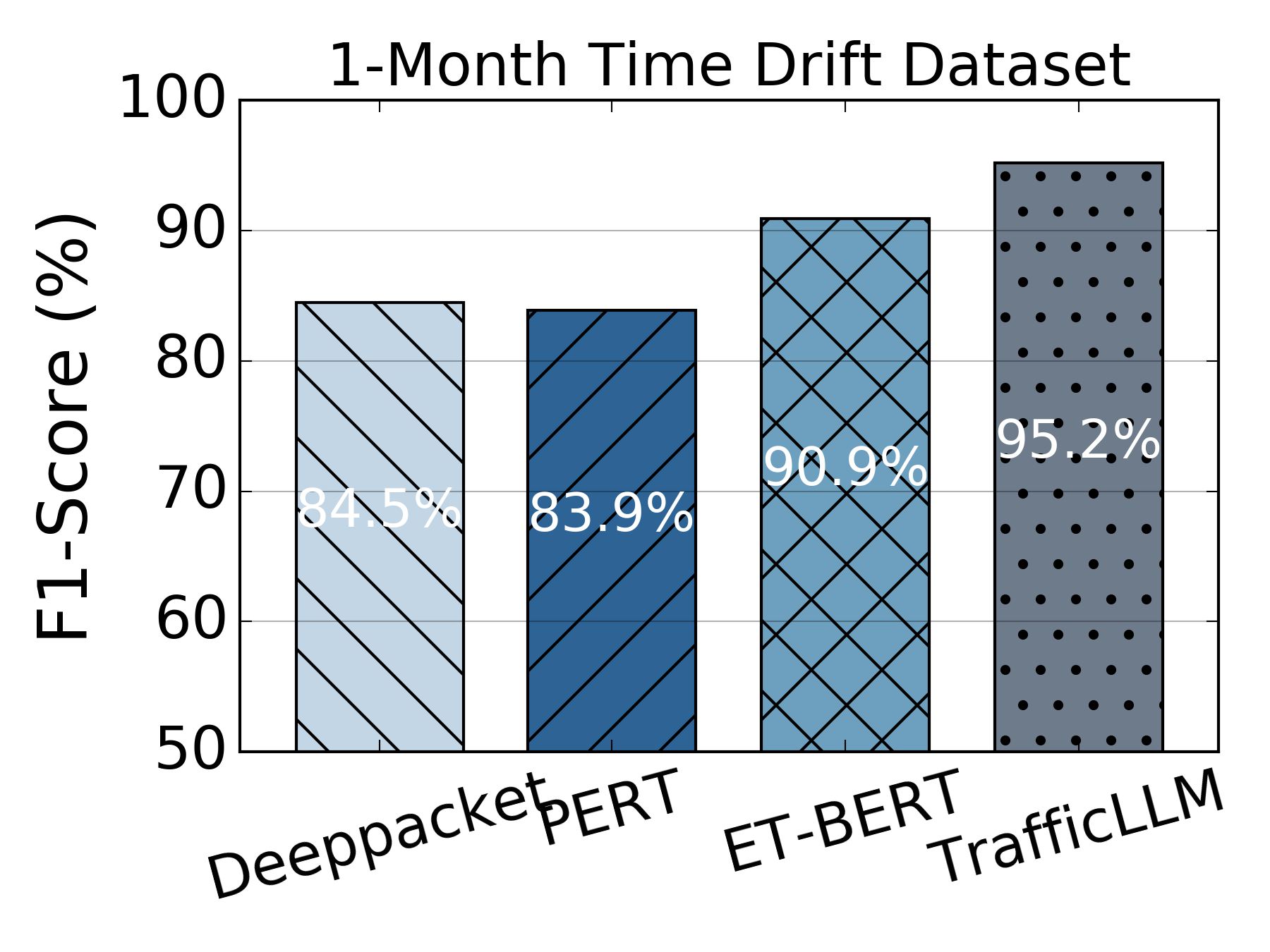}
}\label{fig11b}
\subfigure[Version Drift]{ 
\includegraphics[width=2.6cm]{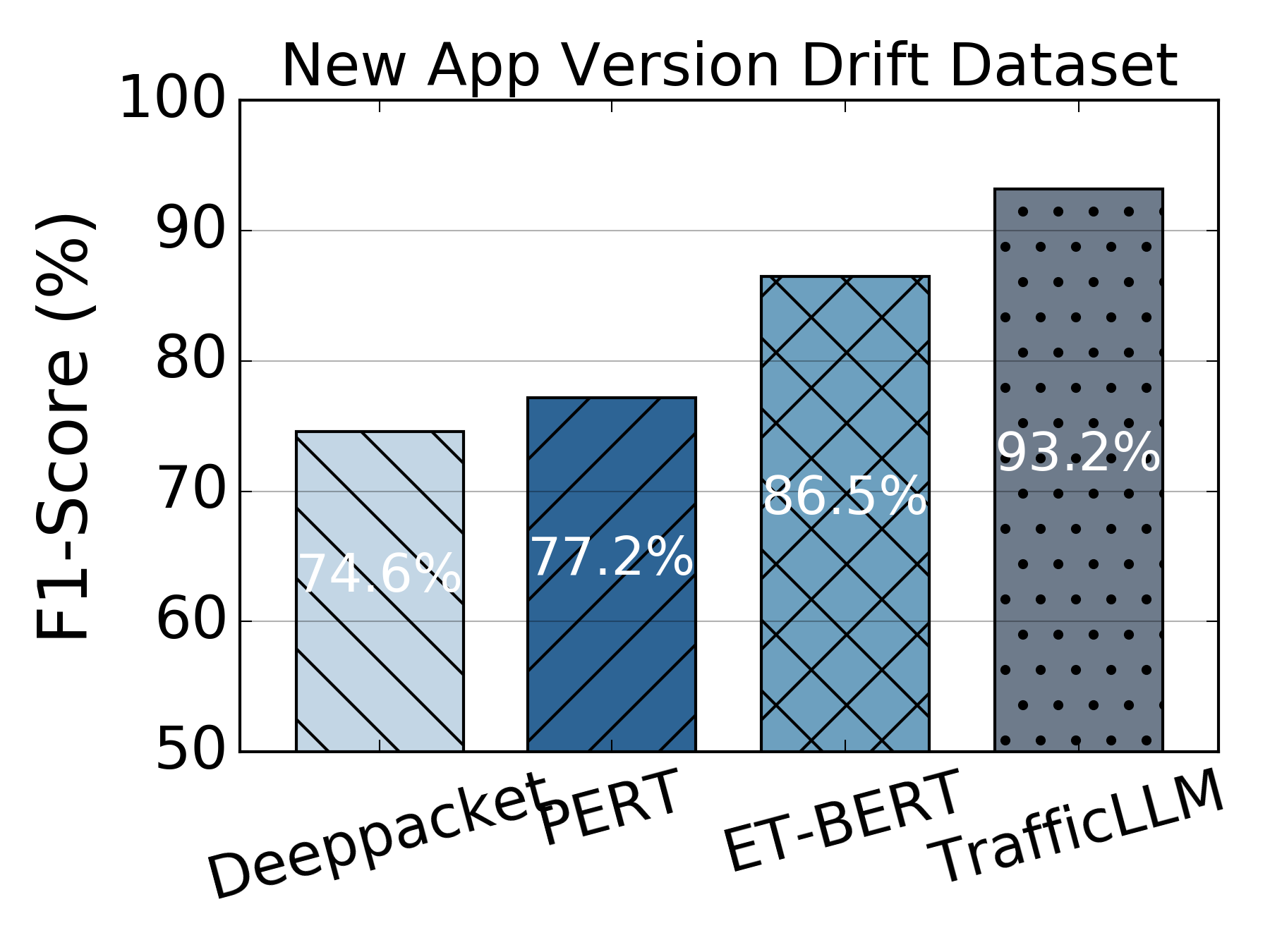}
}\label{fig11c}
\centering
\setlength{\abovecaptionskip}{-0.03cm}
\caption{Concept drift experiments with 1-month time drift and new App version drift settings on APP-53 2023 datasets.} 
\label{fig11}
\vspace{-0.5cm}
\end{figure}

\begin{figure}[t!]
\centering
\subfigure[Stage-2 APT Attacks]{    
\includegraphics[width=2.6cm]{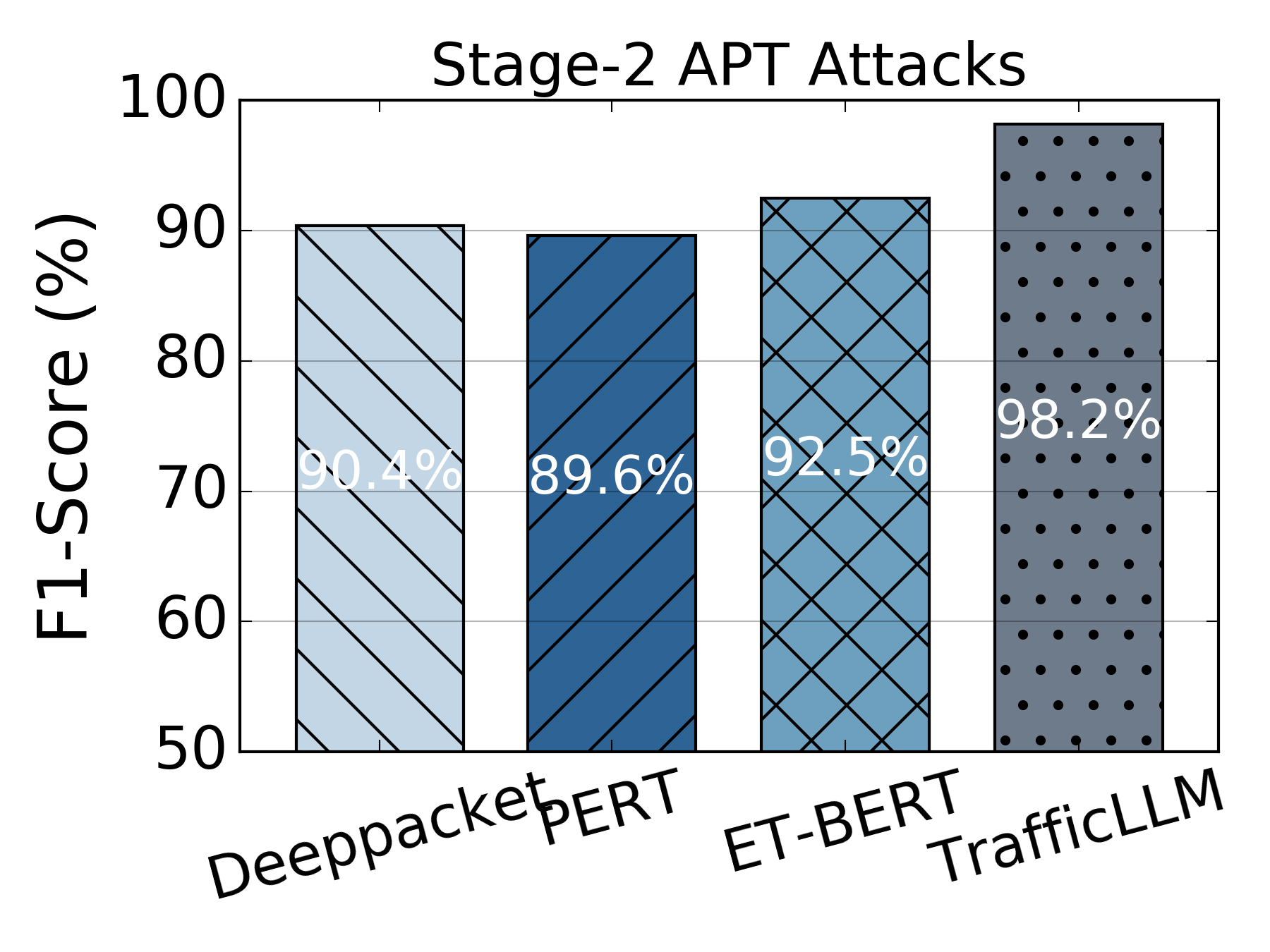}
}\label{fig12a}
\subfigure[Stage-3 APT Attacks]{ 
\includegraphics[width=2.6cm]{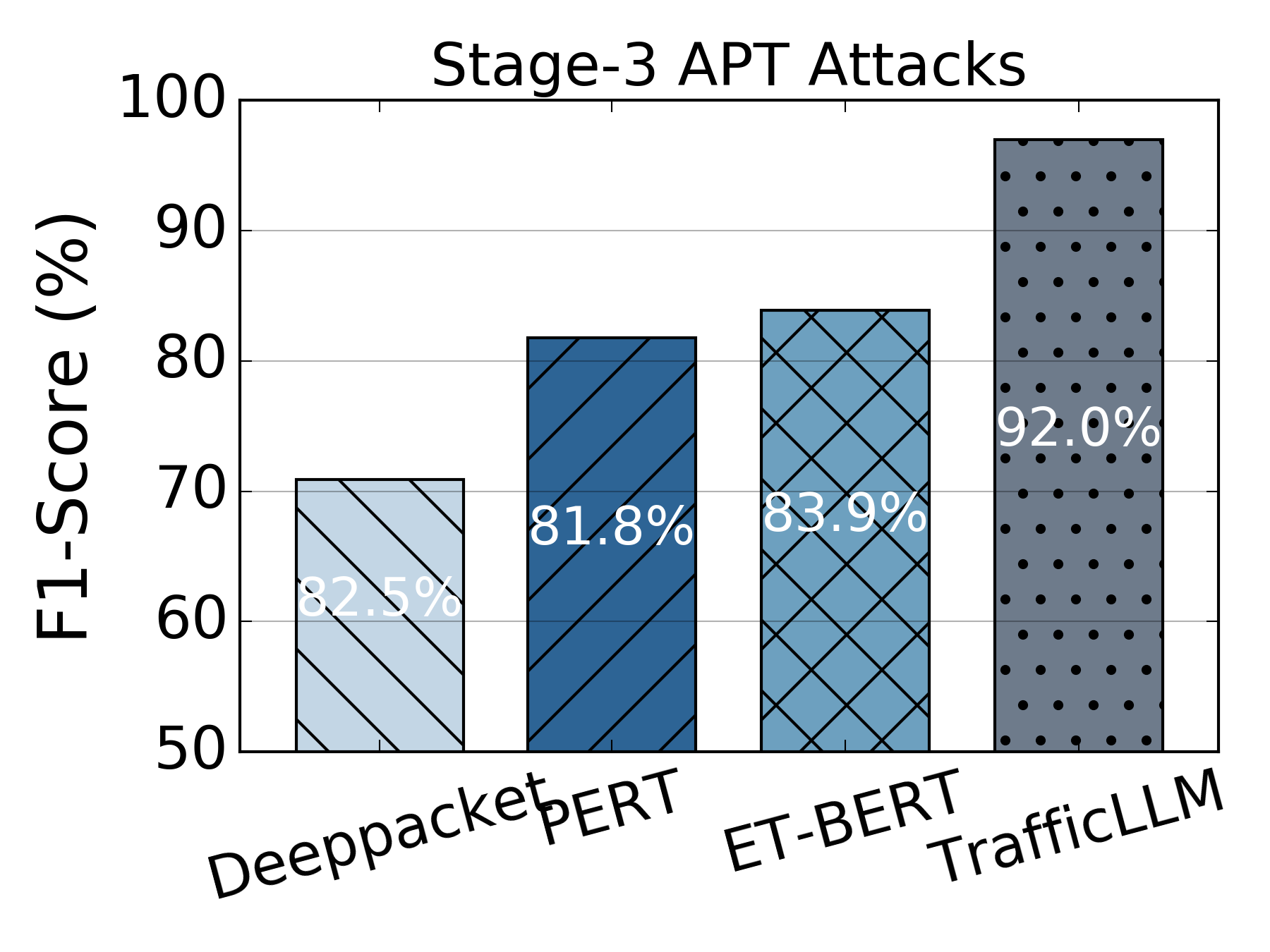}
}\label{fig12b}
\subfigure[Stage-4 APT Attacks]{ 
\includegraphics[width=2.6cm]{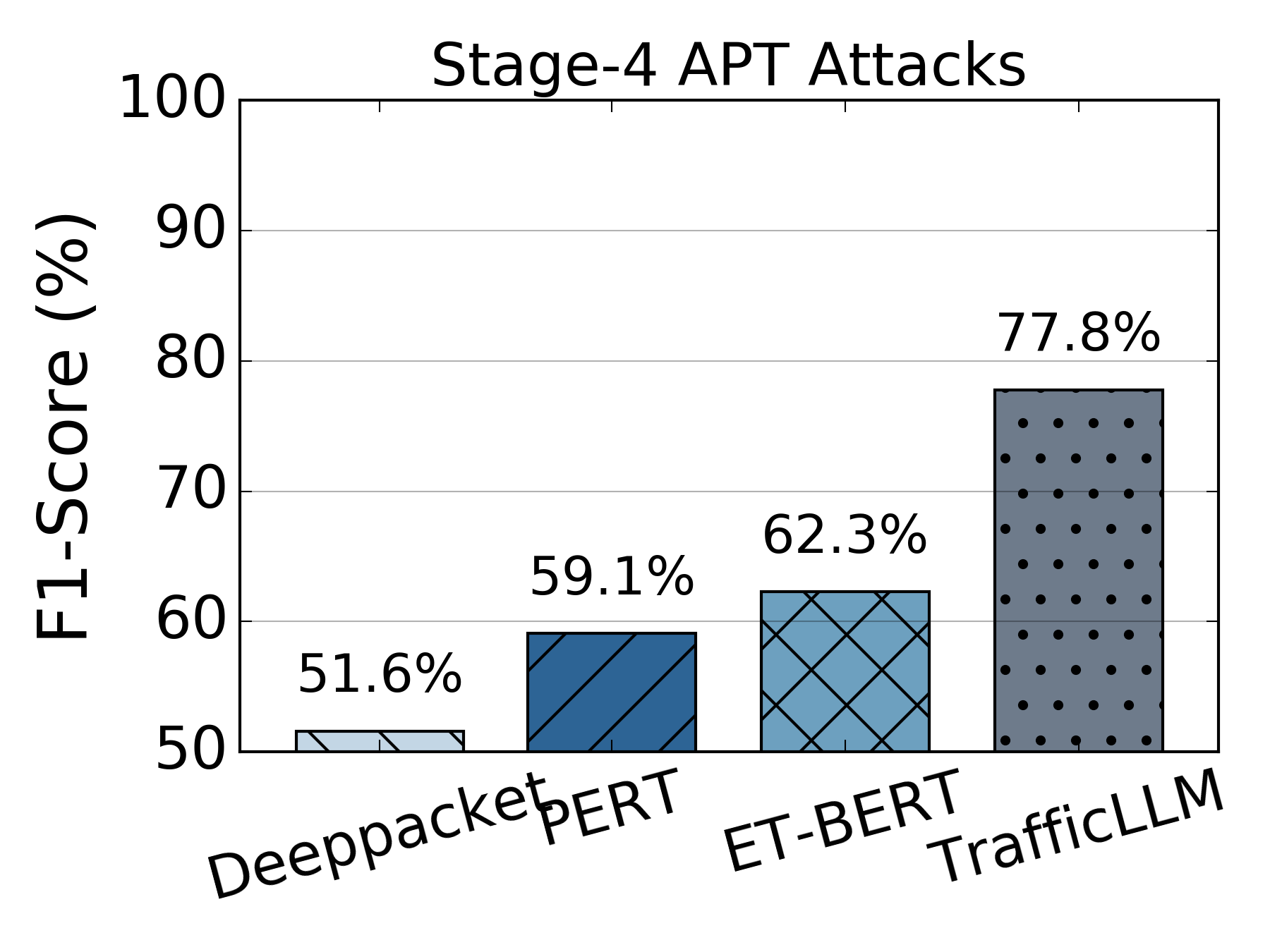}
}\label{fig12c}
\centering
\setlength{\abovecaptionskip}{-0.03cm}
\caption{Future stage APT attack detection based on historical stage-1 APT attack knowledge on DAPT 2020 datasets.} 
\label{fig12}
\vspace{-0.5cm}
\end{figure}


\subsection{Generalization to Unseen Data}
Unlike traditional ML-based methods, a salient property of \textsf{TrafficLLM} is its generalization abilities on unseen data. We utilize concept drift and APT attack datasets~\cite{jiang2023zero,myneni2020dapt} to set up unseen data detection scenarios for evaluation.



\vspace{0.1cm}
\noindent \textbf{Time and Version Drift.} We evaluate \textsf{TrafficLLM} under concept drift~\cite{yang2021cade} scenarios that the detecting traffic distribution often shifts from the original training data over time due to dynamic behaviors (e.g., application updating). In this setting, we use the APP-53 2023 dataset, including 53-type mobile App traffic and its drifted traffic with a 1-month time interval and new App version updates. We train \textsf{TrafficLLM} on the historical traffic and evaluate its generalization performance on the no drift, 1-month time interval, and new App version drift datasets. In Figure~\ref{fig11}, \textsf{TrafficLLM} effectively maintains the detection performance when facing the concept drift scenarios. It outperforms baselines with 4.3\%-11.3\% and 6.7\%-18.6\% F1-score under time and version drift respectively. Since \textsf{TrafficLLM} inherits the strong generalization from LLMs, it captures the common traffic representations in drifted environments, which ensures the robustness of detecting unseen drifted traffic.


\vspace{0.1cm}
\noindent \textbf{APT Attack Detection.} We then evaluate \textsf{TrafficLM} under APT attacks~\cite{myneni2020dapt} that the traffic contains unseen distribution due to dynamic behavior changes of attackers. We evaluate it on the multi-stage APT attack detection tasks using the DAPT 2020 dataset~\cite{myneni2020dapt}. In this setting, the adversary conducts 4-stage attacks including reconnaissance, foothold establishment, lateral movement, and data exfiltration in a few days. We train \textsf{TrafficLLM} with the benign and stage-1 APT attack traffic and test it on the attacks of future stages. As shown in Figure~\ref{fig12}, results indicate that \textsf{TrafficLLM} achieves 89.3\% average F1-score to detect future-stage attack traffic. Although the attack traffic is extremely different in these stages (e.g., stage-1 attack mainly aims to identify vulnerabilities with attack tools, while the stage-4 attack contains the traffic that downloading files from the victim server using C\&C tunnels), the pattern mining ability of \textsf{TrafficLLM} can differ the representation of malicious traffic from the benign. This robust anomaly detection ability helps \textsf{TrafficLLM} achieve better generalization performance on the dynamic traffic distribution compared to the three baselines.



\begin{figure}[t!]
\centering
\subfigure[Avg. F1-Score]{    
\includegraphics[width=4.1cm]{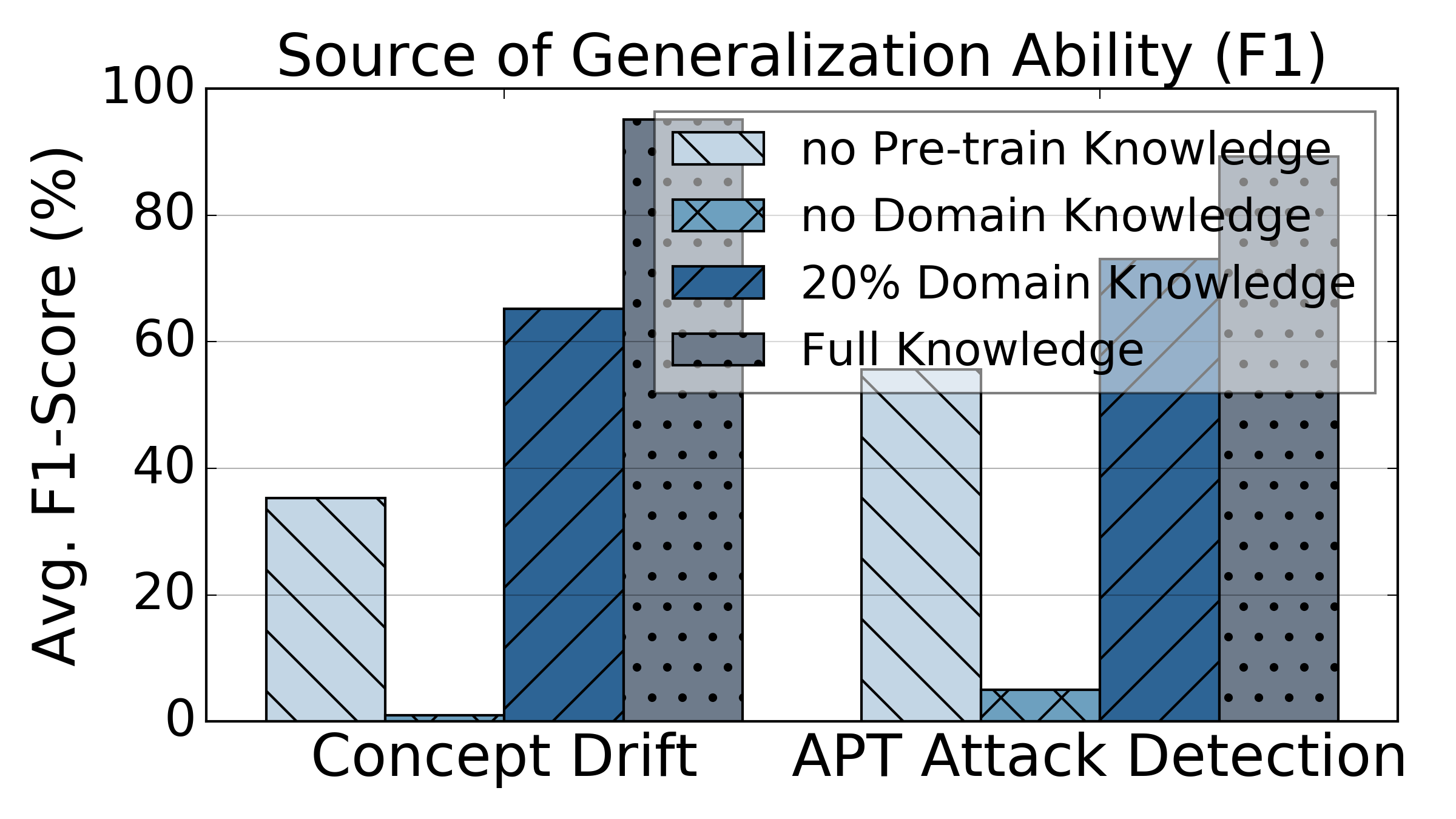}
}\label{fig18a}
\subfigure[False Positives]{ 
\includegraphics[width=4.1cm]{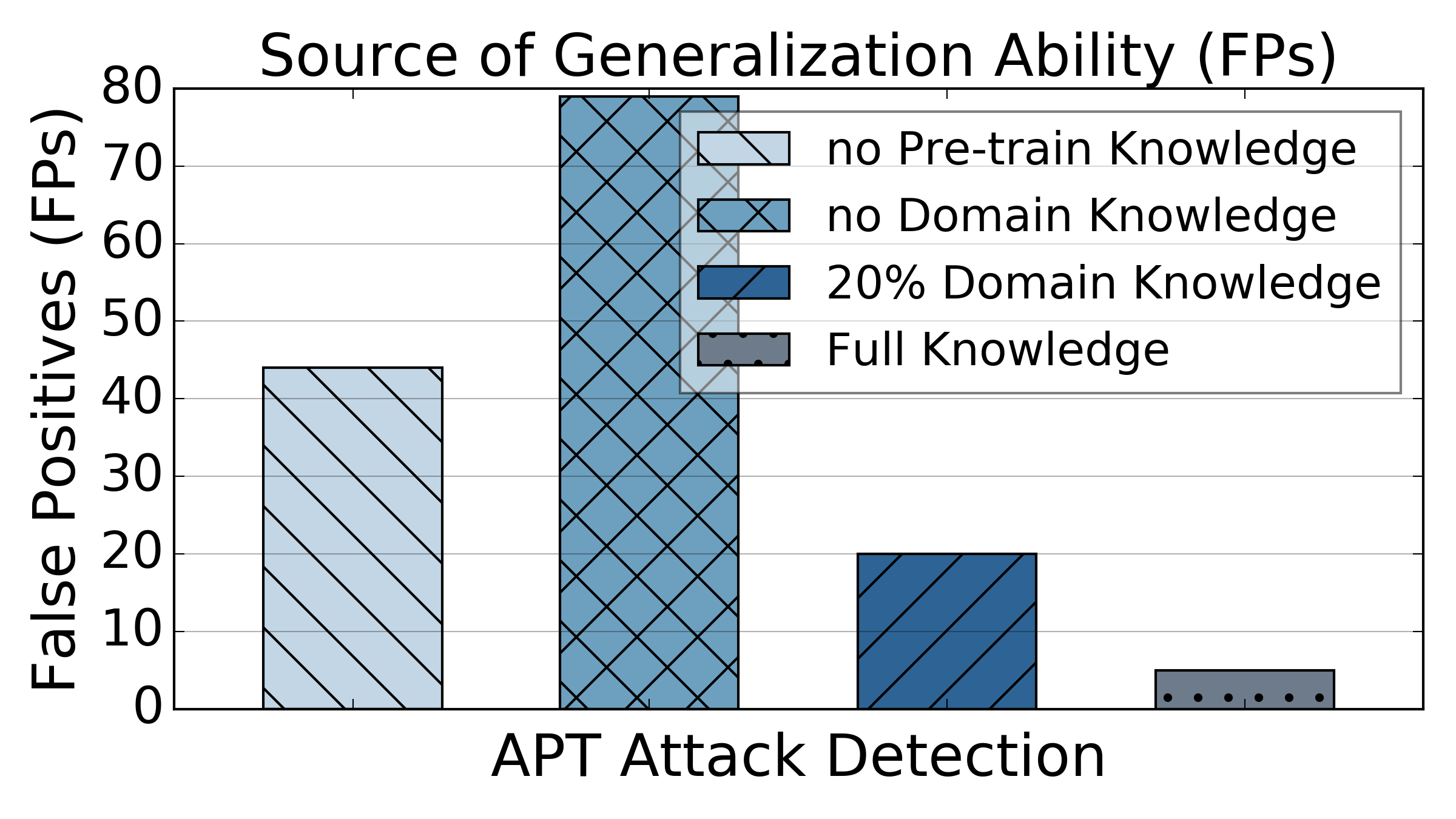}
}\label{fig1b}
\centering
\setlength{\abovecaptionskip}{-0.03cm}
\caption{Exploring the influence of the Pre-trained and traffic-domain knowledge on the generalization ability.} 
\label{fig18}
\vspace{-0.3cm}
\end{figure}

\vspace{0.1cm}
\noindent \textbf{Remark on Generalization Ability.} To further analyze why \textsf{TrafficLLM} keeps such generalization ability that traditional ML-based methods do not have, we investigate the influence of the capability from native base LLMs and tuned traffic-domain PEFT models. First, we randomly initialize LLM's weights to remove the pre-trained knowledge and directly tune it with traffic datasets. We reuse the same settings of APP-53 2023 and DAPT 2020 datasets and report the average F1-score and FP. As shown in Figure~\ref{fig18}, \textsf{TrafficLLM} dramatically drops the performance without pre-trained knowledge. This is because the pre-trained knowledge helps LLM acquire pattern mining and reasoning abilities through massive corpora (e.g., planning and calculation~\cite{llama2}), which is also critical for traffic analysis. Second, we disable the traffic-domain knowledge by removing PEFT models and directly testing on the base model. Without any guidance of traffic-domain knowledge, LLM tends to randomly generate labels, making it not qualified for the work. However, when using 20\% training data to inject partial traffic-domain knowledge into LLMs, \textsf{TrafficLLM} acquires great improvement on the unseen traffic data. The domain knowledge helps LLM effectively activate the generalization ability on the traffic data, building generic traffic representations on unseen traffic.

\begin{figure}[t!]
\centering
\subfigure[Robustness of Different LLMs]{    
\includegraphics[width=4.1cm]{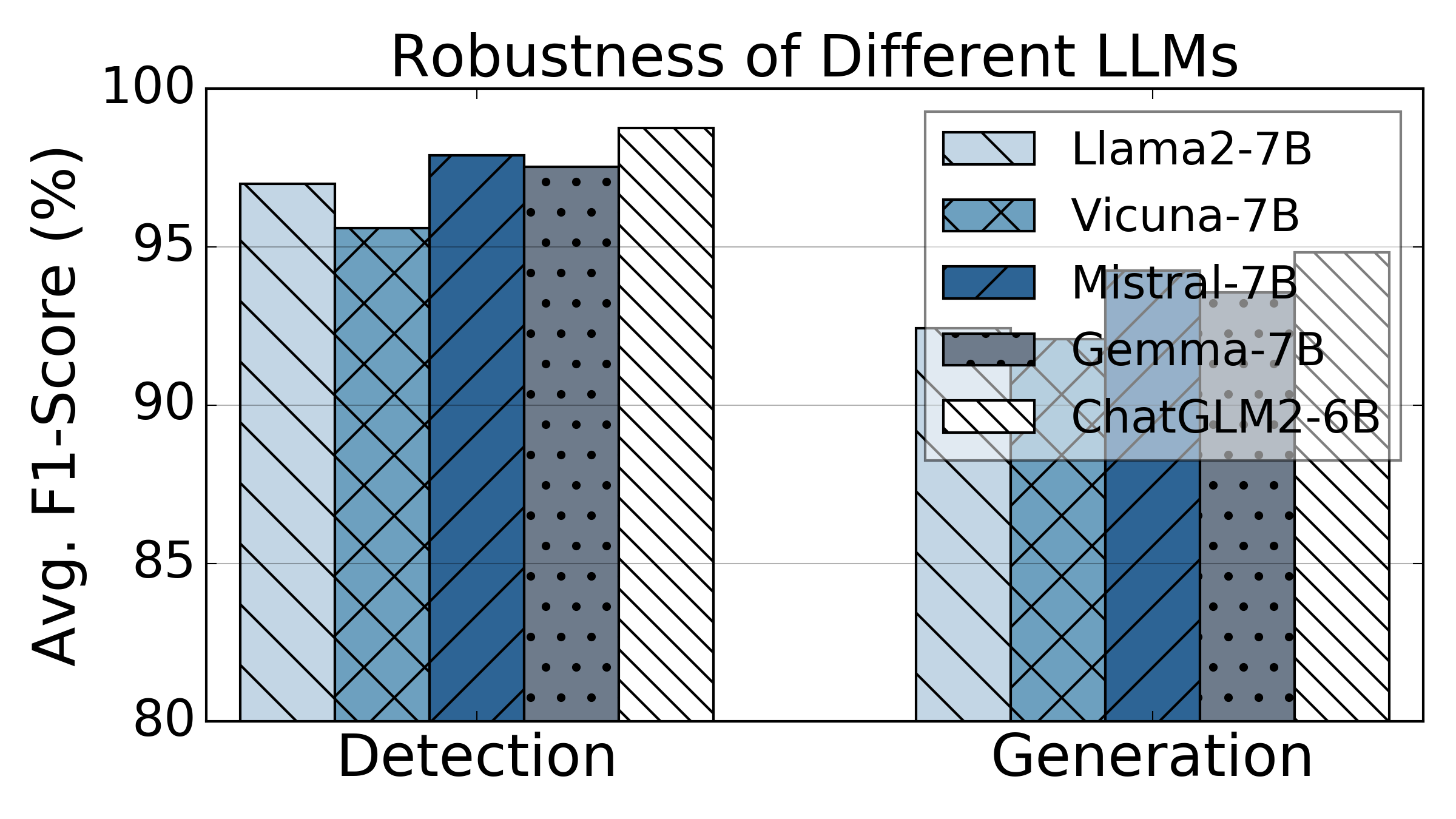}
}\label{fig13a}
\subfigure[Impact of LLM Parameter Size]{ 
\includegraphics[width=4.1cm]{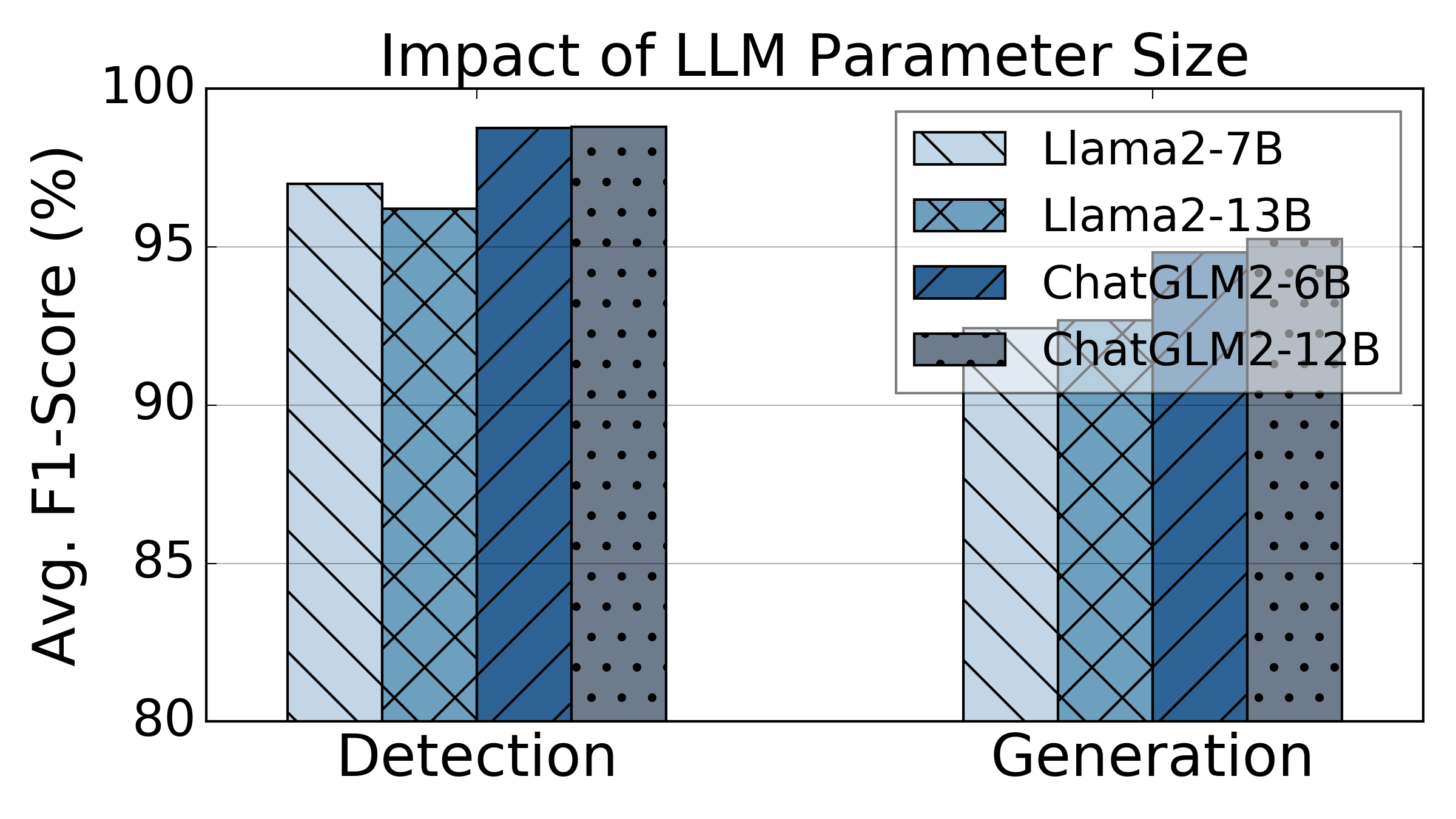}
}\label{fig13b}
\centering
\setlength{\abovecaptionskip}{-0.03cm}
\caption{Different types and model sizes of LLM adapted in \textsf{TrafficLLM}. The generation performance uses the R-Train and S-Test settings (the same as below).} 
\label{fig13}
\vspace{-0.5cm}
\end{figure}

\begin{figure}[t!]
\centering
\subfigure[Traffic Detection]{    
\includegraphics[width=2.3cm]{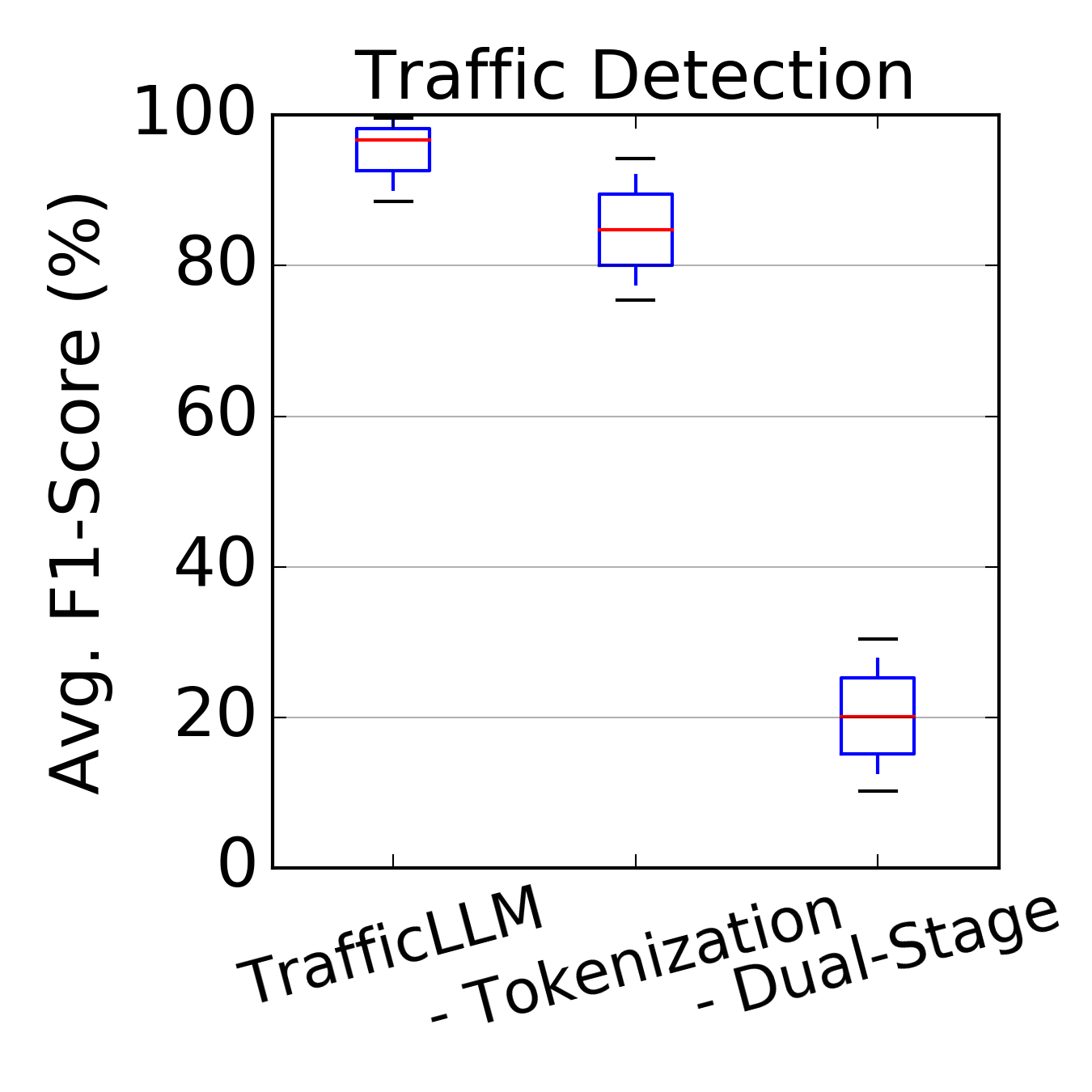}
}\label{fig14a}
\subfigure[Traffic Generation]{ 
\includegraphics[width=2.3cm]{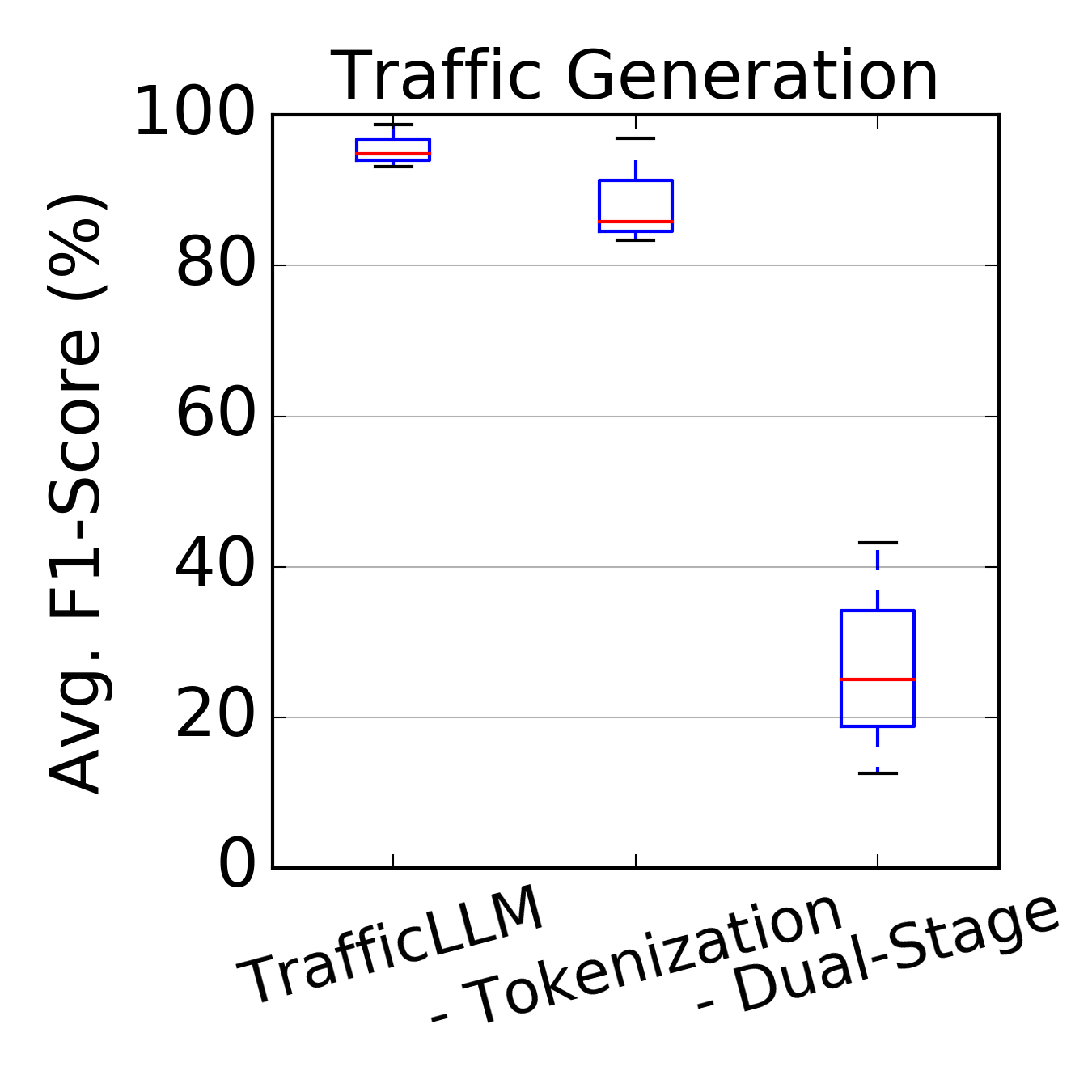}
}\label{fig14b}
\subfigure[Computation Overhead]{ 
\includegraphics[width=3.1cm]{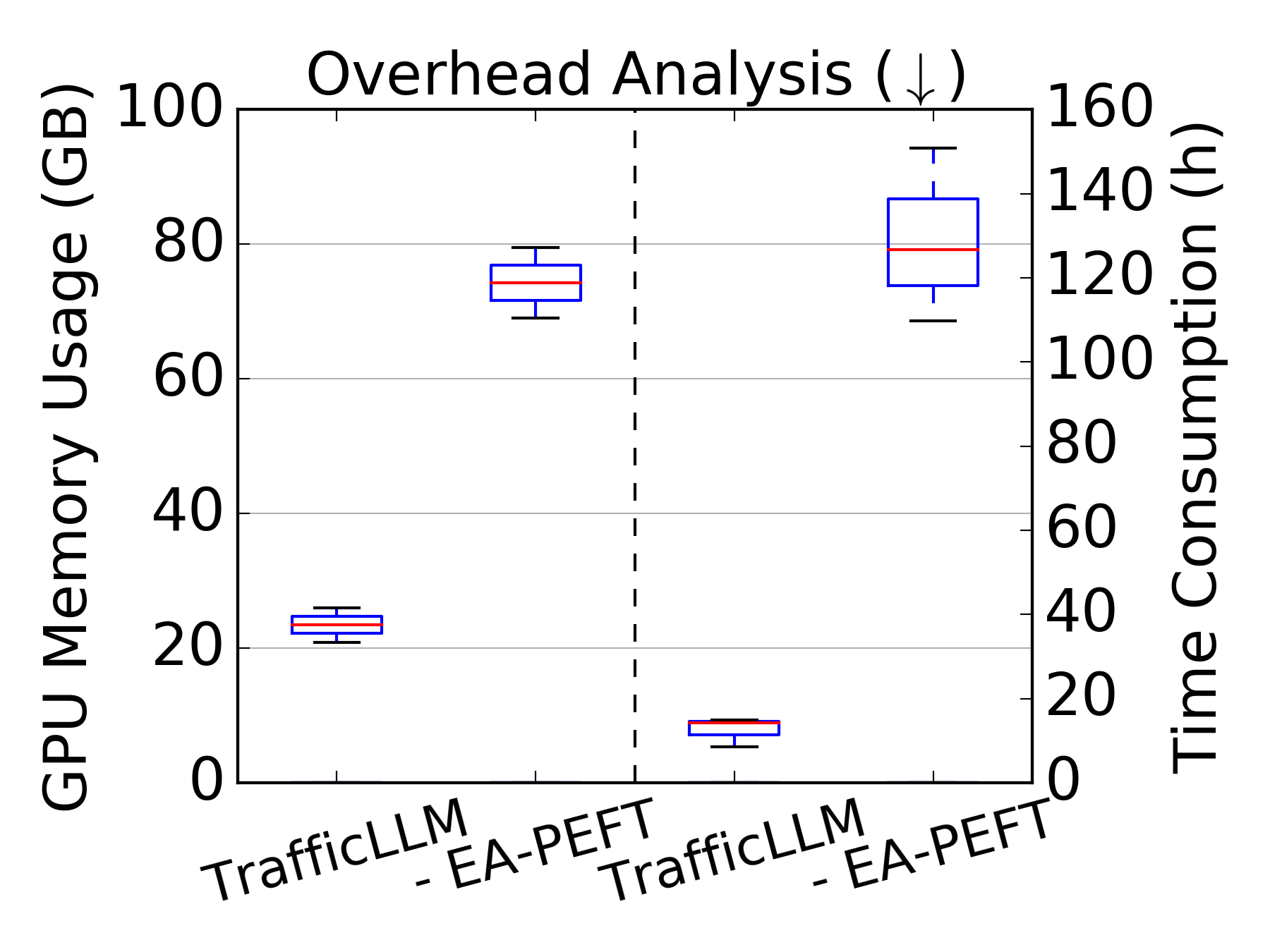}
}\label{fig14c}
\centering
\setlength{\abovecaptionskip}{-0.03cm}
\caption{The effectiveness of traffic-domain tokenization, dual-stage tuning pipeline, and EA-PEFT in \textsf{TrafficLLM}.} 
\label{fig14}
\vspace{-0.3cm}
\end{figure}

\subsection{Deep Dive}\label{sec:evaluation-interpretability}

\noindent \textbf{Adaptability across different LLMs.} To verify whether \textsf{TraffiLLM} is applicable to different LLMs, except for Llama2 and ChatGLM2, we employ additional state-of-the-art LLMs to build \textsf{TrafficLLM}, which includes Vicuna~\cite{vicuna}, Mistral~\cite{mistral}, and Gemma~\cite{gemma}. As shown in Figure~\ref{fig13a}, results indicate that the framework of \textsf{TraffiLLM} can be easily applied to all open-sourced LLMs with strong performance.

\vspace{0.1cm}
\noindent \textbf{Impact of LLM Parameter Size.} In Figure~\ref{fig13b}, we explore the impact of LLM's parameter size on traffic detection and generation performance. Although Llama2-13B and ChatGLM2-12B's parameter size is almost 2-fold that of their 7B and 6B versions, the four LLMs achieve similar performance. The 6B model is enough to outperform existing ML-based baselines on traffic detection and generation tasks.

\vspace{0.1cm}
\noindent \textbf{Effectiveness of Core Components.} To validate the effectiveness of \textsf{TrafficLLM}'s core components, we build three \textsf{TrafficLLM}'s variants by removing the traffic-domain tokenizer (- Tokenization), replacing the dual-stage tuning pipeline with default tuning (- Dual-Stage), and replacing EA-PEFT scheme with full fine-tuning (- EA-PEFT). As shown in Figure~\ref{fig14}, modifying these components will entail 7.2\%-78.7\% performance reduction and 927.9\% time and 216.2\% GPU memory overhead increase among five LLMs mentioned above, which indicates the significance of these components.

\vspace{0.1cm}
\noindent \textbf{Overhead Analysis.} We investigate \textsf{TrafficLLM}'s computation overhead on a NVIDA A100-80GB GPU. Training a 6B model like ChatGLM2-6B requires 23GB GPU memory and 14h training time for a new PEFT model update (involving 20,000 training steps on 50,000 task-specific samples). During the inference stage, loading \textsf{TrafficLLM} requires 13GB GPU memory and takes about 0.2s or 10s to generate a predicted label or a 1000-token synthetic packet. To reduce the overhead, we consider employing a smaller LLM or using compression methods~\cite{xu2023survey} can help speed up the adaptation. 

\subsection{Real-World Evaluation}
To evaluate the practicality in real-world scenarios, we opened \textsf{TrafficLLM}'s framework for extensive security practitioners in an LLM competition and deployed \textsf{TrafficLLM} in a top security company.

\begin{figure}[t]
\centering
\subfigure[Evaluation on the ATEC 2023]{    
\includegraphics[width=3.6cm]{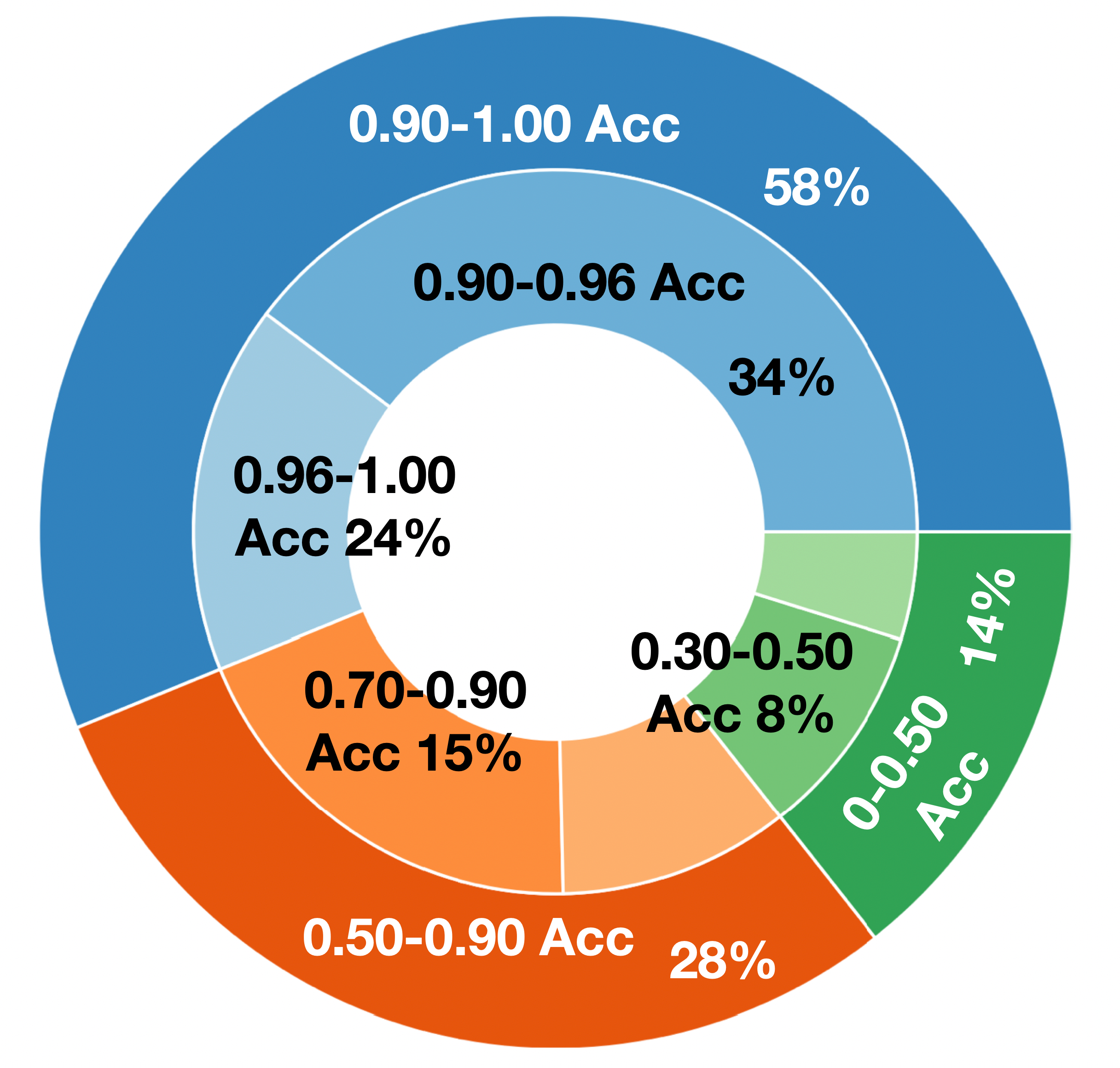}
}\label{fig15a}
\subfigure[Enterprise Deployment]{ 
\includegraphics[width=4.3cm]{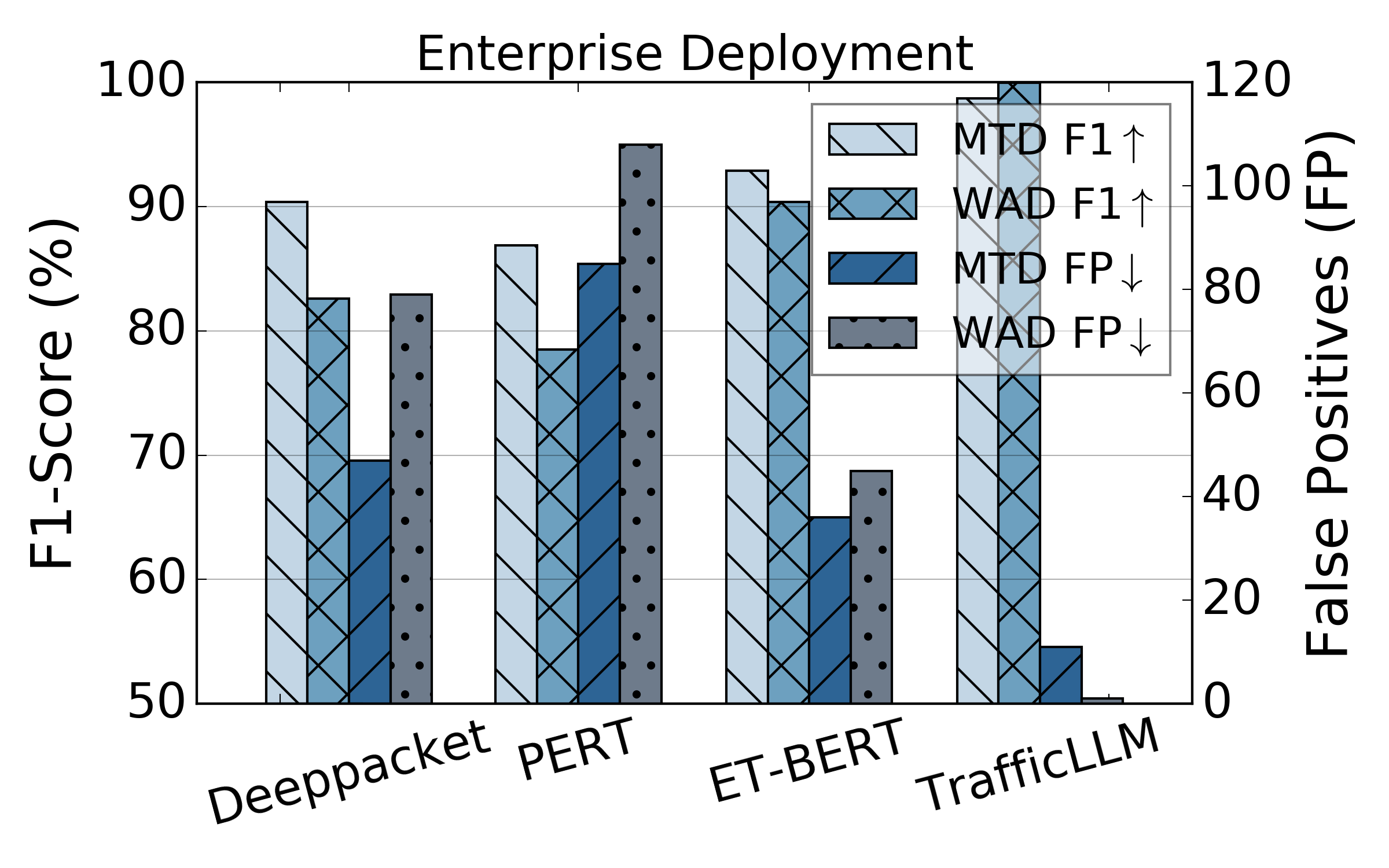}
}\label{fig15b}
\centering
\setlength{\abovecaptionskip}{-0.03cm}
\caption{Players' model performance in the competition and detection performance under enterprise deployment.} 
\label{fig15}
\vspace{-0.4cm}
\end{figure}

\vspace{0.1cm}
\noindent \textbf{Individual Use Evaluation.} To investigate \textsf{TrafficLLM}'s effectiveness in research and collect feedback from the community, we integrate \textsf{TrafficLLM} as a race track on a national LLM competition\footnote{ATEC 2023 Website: \href{https://www.atecup.cn/matchHome/100001}{https://www.atecup.cn/matchHome/100001}} with 1,901 teams and over 3,000 players from about 200 institutions from November 2023 to March 2024. We have united many universities and Internet companies to develop the competition platform. In this track, players must use \textsf{TrafficLLM}'s framework with custom instructions and traffic data to tackle MTD, BND, and EVD tasks. Each player can train their model and submit it in our scoring system online. At the end of the competition, we draw up the statistics for all the players' scores. As shown in Figure~\ref{fig15a}, 58\% of players achieve above 90\% accuracy in this competition. Even 24\% of player models performed better than 96\% Accuracy. After extensive verification by players during the competition, \textsf{TrafficLLM} was proven to adapt LLMs with powerful performance easily.

\vspace{0.1cm}
\noindent \textbf{Enterprise Deployment.} We deploy \textsf{TrafficLLM} in a security company that offers signature-based WAF and NIDS services to hundreds of its enterprise customers. The services are operated by security specialists to record the daily malware and Web attacks through manual analysis behind the services. They generated the ground truth by matching the WAF rules or manually analyzing the traffic log with abnormal behaviors (e.g., containing attack payloads). We replay the recorded traffic that contains 17,556 and 7,083 flows of malware and Web attack traffic to conduct MTD and WAD tasks using \textsf{TrafficLLM}. As shown in Figure~\ref{fig15b}, \textsf{TrafficLLM} outperforms baselines by reaching 98.7\% and 99.8\% F1-scores in the real-world MTD and WAD task. \textsf{TrafficLLM} effectively reduces at least 69\% and 95\% of FPs compared to existing ML-based methods. This is attributed to \textsf{TrafficLLM}'s robust representation learning to differ the pattern of benign and abnormal traffic, making \textsf{TrafficLLM} release strong generalization in the real-world scenario.

\section{Related Work}\label{sec:related-work}

\noindent \textbf{Encrypted Traffic Classification.}
Encrypted traffic classification is a crucial technique for network management and security monitoring. With the inability to inspect the content of encrypted packets directly, researchers have turned to various machine-learning algorithms~\cite{taylor2017robust,van2020flowprint,fu2023detecting} to analyze patterns, timings, packet sizes, and other metadata to classify traffic. For instance, Van et al.~\cite{van2020flowprint} used semi-supervised methods to model device, certificate, and statistical features for mobile app fingerprinting. Taylor et al.~\cite{taylor2017robust} utilized packet size to train traffic classifiers with Random Forest. Fu et al.~\cite{fu2023detecting} exploited flow interaction graphs to distinguish malicious behaviors from benign traffic. Recent research~\cite{lin2022bert,he2020pert} focused more on model generalization using pre-trained models. For instance, Lin et al.~\cite{lin2022bert} leveraged BERT to build the pre-trained model for multi-type traffic classification tasks. However, existing works only focus on handling traffic data. Unlike these works, \textsf{TrafficLLM} jointly learns expert instructions and raw traffic data, making \textsf{TrafficLLM} more powerful in various traffic analysis tasks.

\vspace{0.1cm}
\noindent \textbf{Traffic Data Augmentation.} Data augmentation has been widely applied in the few-shot scenarios of traffic analysis to increase the amount and diversity of traffic data. For example, Qing et al.~\cite{qing2023low} utilized distributions of traffic data in the feature space to augment training data for model training. Jan et al.~\cite{jan2020throwing} generated labeled datasets for botnet detection using generative models. A bunch of prior works~\cite{yin2022practical,wang2020packetcgan,cheng2019pac} leveraged Generative Adversarial Nets (GANs) to synthesize metadata in the traffic. Unlike these works, \textsf{TrafficLLM} can generate entire packets including accurate headers and synthetic payloads based on traffic representation learning.

\section{Conclusion}\label{sec:conclusion}

In this paper, we develop a powerful framework to adapt LLMs for network traffic analysis with strong generalization. \textsf{TrafficLLM} employs three core techniques including traffic-domain tokenization to process instructions and traffic data, the dual-stage tuning pipeline to learn generic traffic representations with instructions and raw traffic data, and the EA-PEFT technique to update model parameters for new scenario adaptation. We evaluate \textsf{TrafficLLM} on 10 open-sourced datasets. Extensive experiments indicate that \textsf{TrafficLLM} shows remarkable generalization abilities across different tasks and unseen traffic scenarios compared to existing ML-based models. We release the source code and datasets to facilitate future research and hope that \textsf{TrafficLLM} can serve as the stepping stone for more LLM adaptation designs in the traffic analysis community.

\bibliographystyle{IEEEtran}
\bibliography{main}

{\appendices
\section{Open Science \& Ethical Considerations}\label{appendix-a}

\subsection{Open Science}
To facilitate future research to adapt LLMs in the traffic analysis domain, we release the source code of \textsf{TrafficLLM} and the datasets at \href{https://github.com/ZGC-LLM-Safety/TrafficLLM}{https://github.com/ZGC-LLM-Safety/TrafficLLM}. The dataset consists of over 0.4M tuning data that include human instructions and traffic features supervised by manual annotation. To the best of our knowledge, this is the largest LLM adaptation dataset for traffic domain to date. To enhance the reproducibility of our work, we mainly use open-sourced datasets and LLMs to conduct the experiments in this paper. All the original traffic data and LLMs are publicly available in their released repositories.


\subsection{Ethical Considerations}
In this paper, we discuss ethical considerations about dataset construction, human evaluation, and real-world deployment.

\vspace{0.1cm}
\noindent \textbf{Dataset Construction.} We leverage the human instruction and traffic data to construct the tuning data for LLM adaptation. The released traffic data are all extracted from the open-sourced datasets. For the 
natural language instruction dataset, we invite five network security practitioners and students to generate task-specific instructions with manual annotation and AI assistance. The instructions are required not including corporate confidentiality and privacy data. We have submitted the data review material to our institutional ethics review body (IRB). Our work has been approved by IRB to ensure ethical soundness and justification.

\vspace{0.1cm}
\noindent \textbf{Individual Use Evaluation.} We evaluate \textsf{TrafficLLM}'s effectiveness with extensive participation of players in the national LLM competition. All the players have signed an informed consent agreement to allow us to collect the competition data. In the competition system, each player is assigned a model submitter ID. We did not over-explore data involving personal information and designed a detailed exit mechanism to remove user records only necessary statistical data used in our experiments. The released statistical data has been approved by the competition organizer and IRB.

\vspace{0.1cm}
\noindent \textbf{Enterprise Deployment.} The recorded traffic has been anonymized and only contains necessary meta-information that is helpful for traffic detection. Any fields containing sensitive user information have been removed. We do not use the traffic data to identify individuals. All the traffic data used in the real-world evaluation has been approved by the company's IRB. The experiments are conducted in an isolated environment. We further guarantee that our evaluation process does not disrupt or harm other hosts or targets.

\begin{tcolorbox}[colback=gray!10, title=Instruction Examples of Different Tasks]
\textbf{TBD Task Instruction}: The Tor network transmits traffic through multiple layers of encryption and relay nodes, providing users with online privacy and anonymity. Please classify the traffic application or behavior according to the traffic data I provided.\\
\textbf{EAC Task Instruction}: Hi, I have noticed at home that my children spend a lot of time online and worry that they may be exposed to unsafe apps. I have obtained some network traffic data. Please help to analyze it to determine which applications the children are accessing. \\
\textbf{MDD Task Instruction}: The traffic may be DoH data generated by an encrypted DNS server. Please help me determine whether it contains malicious DoH behavior.\\
\textbf{WAD Task Instruction}: Analyze the pattern characteristics of the Web request data and determine whether it contains malicious attacks.\\
\textbf{BND Task Instruction}: The traffic data generated by the communication between the server and the client is obviously different from that of the normal network. The data is shown as follows. Please guess the botnet type to which this data belongs.\\
\textbf{MTD Task Instruction}: I have a piece of traffic data that may be malicious software communication behavior, but I don't know the specific software type behind the traffic. Could you provide more information? \\
\textbf{EVD Task Instruction}: I have used a specific encrypted VPN protocol to transmit the traffic generated by an application. Please check the application type in the traffic.\\
\textbf{CD Task Instruction}: I would like you to help analyze the following unknown traffic, determine whether there is a concept drift phenomenon, and identify the type of the traffic. Please note that the unknown patterns that may result from version updates require special attention.
\end{tcolorbox}

\section{Details of LLM Investigation}\label{appendix-a-existing-llm}

We investigate the capabilities of existing LLMs and earlier PLMs to achieve the traffic analysis tasks in Section II. For a more detailed study of existing LLMs, Table~\ref{tab:appendix-llm-response} shows examples of 3 state-of-the-art LLMs' responses to traffic analysis instructions. We can see that existing LLMs tend to give random guesses, only descriptions of the traffic data, or only recommendations of manual steps when receiving 2 types of traffic analysis instructions. For instance, when facing a traffic detection instruction of EAC or MTD tasks, native LLM usually generate a random label based on its knowledge (e.g., Alipay), give a detailed description of the traffic features (e.g., IP and protocol), and give a recommendation to use Wireshark or other network traffic analyzer. This investigation demonstrates that existing LLMs still cannot obtain accurate results for traffic detection and generation.

 \begin{table*}[t!]
\caption{Examples of Gemini (Gemini 1.5), Claude (Claude 3 Sonnet), and ChatGPT (GPT-4)'s responses to the instructions for conducting traffic detection and traffic generation tasks.}
\begin{center}
\vspace{-0.2cm}
\newcolumntype{M}[1]{>{\centering\arraybackslash}m{#1}}
\resizebox{\textwidth}{!}{
\begin{tabular}{c|l|l|l|l}
\toprule
\textbf{Task} & \textbf{Instruction} & \multicolumn{3}{l}{\textbf{Response}}\\
\midrule
\multirow{12}{*}{\rotatebox[origin=c]{90}{Traffic Detection}} & \ding{182} Please conduct the ENCRYPTED APP & \textbf{Gemini:} Based on the provided traffic & \textbf{Claude:} The most likely app category for & \textbf{ChatGPT:} To determine which class  \\
& CLASSIFICATION TASK to determine   & data, the predicted app category is: Alipay. &  the provided encrypted traffic is Apple. & encrypted traffic belongs to based on  \\
& which APP category the encrypted traffic & Explanation: The packet size and structure & The packet data contains patterns and   & packet data, a detailed analysis of   \\
& belongs to. The categories include 'Adobe, & may resemble previously observed patterns  &  characteristics that strongly suggest the & patterns and features within the \\
&Alipay, ..., and Booking'. <packet>: 04d9... & associated with Alipay's protocols ...& traffic is related to Apple services ... & encrypted traffic itself is required ...\\
 & \quad\quad\quad\quad\quad \textsf{[Encrypted App Classification]} & \quad\quad\quad\quad\quad\quad\quad\quad\quad\quad\quad \textsf{[Random Guess]}  & \quad\quad\quad\quad\quad\quad\quad\quad\quad\quad \textsf{[Random Guess]} & \quad\quad\quad\quad\quad \textsf{[Only Recommendations]}\\
\cmidrule{2-5}

& \ding{183} The following is a segment of network & \textbf{Gemini:} The provided packet snippet  & \textbf{Claude:} Without additional context or  & \textbf{ChatGPT:} Considering the DNS query   \\
& traffic data, you need to analyze whether it & appears to be a DNS query, but without  & analysis, it's difficult to definitively  & characteristics, and without additional \\
& is generated by the communication behavior & additional context and analysis, it's difficult  & determine if this specific packet is gener- &  payload decryption or further traffic  \\
& of malware. If so, please categorize it as & to definitively determine if it's generated & ated by malware communication behavior.   & analysis, this packet could be related \\
& Cridex, Geodo, ..., and Zeus. <packet>: 04...& by malware. However, we can examine ...& However, I can provide an overview ... & to families like Zeus or Shifu ...\\
& \quad\quad\quad\quad\quad \textsf{[Malware Traffic Detection]}& \quad\quad\quad\quad\quad\quad\quad\quad\quad\quad \textsf{[Only Description]} & \quad\quad\quad\quad\quad\quad\quad\quad\quad\quad \textsf{[Only Description]} & \quad\quad\quad\quad\quad\quad\quad\quad\quad \textsf{[Random Guess]} \\
\midrule

\multirow{10}{*}{\rotatebox[origin=c]{90}{Traffic Generation}} & \ding{182} Based on the protocol of fields, traffic & \textbf{Gemini:} Due to the limitations of my & \textbf{Claude:} Unfortunately, I cannot generate & \textbf{ChatGPT:} IP Header: Source IP: 192.16 \\
& features, and payloads of different types of & knowledge cutoff and the dynamic nature of  & packet capture (pcap) files or raw network   & 8.1.2, Destination IP: 216.58.214.14, Pro- \\
& traffic in your knowledge. Please generate & YouTube traffic, I cannot directly generate a  & traffic data. As an AI assistant without & tocol: TCP, TCP Header: Source Port: \\
& a packet of YouTube with the pcap format. & specific packet in hexadecimal format ...  &  direct access to network devices or the ... &  49152, Destination Port: 443, Flags: ...\\
& \quad\quad\quad\quad\quad\quad \textsf{[Youtube Traffic Generation]}& \quad\quad\quad\quad\quad\quad\quad \textsf{[Only Recommendations]} & \quad\quad\quad\quad\quad\quad\quad \textsf{[Only Recommendations]} & \quad\quad\quad\quad\quad\quad\quad \textsf{[Random Generation]}\\
\cmidrule{2-5}

& \ding{183} I need a packet of Weibo traffic. Based  & \textbf{Gemini:} Unfortunately, due to limitations & \textbf{Claude:} I'm afraid I cannot provide & \textbf{ChatGPT:}  To obtain a real packet of   \\
& on Weibo's traffic feature, please provide &  in my capabilities and knowledge cutoff, I & you with an actual packet capture (pcap) & Weibo's traffic in PCAP format, you  \\
&  me with a packet of Weibo's traffic  & cannot directly generate a pcap file & file containing Weibo traffic data. & would typically use a network traffic \\
&  data with the pcap format. & containing Weibo traffic. However, ... & As an AI assistant without direct access ... & capturing tool like Wireshark ... \\
& \quad\quad\quad\quad\quad\quad\quad \textsf{[Weibo Traffic Generation]}& \quad\quad\quad\quad\quad\quad\quad \textsf{[Only Recommendations]} & \quad\quad\quad\quad\quad\quad\quad \textsf{[Only Recommendations]} & \quad\quad\quad\quad\quad\quad \textsf{[Only Recommendations]}\\
\bottomrule
\end{tabular}
}
\label{tab:appendix-llm-response}
\end{center}
\vspace{-0.1cm}
\end{table*}

\section{Details of Experimental Settings}
\subsection{Instruction Dataset}\label{appendix-b-instruction-dataset}
The human instructions for LLM adaptation are constructed by 5 experts in the network security industries and PH.D. students majoring in cybersecurity. They are required to generate instruction templates using domain knowledge across different traffic detection and generation tasks and extend the scale via generative AI assistants. We generate expert instructions according to the backgrounds of 10 traffic classification tasks and 229 types of traffic generation tasks. After collecting professional instructions with extensive traffic-domain knowledge from security experts, we use them as templates for AI assistants to supplement the instruction dataset. We collect diverse AI-generated instructions using ChatGPT~\cite{ChatGPT} with a rewrite instruction like `The following is a traffic analysis instruction provided by network security experts. Please consider different scenarios, security goals, question subjects, writing styles, and text descriptions. Rewrite the following instructions and generate 20 new different traffic analysis instructions'. All the generated instructions have been supervised by manual annotation. We removed the overlapped text and mistakes. We released all our collected traffic-domain instructions for future research. We show instruction examples of partial tasks.

\subsection{Description of Baselines}\label{baselines}
In this paper, we compare the performance of \textsf{TrafficLLM} with 15 baselines across different tasks, including state-of-the-art traffic detection and generation algorithms.

\vspace{0.1cm}
\noindent \textbf{Traffic Detection.} The following provides an overview of the traffic classification baseline used in our evaluation.
\begin{itemize}[nolistsep,leftmargin=*]
\item \textbf{AppScanner}~\cite{taylor2017robust}. AppScanner is a statistical feature method using 54 statistical features for smartphone App identification based on ML models.
\item \textbf{CUMUL}~\cite{panchenko2016website}. CUMUL extracts packet size, direction, and ordering features for website fingerprinting on the Tor network.
\item \textbf{BIND}~\cite{al2016adaptive}. BIND uses bi-directional burst features for encrypted traffic fingerprinting in a wide range.
\item \textbf{k-fingerprinting (K-FP)}~\cite{hayes2016k}. K-FP extracts the most important features using random forests for large-scale traffic classification.
\item \textbf{FlowPrint}~\cite{van2020flowprint}. FlowPrint extracts statistical features for encrypted traffic classification tasks using the semi-supervised clustering method.
\item \textbf{FS-Net}~\cite{liu2019fs}. FS-Net uses RNN models to learn packet length sequences for traffic classification.
\item \textbf{Deep Fingerprinting (DF)}~\cite{sirinam2018deep}. DF uses deep Convolutional Neural Networks to realize website fingerprinting.
\item \textbf{GraphDApp}~\cite{shen2021accurate}. GraphDApp extracts traffic interaction graphs and uses Graph Neural Networks to detect traffic. 
\item \textbf{TSCRNN}~\cite{lin2021tscrnn}. TSCRNN combines CNN and RNN to extract abstract features of the flow for traffic classification.
\item \textbf{Deeppacket}~\cite{lotfollahi2020deep}. Deeppacket uses deep Convolutional Neural Networks to realize encrypted traffic classification.
\item \textbf{PERT}~\cite{he2020pert}. PERT uses a Transformer-based encoder and pre-training technique to learn traffic representation.
\item \textbf{ET-BERT}~\cite{lin2022bert}. ET-BERT pre-trains BERT with large-scale traffic to realize traffic classification across different tasks.
\end{itemize}

\vspace{0.1cm}
\noindent \textbf{Traffic Generation.} We use the following traffic generation algorithms as the baselines in the experiments.
\begin{itemize}[nolistsep,leftmargin=*]
\item \textbf{Netshare}~\cite{yin2022practical}. Netshare uses the GAN-based method to generate IP header traces at a large scale. 
\item \textbf{PacketCGAN}~\cite{cheng2019pac}. PacketCGAN uses conditional GANs to generate the encrypted traffic using bit vectors.
\item \textbf{PAC-GAN}~\cite{wang2020packetcgan}. PAC-GAN uses CNN GAN to encode packets into images and use them to generate packets.
\end{itemize}
﻿
\noindent Note that we choose these representative methods to cover a wide range of traffic detection and generation tasks. To help all baselines be evaluated reasonably, we use the raw traffic data for most of datasets and preprocess them following the baselines' instructions in their repositories. We contacted the authors to obtain the datasets that do not open-source their raw PCAP files. We keep similar setups to prior work~\cite{lin2022bert,zhang2023tfe} to evaluate baselines on different scenarios.

\begin{table}[t]
\caption{The details of traffic data in the evaluation under the LLM competition and enterprise deployment.}
\vspace{-0.3cm}
\begin{center}
\resizebox{0.48\textwidth}{!}{
\begin{tabular}{c|c|cccc}
\toprule
\textbf{Experiment} & \textbf{Task} & \textbf{\#Flows} & \textbf{\#Packet}& \textbf{\#Labels}\\
\midrule
\multirow{2}{*}{Evaluation on} & MTD & 2,855 & 9,458 & 20\\
\multirow{2}{*}{the LLM Competition} & BND & 16,930 & 52,278 & 5 \\
& EVD & 1,025 & 3,392 & 19\\
\midrule
\midrule
\textbf{Experiment} & \textbf{Task} & \textbf{\#Malicious} & \textbf{\#Benign}& \textbf{\#Labels}\\
\midrule
\multirow{2}{*}{Enterprise Deployment} & MTD & 17,556 & 219,450 & 2\\
& WAD & 7,083 & 215,323 & 2\\
\bottomrule
\end{tabular}
}
\label{tab11}
\vspace{-0.1cm}
\end{center}
\end{table}

\begin{figure*}[t]
\centering
\subfigure[5-Class packets generated from ISCX Botnet 2014]{    
\includegraphics[width=8.5cm]{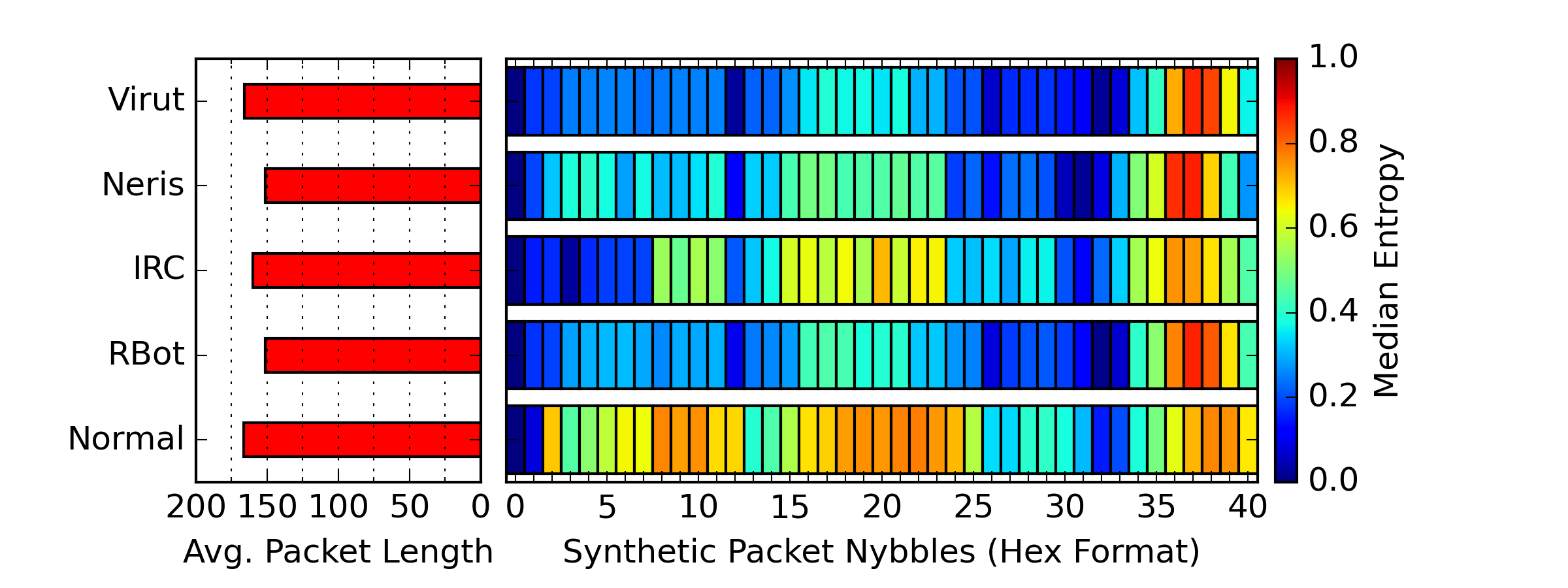}
}
\subfigure[10-Class packets generated from CSTNET 2023]{ 
\includegraphics[width=8.5cm]{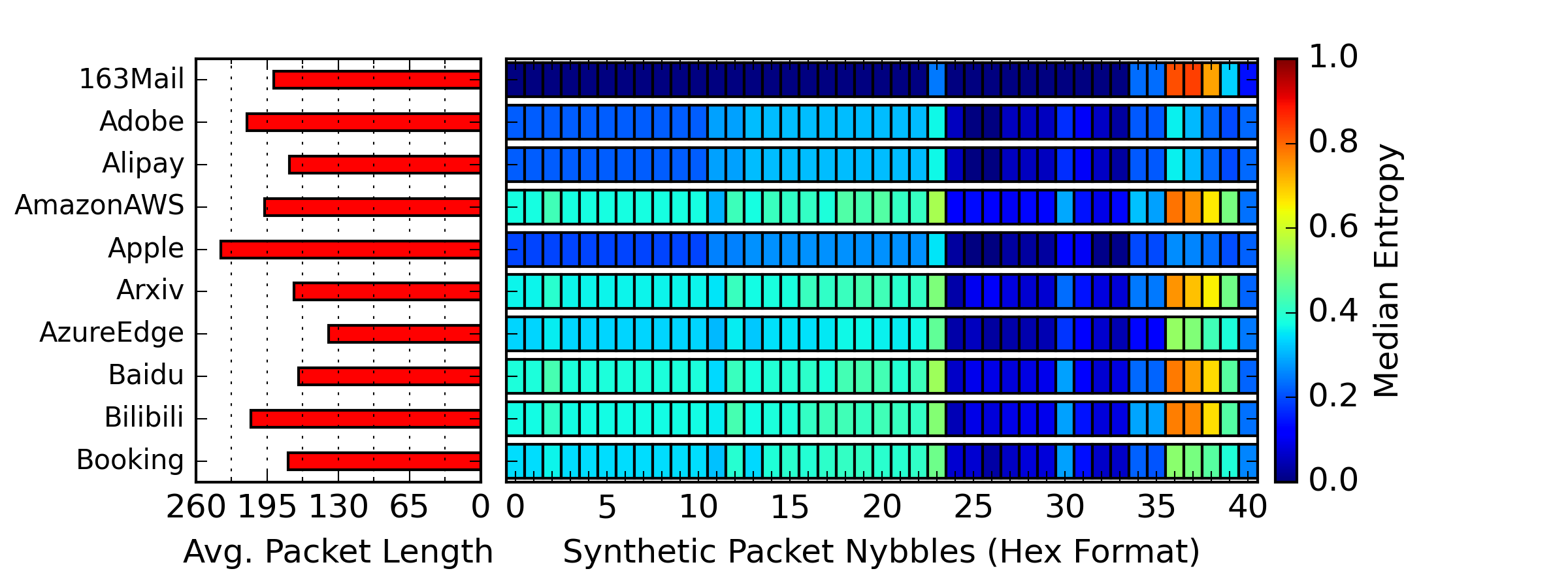}
}
\setlength{\abovecaptionskip}{-0.01cm}
\caption{Entropy analysis of synthetic packets on ISCX Botnet 2014 and CSTNET 2023. In each index, a higher entropy score (red) refers to more diverse nybbles generated; A lower entropy score (blue) refers to more fixed nybbles generated.}
\centering
\label{fig9}
\end{figure*}

\subsection{Real-World Evaluation Setup}\label{real-world}
\noindent \textbf{Individual Use Evaluation Setup.} To investigate the effectiveness of deploying \textsf{TrafficLLM} in the community, we conduct an extensive evaluation of \textsf{TrafficLLM} by integrating the framework as a track in a national LLM competition. In this track, we collect the traffic including 20 types of malware, 5 types of botnets, and 19 types of VPN-encrypted Apps in our experimental environments. The competition requires players to build traffic-domain LLMs using \textsf{TrafficLLM}'s framework with strong performance on all 3 tasks. Table~\ref{tab11} shows the detail of the competition dataset to evaluate \textsf{TrafficLLM}. As one of the organizers of the competition, we join the development of the competition platform. The platform can run the model image submitted by players and record the score on a real-time ranking web page. Each player is allocated an account to log in to the fortress machine and access the development machine to build their models. The players must upload their Docker~\cite{docker} images to the cloud environment to load GPU servers. We use average accuracy (Acc) as the metric and build the scoring procedures to update player ranking. At the end of the competition, we analyze the overall performance of all players to evaluate \textsf{TrafficLLM}'s practicality. 

\vspace{0.1cm}
\noindent \textbf{Enterprise Deployment Setup.} In the real-world deployment scenario, we implement \textsf{TrafficLLM} prototype and deploy it in a top global security company. The company has supported signature-based WAF and NIDS services for over 20 years. It has provided security services for 100K+ enterprises. To evaluate \textsf{TrafficLLM}'s practical performance, we collected 17,556 and 7,083 flows of malware and Web attack traffic according to WAF rules hitting and manual analysis. The Web attack traffic contains many attack payloads such as command injection, which is different from normal user behavior of Web applications. The malware traffic includes the communication behavior of malware such as Cobalt Strike~\cite{yang2024petnet}. Otherwise, the traffic that bypasses the rules and is confirmed by the security engineers is considered benign traffic. Table~\ref{tab11} shows the detailed setting of the dataset. We build \textsf{TrafficLLM} and its API in an isolated environment with a super GPU server. We leverage \textsf{TrafficLLM}'s API to conduct MTD and WAD tasks using its command mode. The security engineers report F1-scores and false positives to evaluate the performance in real-world scenarios.

\section{Details of Evaluation}\label{detail-evaluation}
\subsection{Packet Generation}\label{packet-generation}

\textsf{TrafficLLM} can rebuild the raw traffic across different scenarios using the representation learned from the tuning stage. Leveraging the generated fine-grained meta-information, \textsf{TrafficLLM} can be equipped with packet manipulation tools like Scapy~\cite{scapy} to generate PCAP packets. These synthetic packets follow standard formats that can be completely readable by Wireshark~\cite{wireshark}. \textsf{TafficLLM} does not directly generate the Ethernet layer since it often varies depending on the different types of physical devices at the viewpoints. We use Scapy to synthesize a default Ethernet layer for every generated packet. To further analyze the integrity of the generated packets, Figure~\ref{fig9} shows the entropy analysis results of the packets generated from ISCX Botnet 2014 and CSTNET 2023. We find that \textsf{TrafficLLM} can learn fixed field values (e.g., TCP Flags) effectively and imitate changed field values (e.g., Source and Destination Port) according to the distribution in the dataset. Moreover, \textsf{TrafficLLM} can capture the traffic characteristics for different network scenarios. For instance, \textsf{TrafficLLM} keeps an average entropy of 0.55 to generate the packet nybbles of the target botnet (i.e., Virut, Neris, IRC, and RBot), while the normal network is 0.64 due to the diversity of protocols. 

\begin{figure*}[t!]
\centering
\subfigure[LLM Instruction Embeddings]{    
\vspace{-0.03cm}
\includegraphics[width=3.8cm]{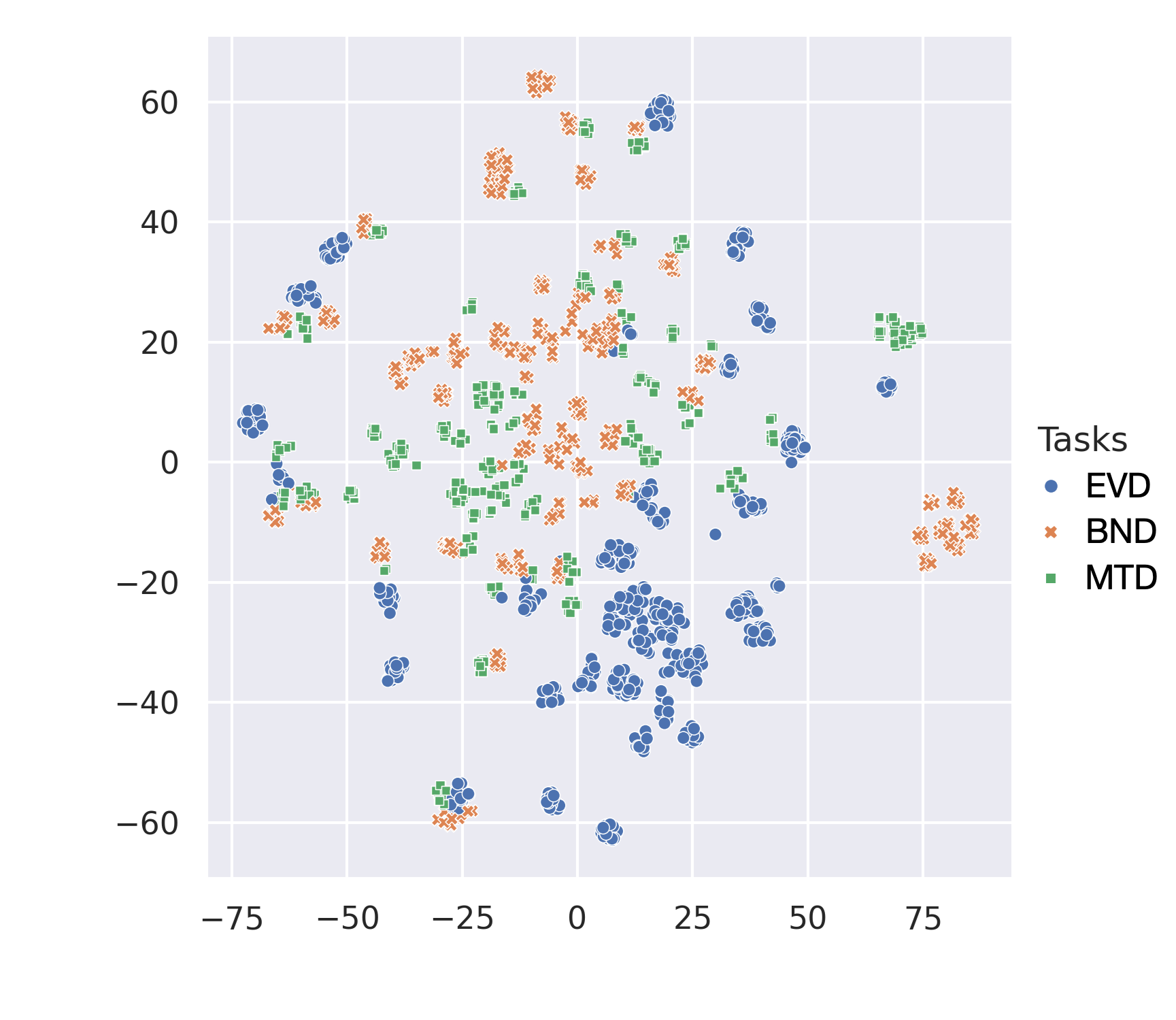}
}\label{fig11a}
\subfigure[\textsf{TrafficLLM} Instructions]{ 
\vspace{-0.03cm}
\includegraphics[width=3.8cm]{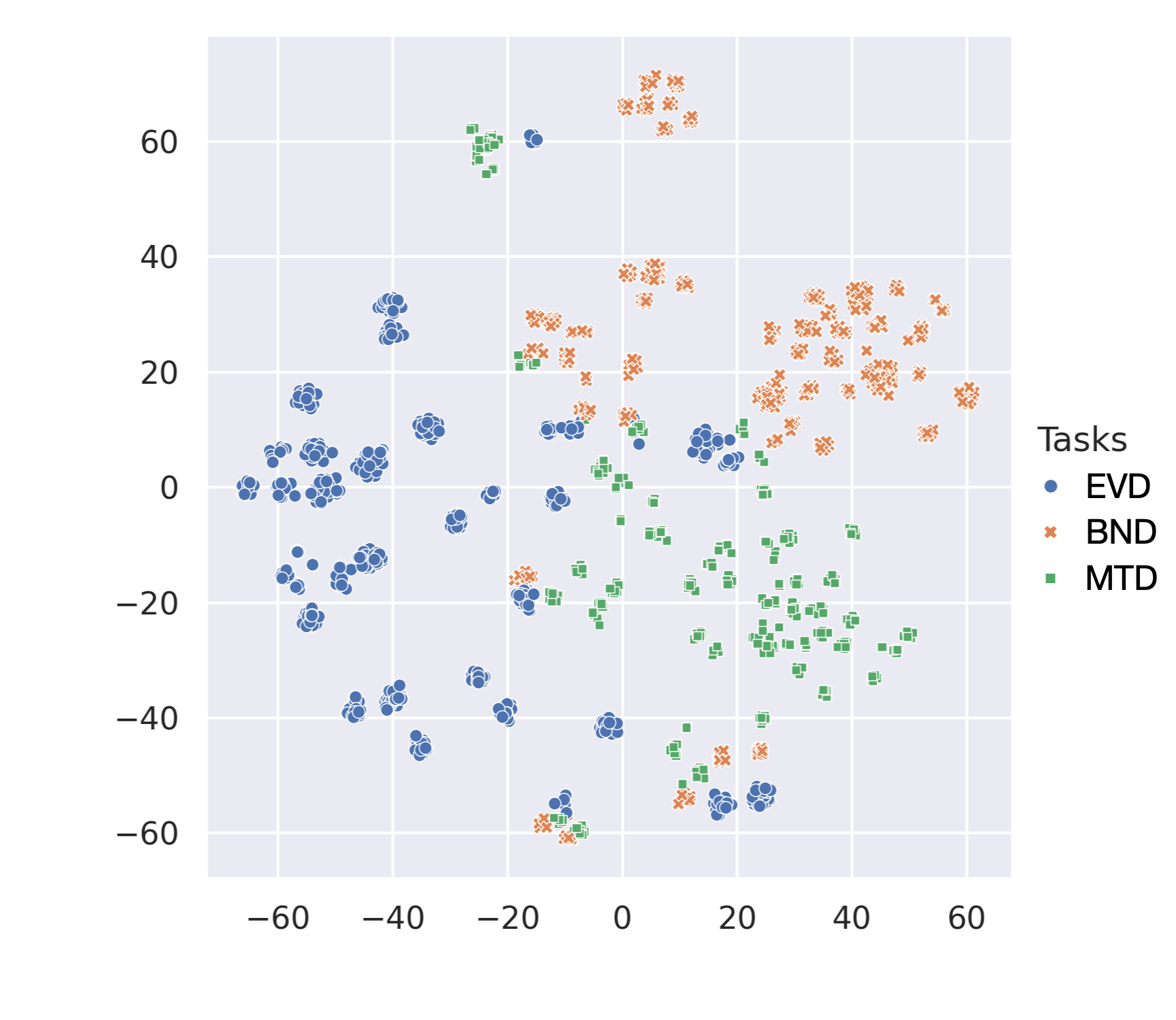}
}\label{fig11b}
\subfigure[LLM Traffic Embeddings]{ 
\vspace{-0.03cm}
\includegraphics[width=4.1cm]{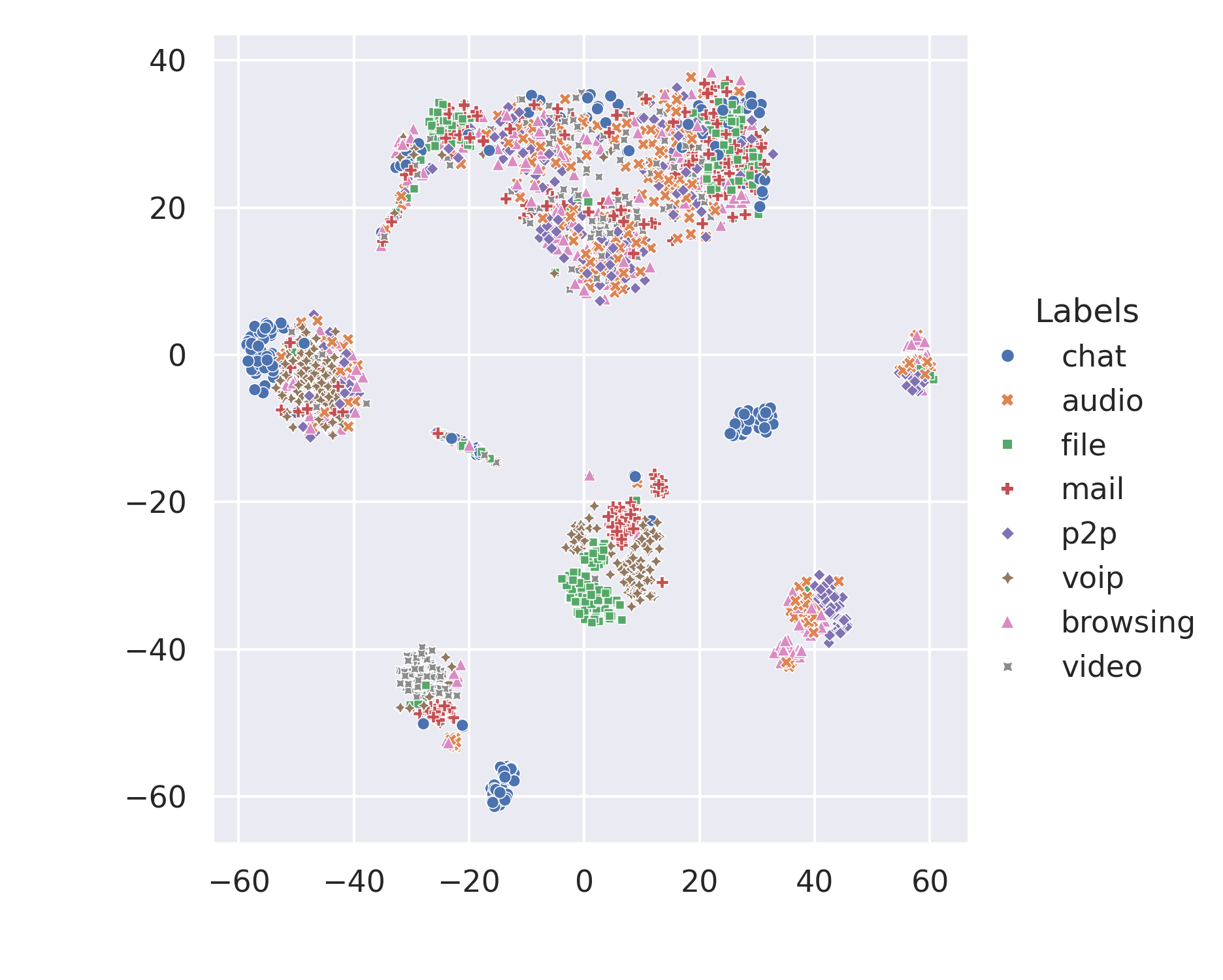}
}\label{fig11c}
\subfigure[\textsf{TrafficLLM} Traffic Embeddings]{ 
\vspace{-0.03cm}
\includegraphics[width=4.1cm]{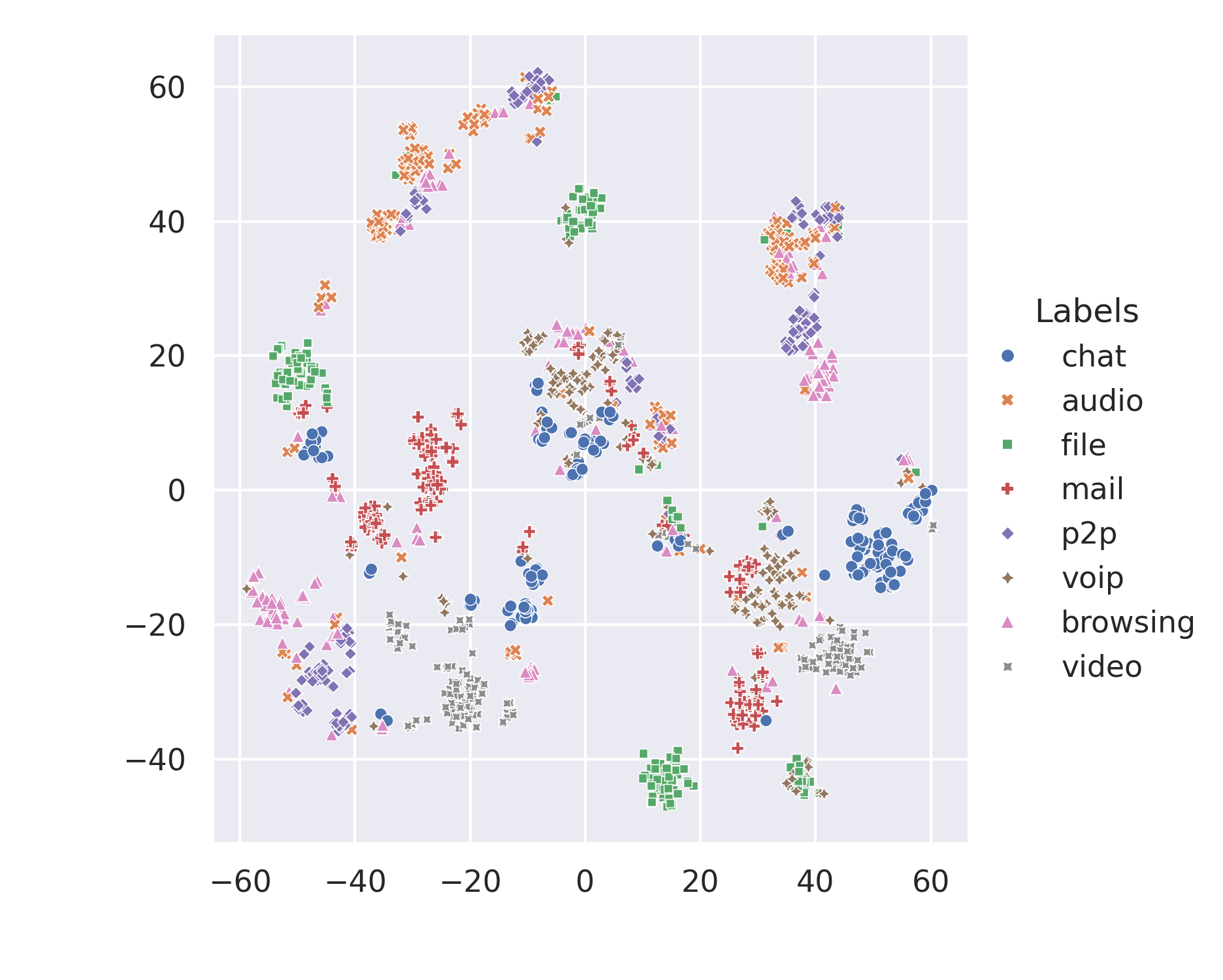}
}\label{fig11d}
\centering
\setlength{\abovecaptionskip}{-0.03cm}
\caption{The hidden state visualization of task instructions and traffic data in ChatGLM2 and \textsf{TrafficLLM}. \textsf{TrafficLLM} can learn better representations under traffic detection instructions of EVD, BND, and MTD tasks and ISCX Tor 2016 traffic datasets.} 
\label{fig16}
\end{figure*}

\subsection{Efficiency Evaluation}\label{efficiency}
Model quantization~\cite{xu2023survey} is a widely used technique to deploy LLM-based systems in real-world scenarios. To further evaluate \textsf{TrafficLLM}'s efficiency, we employ INT4 quantization to help \textsf{TrafficLLM} overcome LLM's computation overhead and reach faster inference speed. We observe that \textsf{TrafficLLM}'s average prediction latency for a sample is 0.1408s, which is 42\% better than no quantization on \textsf{TrafficLLM} (BF16). \textsf{TrafficLLM} only consumes 8.3GB memory, which is a 3.5-fold decrease. The performance has almost no degradation. Additionally, we compare \textsf{TrafficLLM}'s inference latency with existing ML-based models. \textsf{TrafficLLM} incurs lower detection latency which is 3.51 times lower than that of the existing method, i.e., FS-Net, which incurs 0.4950s per flow latency. It is attributed to traffic-domain tokenization and quantization techniques to achieve a significant reduction on the traffic input and model overhead. 

\begin{figure}[t]
\centering
\subfigure[RGE and SMS Metrics]{    
\includegraphics[width=4.1cm]{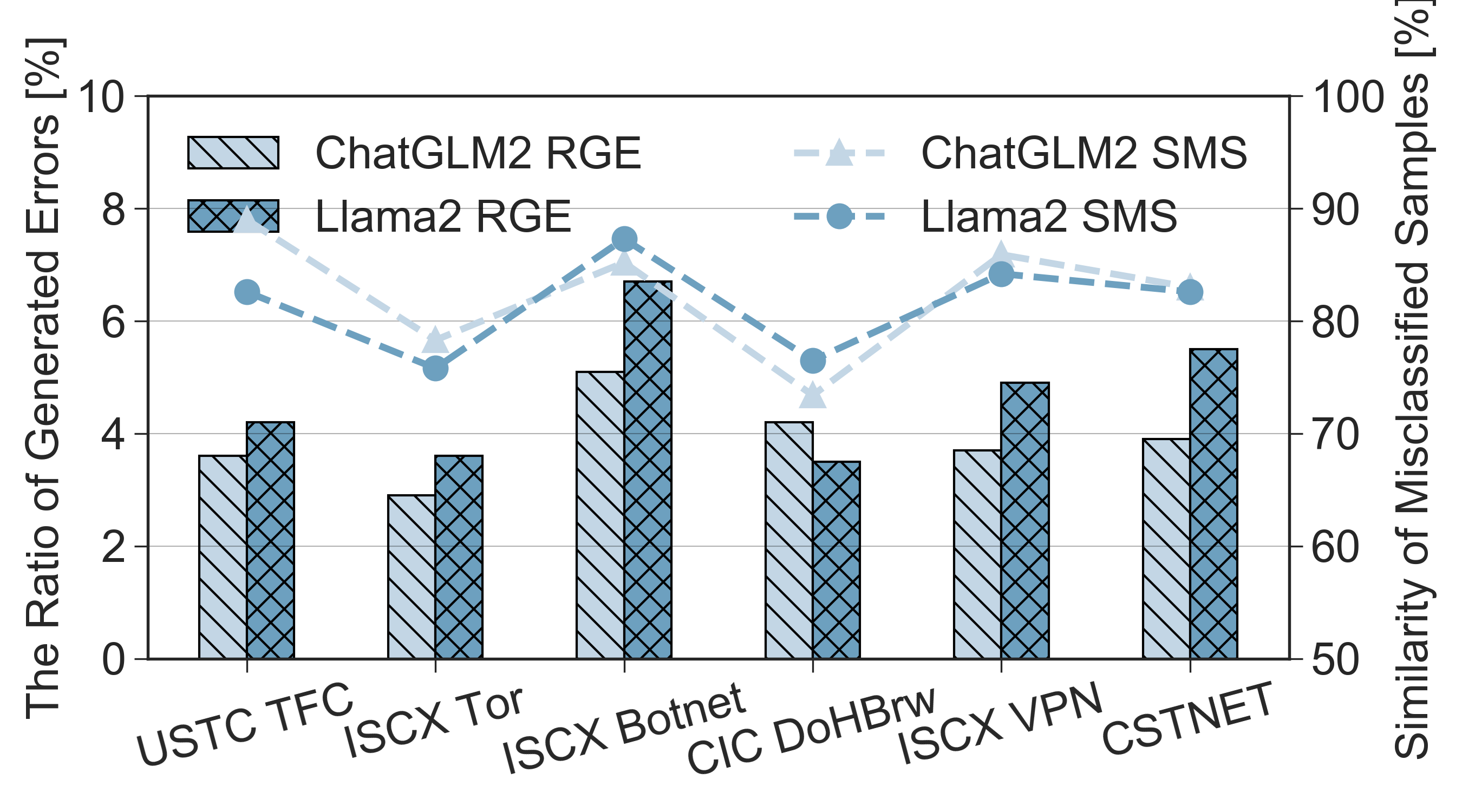}
}\label{tc_error}
\subfigure[Top-p and Temperature]{ 
\includegraphics[width=4.1cm]{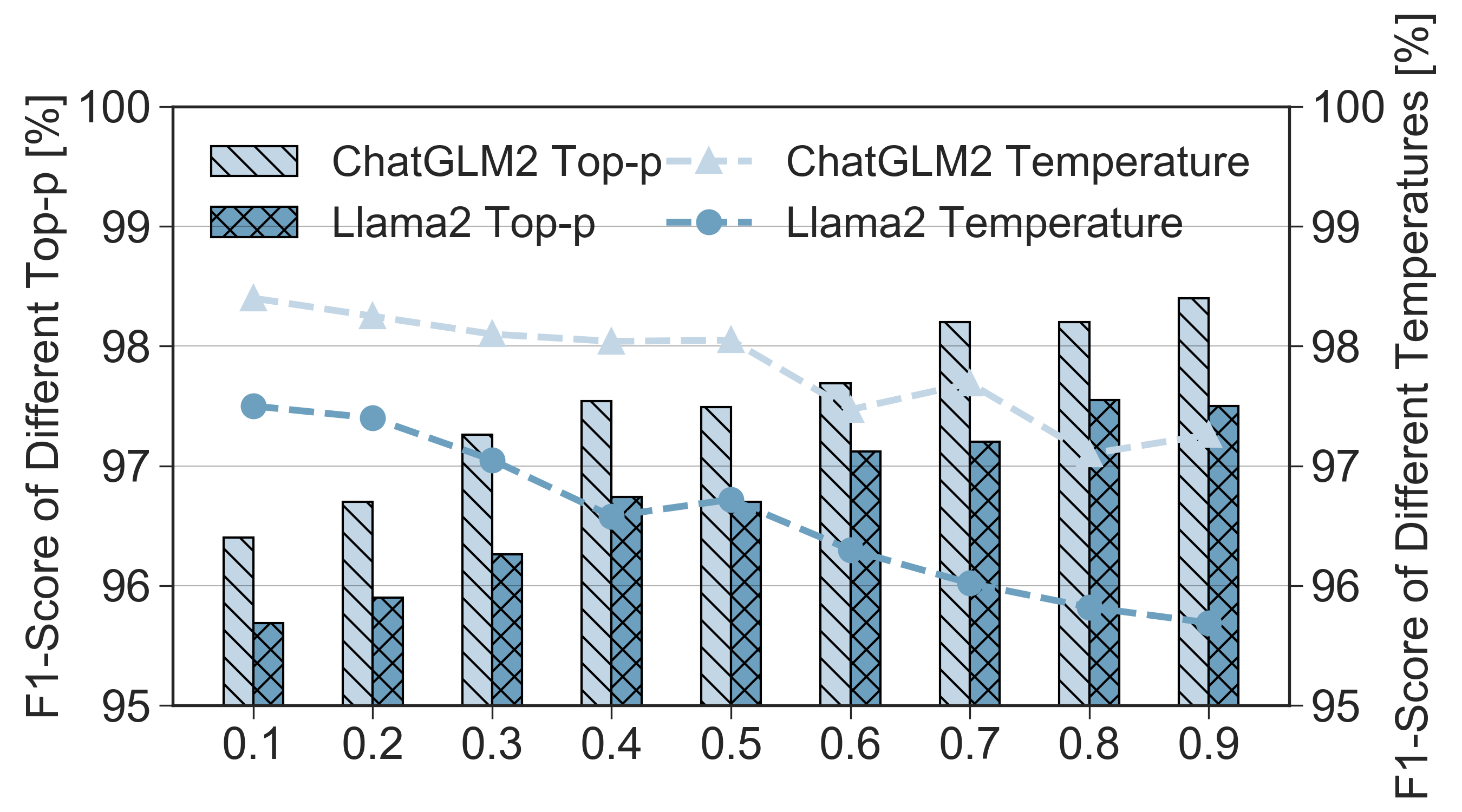}
}\label{tc_p}
\centering
\setlength{\abovecaptionskip}{-0.01cm}
\caption{Left: The ratio of generated errors (RGE) and the similarity of misclassified samples (SMS) metric; Right: The performance of different top-p and temperature.} 
\label{fig17}
\vspace{-0.2cm}
\end{figure}

\subsection{Hallucination Evaluation}\label{hallucination}
LLM is prone to hallucination~\cite{JiLFYSXIBMF23} due to the token generation with word sampling strategies. This may influence the detection accuracy during the inference. In the hallucination evaluation experiment, we collect the prediction results from 10 rounds of inference using the same test set. We define the different responses for the same test sample input as the generated errors. Figure 17-a shows the ratio of generated errors in the misclassified samples and the average Jaccard similarity between the misclassified samples and the dataset of the misclassified label. Results indicate that 3.9\% and 4.7\% of the misclassified output are generated errors on average when using ChatGLM2 and Llama2 as the foundation model in \textsf{TrafficLLM}. These misclassified samples usually keep a high similarity (i.e., 82.4\% and 81.5\% on average) to the datasets of misclassified labels, which we consider as the reason for raising the hallucination issues. To address the problem, Figure 17-b measures the performance of different Top-p and Temperature parameters setting in the sampling strategy. A higher Top-p and a lower Temperature parameter can help the model keep strong confidence when predicting labels and mitigate the hallucination in traffic detection.

\begin{table}[t]
\caption{The task understanding abilities of \textsf{TrafficLLM} and the native LLM based on instructions of downstream tasks.}
\vspace{-0.4cm}
\begin{center}
\resizebox{0.48\textwidth}{!}{
\begin{tabular}{c|l|cccc}
\toprule
\textbf{Model} & \textbf{Task} & \textbf{PR} & \textbf{RC} & \textbf{F1} & \textbf{Acc} \\
\midrule
Native LLM& Traffic Detection & 0.4422 & 0.6650 & 0.5312 & 0.6650\\
(Llama2-7B)& Traffic Generation & 0.5776 & 0.7600 & 0.6564 & 0.7600\\
\midrule
\multirow{2}{*}{\textsf{TrafficLLM}}& Traffic Detection & 0.9910 & 0.9925 & 0.9915 & 0.9925 \\
 & Traffic Generation & 0.9935 & 0.9960 & 0.9940 & 0.9960 \\
\bottomrule
\end{tabular}
}
\label{tab:task-understanding}
\end{center}
\end{table}

\subsection{Representation Learning}\label{representation}
To further indicate the robustness of \textsf{TrafficLLM} on different datasets, we show \textsf{TrafficLLM}'s robust traffic representation in the vector space and evaluate its performance by using different instructions. 

\vspace{0.1cm}
\noindent \textbf{Hidden State Visualization.} To explain \textsf{TrafficLLM}'s ability to learn text and traffic data, in Figure~\ref{fig16}, we show the visualization of \textsf{TrafficLLM}'s hidden state representation using T-SNE. Compared to the native LLM, \textsf{TrafficLLM} can learn more distinguishable representations of different traffic analysis instructions through instruction tuning. Moreover, \textsf{TrafficLLM} also learns better traffic representation for each class due to the task-specific traffic tuning with the specialized traffic tokens. \textsf{TrafficLLM}'s traffic representations of each type keep clearer boundaries in the feature space, which ensures accuracy across different tasks.

\vspace{0.1cm}
\noindent \textbf{Task Understanding.} The representation learning ability helps \textsf{TrafficLLM} correctly understand different task instructions from security practitioners. In Table~\ref{tab:task-understanding}, we compare \textsf{TrafficLLM}'s performance to the native Llama2-7B, which is required to choose the correct task labels based on the instructions and given options across different detection and generation tasks. Results indicate that \textsf{TrafficLLM} has effectively acquired the domain knowledge for traffic analysis, achieving strong task understanding performance (0.9920 average F1-score) to conduct different downstream tasks.

}

\vfill

\end{document}